\documentclass[10pt,journal,compsoc,final]{IEEEtran}
\usepackage{float}
\usepackage{hyperref}

\usepackage{wrapfig}
\usepackage{mathtools}
\DeclarePairedDelimiter{\floor}{\lfloor}{\rfloor}
\usepackage{amssymb}
\usepackage{gensymb}
\usepackage{multirow}
\usepackage{ragged2e}
\usepackage{makecell}

\usepackage{xcolor}
\newcommand{\Blue}[1]{\textcolor{black}{#1}}

\usepackage{pifont}
\newcommand{\cmark}{\ding{51}}%
\newcommand{\xmark}{\ding{55}}%

\ifCLASSOPTIONcompsoc
  \usepackage[nocompress]{cite}
\else
  \usepackage{cite}
\fi
\ifCLASSINFOpdf
  \usepackage[pdftex]{graphicx}
\else
\fi
\usepackage{amsmath}
\ifCLASSOPTIONcompsoc
  \usepackage[caption=false,font=footnotesize,labelfont=sf,textfont=sf]{subfig}
\else
  \usepackage[caption=false,font=footnotesize]{subfig}
\fi
\usepackage{url}

\begin{document}
%
\title{TopoTag: A Robust and Scalable \\ Topological Fiducial Marker System}
%
%
%
%

\author{Guoxing Yu,~
        Yongtao Hu,~
        Jingwen Dai,~\IEEEmembership{Member,~IEEE}
\IEEEcompsocitemizethanks{\IEEEcompsocthanksitem G. Yu, Y. Hu and J. Dai are with Guangdong Virtual Reality Technology Co., Ltd. (aka. Ximmerse)\protect\\
E-mail: \{calvin.yu, ythu, dai\}@ximmerse.com}
\thanks{Manuscript received [Month] [Date], [Year]; revised [Month] [Date], [Year].}
}

\IEEEtitleabstractindextext{%
\begin{abstract}
\justifying
Fiducial markers have been playing an important role in augmented reality (AR), robot navigation, and general applications where the relative pose between a camera and an object is required. Here we introduce TopoTag, a robust and scalable topological fiducial marker system, which supports reliable and accurate pose estimation from a single image. TopoTag uses topological and geometrical information in marker detection to achieve higher robustness. \Blue{Topological information is extensively used for 2D marker detection, and further corresponding geometrical information for ID decoding. Robust 3D pose estimation is achieved by taking advantage of all TopoTag vertices.} Without sacrificing bits for higher recall and precision like previous systems, TopoTag can use full bits for ID encoding. TopoTag supports tens of thousands unique IDs and easily extends to millions of unique tags resulting in massive scalability. We collected a large test dataset including in total 169,713 images for evaluation, involving in-plane and out-of-plane rotation, image blur, different distances and various backgrounds, etc. Experiments \Blue{on the dataset and real indoor and outdoor scene tests with a rolling shutter camera both} show that TopoTag significantly outperforms previous fiducial marker systems in terms of various metrics, including detection accuracy, vertex jitter, pose jitter and accuracy, etc. In addition, TopoTag supports occlusion as long as the main tag topological structure is maintained and allows for flexible shape design where users can customize internal and external marker shapes. Code for our marker design/generation, marker detection, and dataset are available at \url{http://herohuyongtao.github.io/research/publications/topo-tag/}.
\end{abstract}

}

\maketitle

\IEEEdisplaynontitleabstractindextext

%
\IEEEpeerreviewmaketitle

\IEEEraisesectionheading{\section{Introduction}\label{sec:introduction}}

%
%
%
%
\IEEEPARstart{I}{n} this paper, we introduce TopoTag, a new fiducial marker and detection algorithm that is more robust and accurate than current fiducial marker systems. Fiducial markers are artificial objects (typically paired with a detection algorithm) designed to be easily detected in an image from a variety of perspectives. They are widely used for augmented reality and robotics applications because they enable localization and landmark detection in featureless environments \cite{Degol2017}. Previous work on fiducial markers mainly focus on one or more of the following areas: (1) improving detection accuracy via specialized tag design \cite{Fiala2005, Fiala2010, Atcheson2010, Goyal2011}; (2) reducing pose estimation error via precise vertex estimation \cite{Wang2016} or introducing more feature points \cite{Bergamasco2011}; (3) increasing unique identities \cite{Cho1998, Flohr2007, Neto2010, Garrido-Jurado2015}; (4) improving robustness under occlusion \cite{Bergamasco2011, Garrido2014} and other use cases \cite{Prasad2015, Calvet2016, birdal2016x, Cruz-Hernandez2018} and (5) speed-up \cite{Molineros2001, Wang2016, Degol2017, Romero-Ramirez2018}.

\begin{figure}[tp]
\centering
\subfloat{\includegraphics[height=0.135\textwidth]{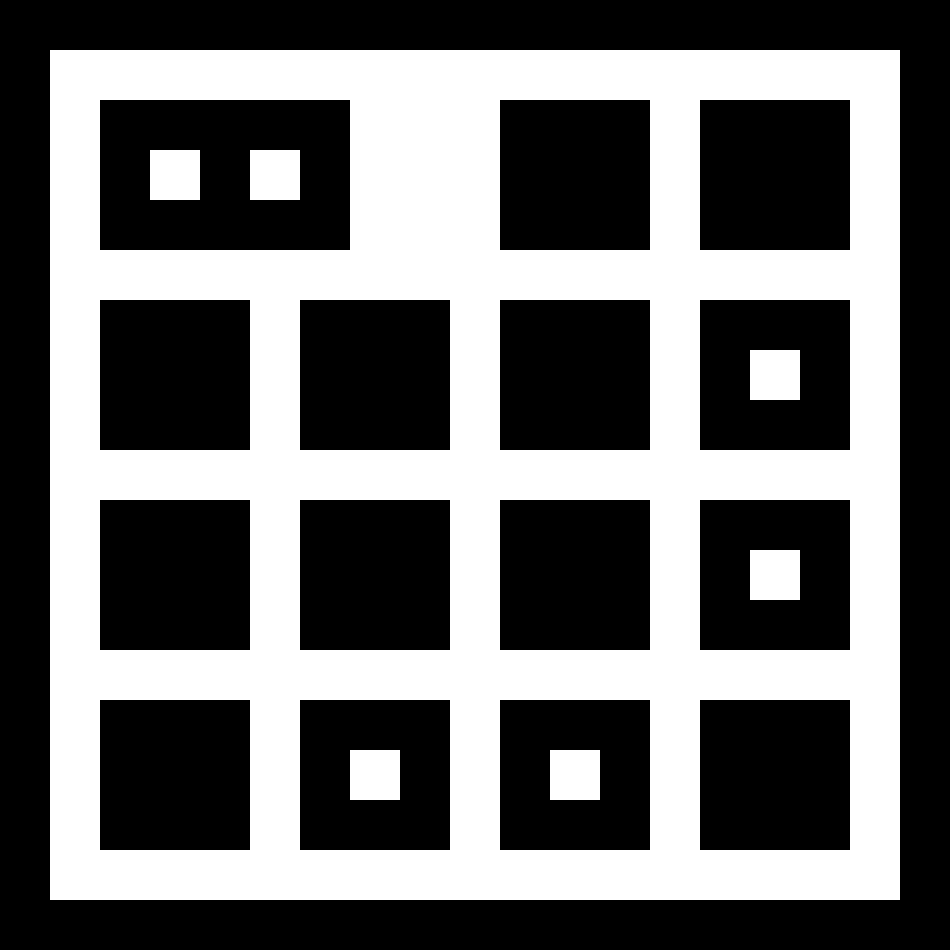}}
\hspace{0.06in}
\subfloat{\includegraphics[height=0.135\textwidth]{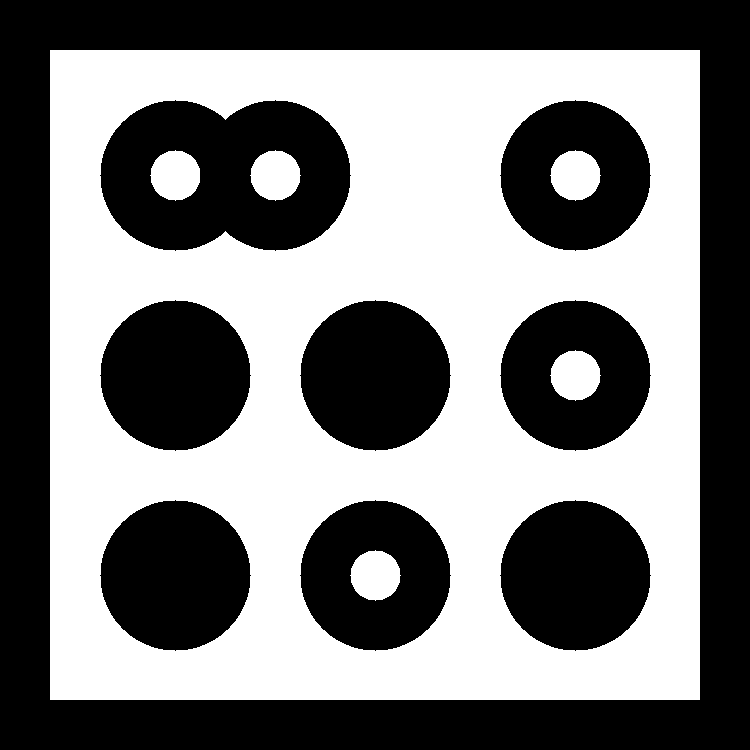}}
\hspace{0.04in}
\subfloat{\includegraphics[height=0.135\textwidth]{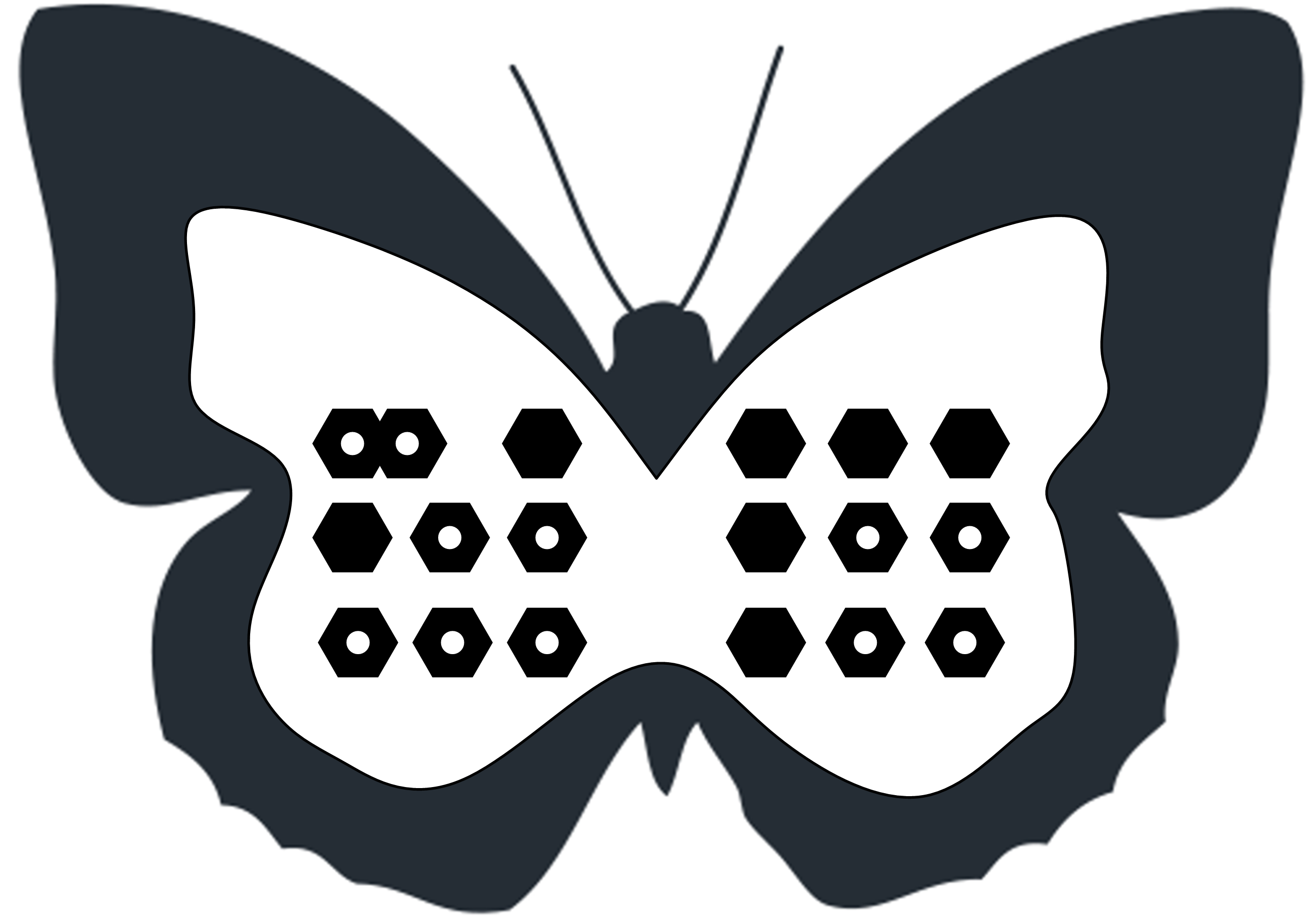}}
\caption{Three TopoTag markers. \Blue{TopoTag supports both customized internal and external shapes. Here shows three TopoTags with various internal shapes like squares, circles, hexagons and different external shapes like square and butterfly.}}
\label{fig:topotag}
\end{figure}

TopoTag utilizes topological information in tag design to improve robustness, which achieves perfect detection accuracy on the large dataset we collected and on datasets from others. We show that all tag bits can be used to encode identities without sacrificing detection accuracy, thus achieving rich identification and massive scalability. In addition, TopoTag offers more feature point correspondences for better pose estimation. Results show that TopoTag achieves the best performance in vertex jitter, pose error and pose jitter. TopoTag also supports occlusion and noise, to some extent, if the main tag topological structure is maintained and supports flexible shape design where users can customize internal and external marker shapes.
\Blue{\autoref{fig:topotag} shows three TopoTag markers.}

We collected a large dataset including 169,713 images with TopoTag and several state-of-the-art tag systems. A robot arm is used to make sure each tag has the same trajectory for consistent comparison. The rich modalities of the dataset include in-plane and out-of-plane rotations, image blur, different distances and various backgrounds, etc. which offer a challenging benchmark evaluation.

In summary, the contributions of this paper are: (1) we present TopoTag, a topological-based fiducial marker system and detection algorithm; (2) we demonstrate that TopoTag achieves the best performance in various metrics including detection accuracy, localization jitter and accuracy, etc. while at the same time supports occlusion, noise and flexible shapes; (3) we show that it's possible in tag design to use full bits for ID encoding without sacrificing detection accuracy, thus achieving scalability; and (4) we collect a large dataset of various tags, involving in-plane and out-of-plane rotation, image blur, different distances and various backgrounds, etc.

The remainder of the paper is organized as follows: In Section \ref{sec:related-work}, we discuss related work in different marker patterns. We introduce the TopoTag design and detection algorithm in Section \ref{sec:TopoTag-Design} and Section \ref{sec:TopoTag-Detection} respectively.
Dataset and experimentation are discussed in Section \ref{sec:Results}. Section \ref{sec:Conclusions} is devoted to the conclusions.

\section{Related Work} \label{sec:related-work}
\begin{figure*}[tp]
\captionsetup[subfigure]{labelformat=empty,font=scriptsize}
\centering
\subfloat[CCC \cite{Gatrell1991}]{\includegraphics[height=0.115\textwidth]{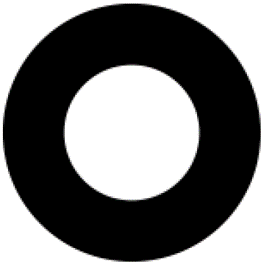}}
\hspace{0.02in}
\subfloat[Cho et al. \cite{Cho1998}]{\includegraphics[height=0.12\textwidth]{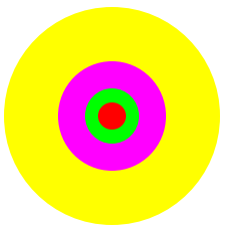}}
\hspace{0.02in}
\subfloat[Knyaz et al. \cite{Knyaz1998}]{\includegraphics[height=0.115\textwidth]{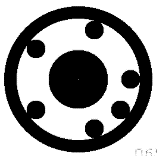}}
\hspace{0.02in}
\subfloat[InterSense \cite{Naimark2002}]{\includegraphics[height=0.115\textwidth]{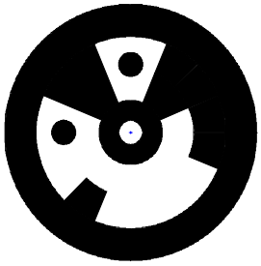}}
\hspace{0.02in}
\subfloat[FourierTag \cite{Sattar2007}\cite{Xu2011}]{\includegraphics[height=0.12\textwidth]{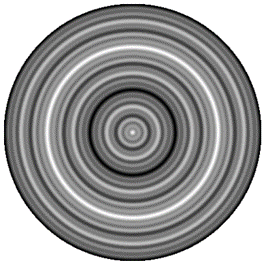}}
\hspace{0.02in}
\subfloat[RuneTag \cite{Bergamasco2011}\cite{Bergamasco2016}]{\includegraphics[height=0.115\textwidth]{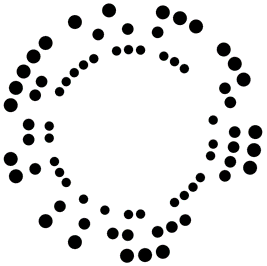}}
\hspace{0.02in}
\subfloat[CCTag \cite{Calvet2016}\cite{Calvet2012}]{\includegraphics[height=0.115\textwidth]{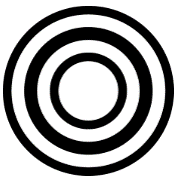}}
\hspace{0.02in}
\subfloat[Pi-Tag \cite{Bergamasco2013}]{\includegraphics[height=0.115\textwidth]{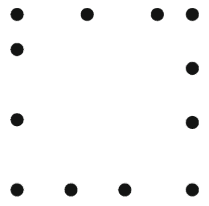}}
\hspace{0.02in}
\subfloat[Prasad et al. \cite{Prasad2015}]{\includegraphics[height=0.1155\textwidth]{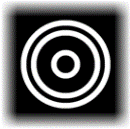}}
\hspace{0.03in}
\subfloat[Matrix \cite{Rekimoto1998}]{\includegraphics[height=0.115\textwidth]{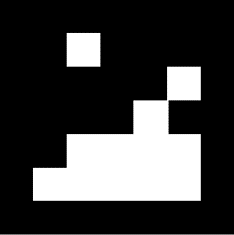}}
\hspace{0.02in}
\subfloat[ARToolKit \cite{Kato1999}]{\includegraphics[height=0.115\textwidth]{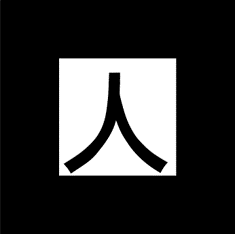}}
\hspace{0.02in}
\subfloat[CyberCode \cite{Rekimoto2000}]{\includegraphics[height=0.115\textwidth,width=0.1\textwidth]{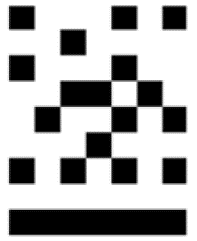}}
\hspace{0.02in}
\subfloat[VisualCode \cite{Rohs2004}]{\includegraphics[height=0.115\textwidth]{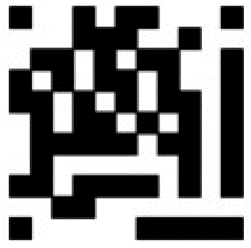}}
\hspace{0.02in}
\subfloat[ARToolKitPlus \cite{Wagner2007}]{\includegraphics[height=0.115\textwidth,width=0.115\textwidth]{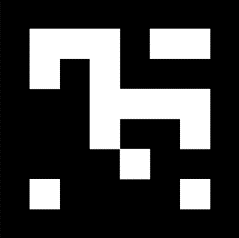}}
\hspace{0.03in}
\subfloat[binARyID \cite{Flohr2007}]{\includegraphics[height=0.115\textwidth]{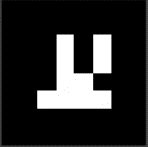}}
\hspace{0.02in}
\subfloat[Tateno et al. \cite{Tateno2007}]{\includegraphics[height=0.115\textwidth]{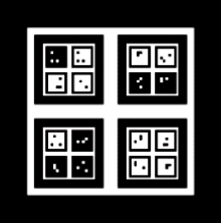}}
\hspace{0.02in}
\subfloat[SIFTTag \cite{Schweiger2009}]{\includegraphics[height=0.115\textwidth]{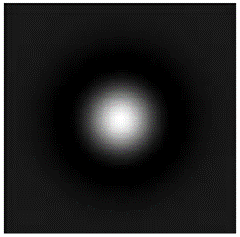}}
\hspace{0.02in}
\subfloat[ARTag \cite{Fiala2005}]{\includegraphics[height=0.115\textwidth]{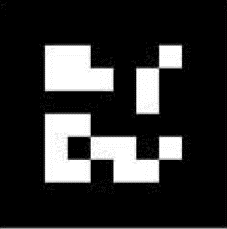}}
\hspace{0.02in}
\subfloat[AprilTag \cite{Goyal2011}\cite{Wang2016}]{\includegraphics[height=0.115\textwidth]{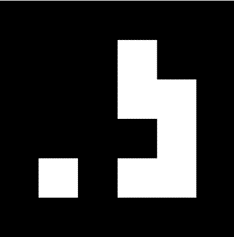}}
\hspace{0.02in}
\subfloat[ArUco \cite{Garrido-Jurado2015}\cite{Garrido2014}\cite{Romero-Ramirez2018}]{\includegraphics[height=0.115\textwidth]{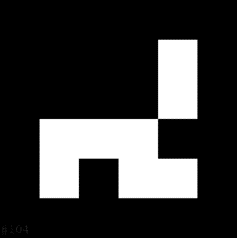}}
\hspace{0.02in}
\subfloat[ChromaTag \cite{Degol2017}]{\includegraphics[height=0.115\textwidth]{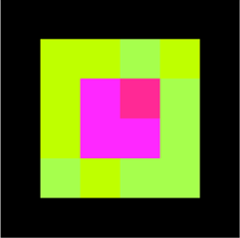}}
\hspace{0.02in}
\subfloat[D-touch \cite{Costanza2003_1}\cite{Costanza2003_2}]{\includegraphics[height=0.115\textwidth]{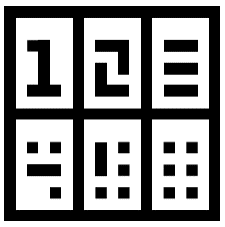}}
\hspace{0.03in}
\subfloat[reacTIVision \cite{Bencina2005_1}\cite{Bencina2005_2}\cite{Kaltenbrunner2007}]{\includegraphics[height=0.115\textwidth]{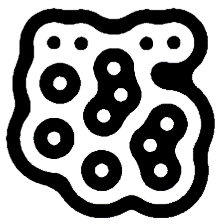}}
\hspace{0.03in}
\subfloat[BullsEye \cite{Klokmose2014}]{\includegraphics[height=0.115\textwidth]{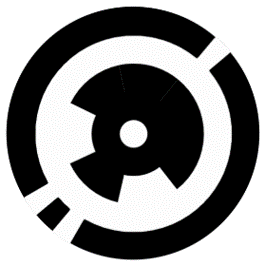}}
\caption{Existing fiducial marker systems.}
\label{fig:tags}
\end{figure*}

\autoref{fig:tags} shows many different fiducial marker systems discussed in this section.

\textbf{Circular patterns.} \quad Among the earliest work, Gatrell et al. \cite{Gatrell1991} propose to use concentric contrasting circle (CCC) for fiducial marker design. It's further enhanced in \cite{Cho1998} by adding colors and multiple scales. In \cite{Knyaz1998, Naimark2002}, dedicated data rings are added to the marker design for rich identification. Sattar et al. \cite{Sattar2007} and Xu et al. \cite{Xu2011} propose FourierTag with a frequency image as the signature. In RuneTag \cite{Bergamasco2011, Bergamasco2016} and Pi-Tag \cite{Bergamasco2013}, they propose using rings of dots to improve robustness to occlusion and provide more points for pose estimation. CCTag \cite{Calvet2012, Calvet2016} and followed work by Prasad et al. \cite{Prasad2015} use multiple rings to increase robustness to blur and ring width for encoding.
\Blue{Circular patterns, e.g. RuneTag, provide the state-of-the-art for most identities. However, the tracking distance is usually limited due to their requirement of finding enough confident ellipses. In comparison, TopoTag can provide even more identities while at the same time offering much larger tracking range.}

\textbf{Square patterns.} \quad To be easily localized, most fiducial systems are designed to contain a thick square border. Matrix \cite{Rekimoto1998}, CyberCode \cite{Rekimoto2000} and VisualCode \cite{Rohs2004} are the first and simplest proposals. ARToolkit \cite{Kato1999} is well known and widely used in many augmented reality applications. It includes a pattern in their internal region for identification via image correlation. ARTag \cite{Fiala2005} and ARToolkitPlus \cite{Wagner2007} improve the recognition technique with a binary coded pattern. In addition, they are designed with an error correction mechanism to increase robustness. BinARyID \cite{Flohr2007} proposes a method to generate markers that attempt to avoid rotation ambiguities. Schweiger et al. \cite{Schweiger2009} propose using SIFT and SURF filters that are specifically designed for SIFT and SURF detectors. Tateno et al. \cite{Tateno2007} propose using nested markers to improve performance under different distances. Several works investigate using multiple fiducial markers in a checkerboard to improve camera calibration \cite{Atcheson2010} and reduce the perspective ambiguity by further adding color \cite{Neto2010}. AprilTag \cite{Goyal2011, Wang2016} is a faster and more robust reimplementation of ARTag. Garrido-Jurado et al. \cite{Garrido2014, Garrido-Jurado2015, Romero-Ramirez2018} propose ArUco using mixed integer programming to generate markers. ChromaTag \cite{Degol2017} uses color over AprilTag to improve marker detection speed.
\Blue{Square patterns are most popular among practical applications due to this technique's detection robustness and large tracking range. However, some encoding bits must be reserved to handle rotation ambiguities and incorporate Hamming distance strategy. In contrast, TopoTag can provide much richer identities by encoding full bits while at the same time achieving the state-of-the-art robustness and tracking range. Moreover, unlike square markers using four corner points for pose estimation (which is the minimum number for unambiguous pose estimation \cite{Owen2002}), TopoTag offers better pose estimation utilizing all vertices of tag bits. It's worth noting that [10] shows the possibility of reducing rotation ambiguities, increases rich identities by adding color information and achieves better pose accuracy by using more inner corners. However, it still needs to reserve some bits for error detection and correction. In comparison, TopoTag offers even richer identities without using color due to the unique baseline node design and can utilize more feature correspondences for better pose estimation.}

\textbf{Topological patterns.} \quad D-touch \cite{Costanza2003_1, Costanza2003_2} is the earliest work to use topological patterns in tag design. Marker detection is based on the region adjacency tree information. D-touch employs a single topology for all markers in the set and does not provide a specific method for computing location and orientation. ReacTIVision \cite{Bencina2005_1, Bencina2005_2, Kaltenbrunner2007} improves over D-touch and provides unique identities purely with the topological structure by building a left heavy depth sequence of the region adjacency graph. BullsEye \cite{Klokmose2014}, which is specially optimized for GPU, consists of a central white dot surrounded by a solid black ring and one or more data rings again surrounded by a solid white ring inside a black ring with three white studs.
\Blue{Topological patterns demonstrate the ability to improve robustness using topological information. However, they (including ReacTIVision and BullsEye) can only recover 2D location and orientation due to the lack of sufficient matched feature points. In comparison, TopoTag offers accurate 3D pose estimation and  state-of-the-art robustness at the same time.}

\textbf{Machine learning.} \quad Claus et. al \cite{Claus2004, Claus2005} use trained classifiers to improve detection in cases of insufficient illumination and blurring caused by fast camera movement. Randomized forests are also used to learn and detect planar objects \cite{Fua2006, Ozuysal2010}.
\Blue{Machine learning methods show the potential to detect natural objects. However, in practice, these algorithms do not achieve detection accuracies on par with detection algorithms specifically designed for marker detection \cite{Degol2017}. In contrast, TopoTag achieves the state-of-the-art detection accuracy over machine learning and other previous types of patterns.}

\section{TopoTag Design} \label{sec:TopoTag-Design}
\begin{figure}[tp]
\centering
\subfloat{\includegraphics[width=0.48\textwidth]{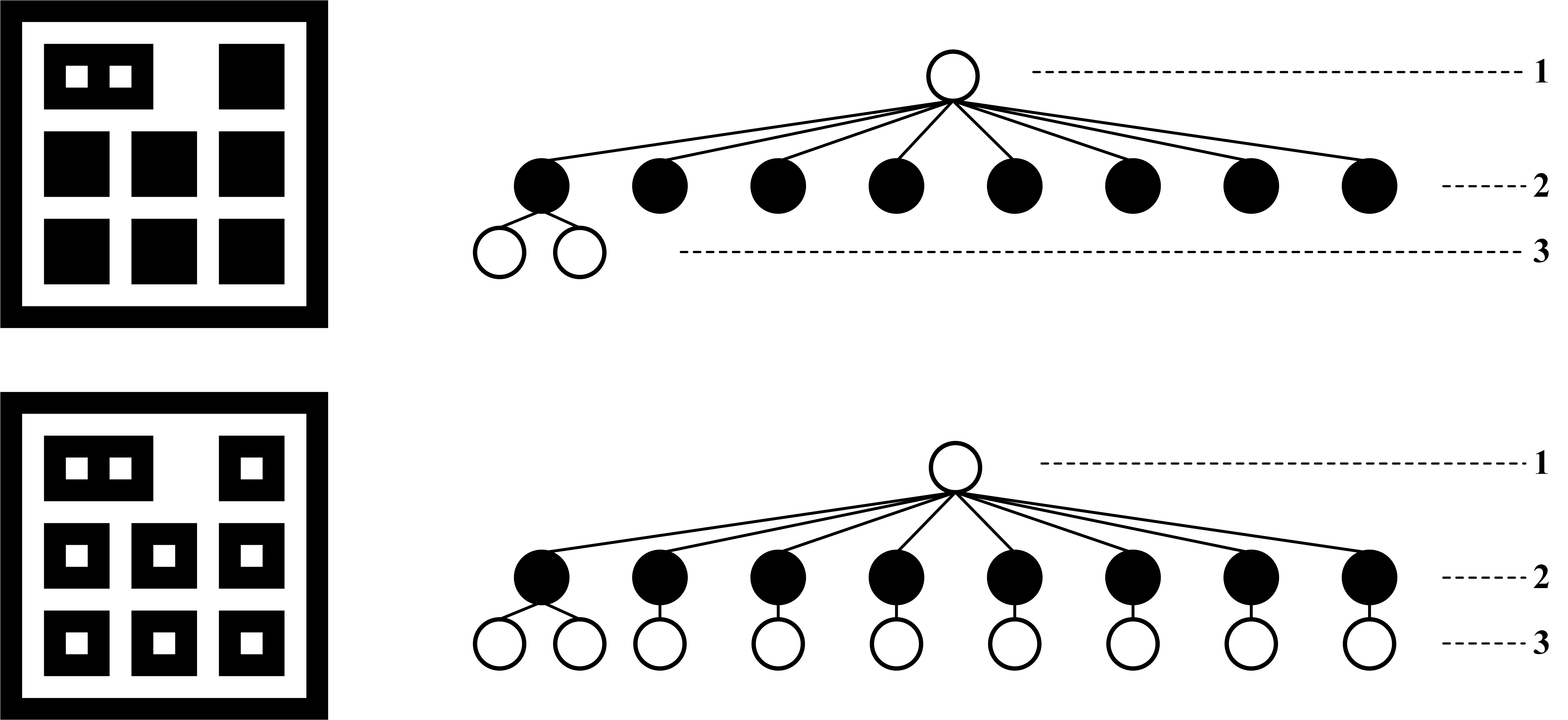}}
\caption{Topological tree of two TopoTags. \Blue{Each node in the topological tree denotes one TopoTag connected component (starting from the inner biggest white connected component). Except the two white nodes inside the baseline node, all leaf nodes are used for identify encoding. The identity encodings for these two markers are $0000000 = 0$ and $1111111 = 127$ respectively.}}
\label{fig:tag-tree}
\end{figure}

TopoTag utilizes topological structure information in tag design. This method has been validated with proven increases in robustness across illumination variation and a reduction in false detection \cite{Costanza2003_1}. Existing fiducial marker systems, especially with square patterns, sacrifice tag encoding bits to handle  rotation ambiguities during decoding \cite{Neto2010}. Additional bits will also be reserved for incorporating Hamming distance strategy in order to improve false positive rejection. Strong robustness with topological design helps by saving tag bits for encoding identities. To avoid rotation ambiguities, TopoTag introduces \textit{baseline node} in its topological structure. The baseline node is specially designed to be different from other nodes in the tag. TopoTag uses a black node with two white children nodes inside as the baseline node and other black nodes, with at most one white child node, as \textit{normal nodes}. Note that, baseline node can be defined with other forms. For example, it can be defined with three or more white children nodes for different needs. Baseline node defines the search starting position of the whole tag, thus avoiding checking rotation ambiguities. All normal nodes are used for identity encoding with $0$ denoting no child node and $1$ otherwise. The identity encoding for the two markers shown in \autoref{fig:tag-tree} is $0000000 = 0$ and $1111111 = 127$ respectively.

For pose estimation, instead of using only four border points in previous square systems \cite{Fiala2005, Wagner2007, Goyal2011, Garrido2014, Garrido-Jurado2015, Wang2016, Degol2017} which is the minimum number required, TopoTag offers more point correspondences resulting in more accurate pose estimation. Baseline node (more specifically its two children nodes) and all normal nodes are all employed as feature points, thus achieving a better pose estimation.

Note that, as TopoTag design is based on topological information, there is no restriction for the shapes used in the tag. Both internal and external shapes can be customized as long as the desired topological structure is preserved. \autoref{fig:topotag} shows three different design samples of TopoTag. For easy searching and model simplicity, in current TopoTag design, we place all internal nodes uniformly spaced and compacted into a $n \times n$ squared shape.

\section{TopoTag Detection} \label{sec:TopoTag-Detection}
\begin{figure*}[tp]
\centering
\subfloat[Input image.]{\includegraphics[width=0.24\textwidth]{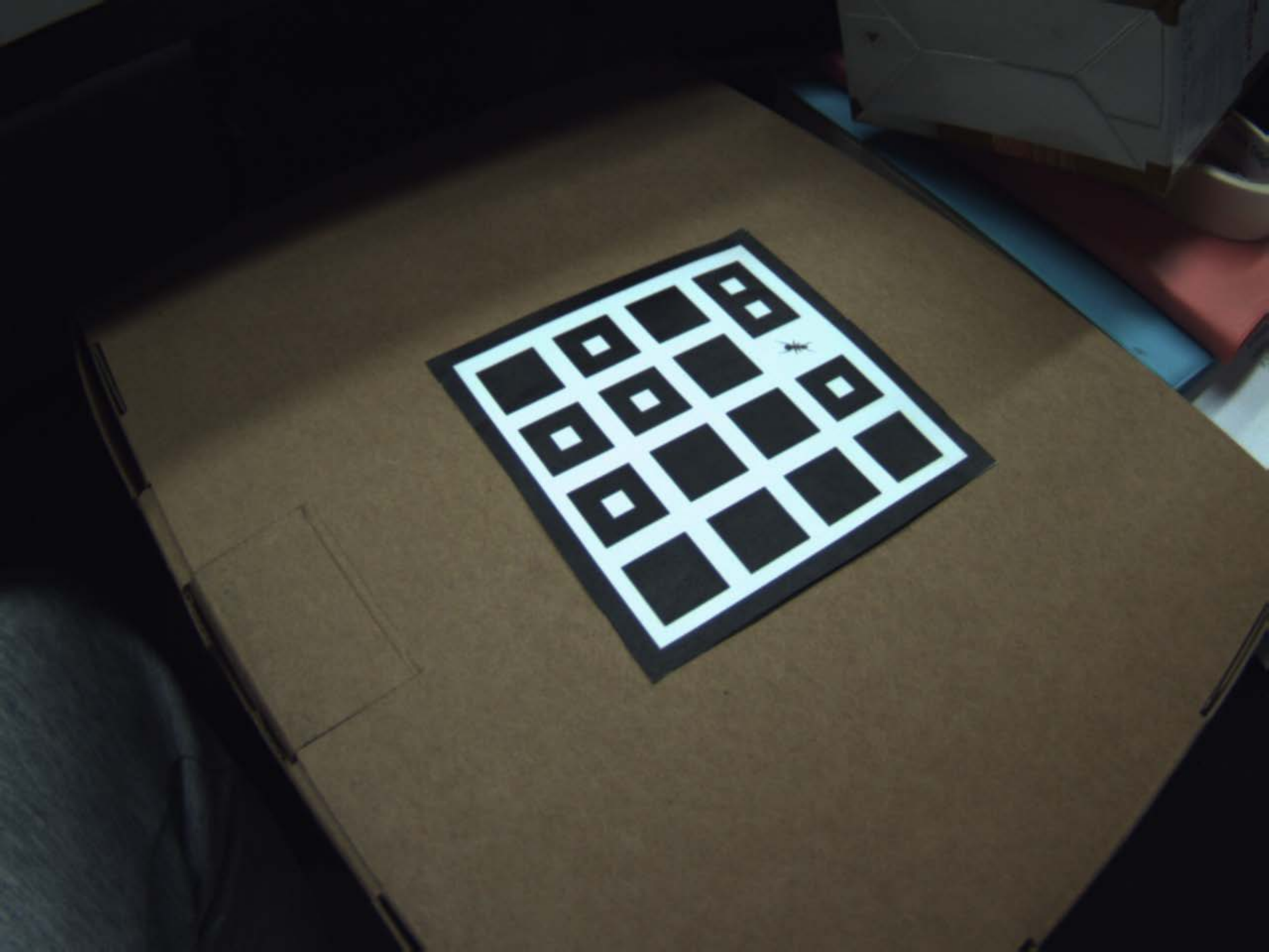}}
\hspace{0.01in}
\subfloat[Threshold map. \label{fig:alg-steps-2}]{\includegraphics[width=0.24\textwidth]{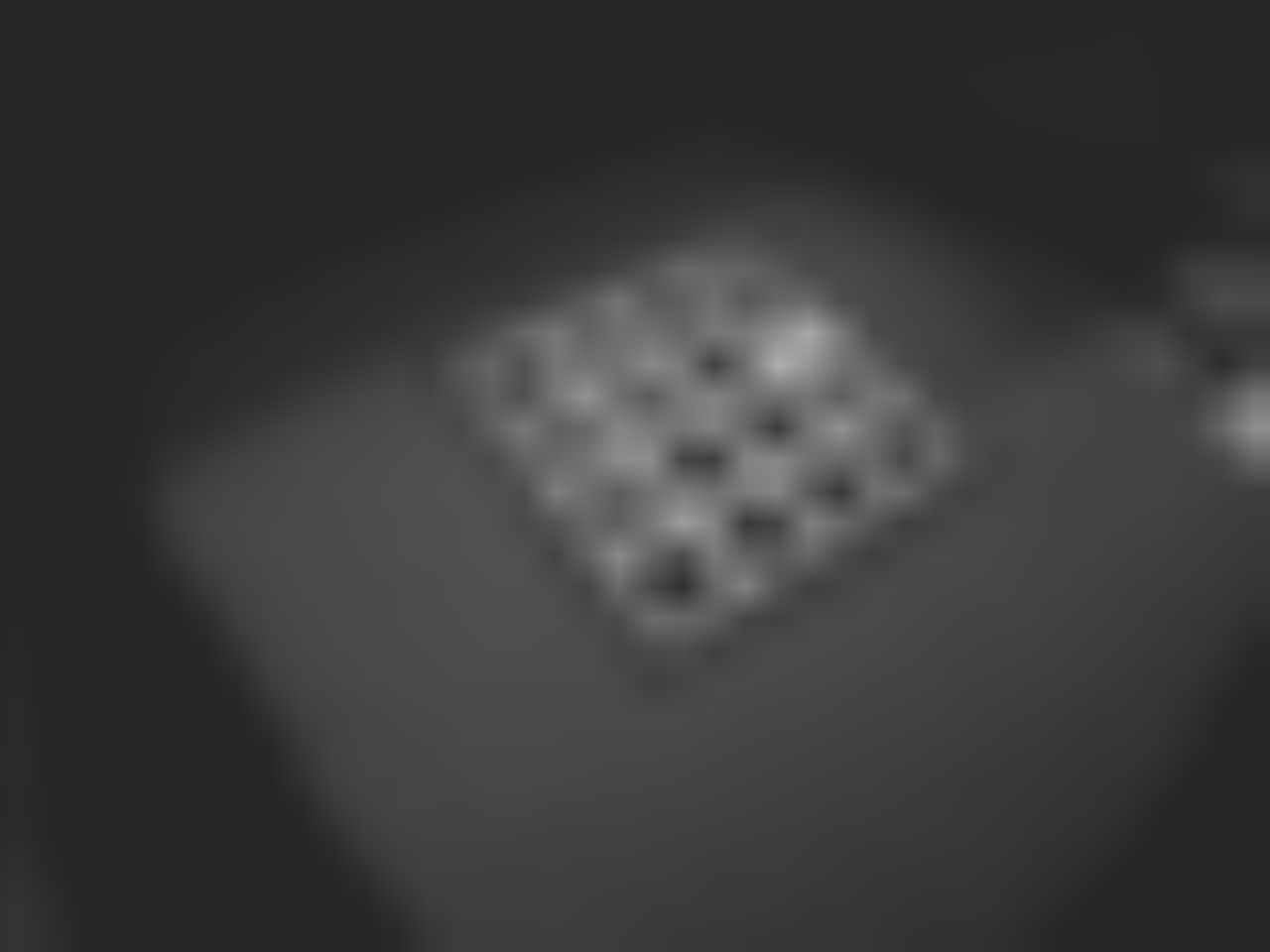}}
\hspace{0.01in}
\subfloat[Binarization. \label{fig:alg-steps-3}]{\includegraphics[width=0.24\textwidth]{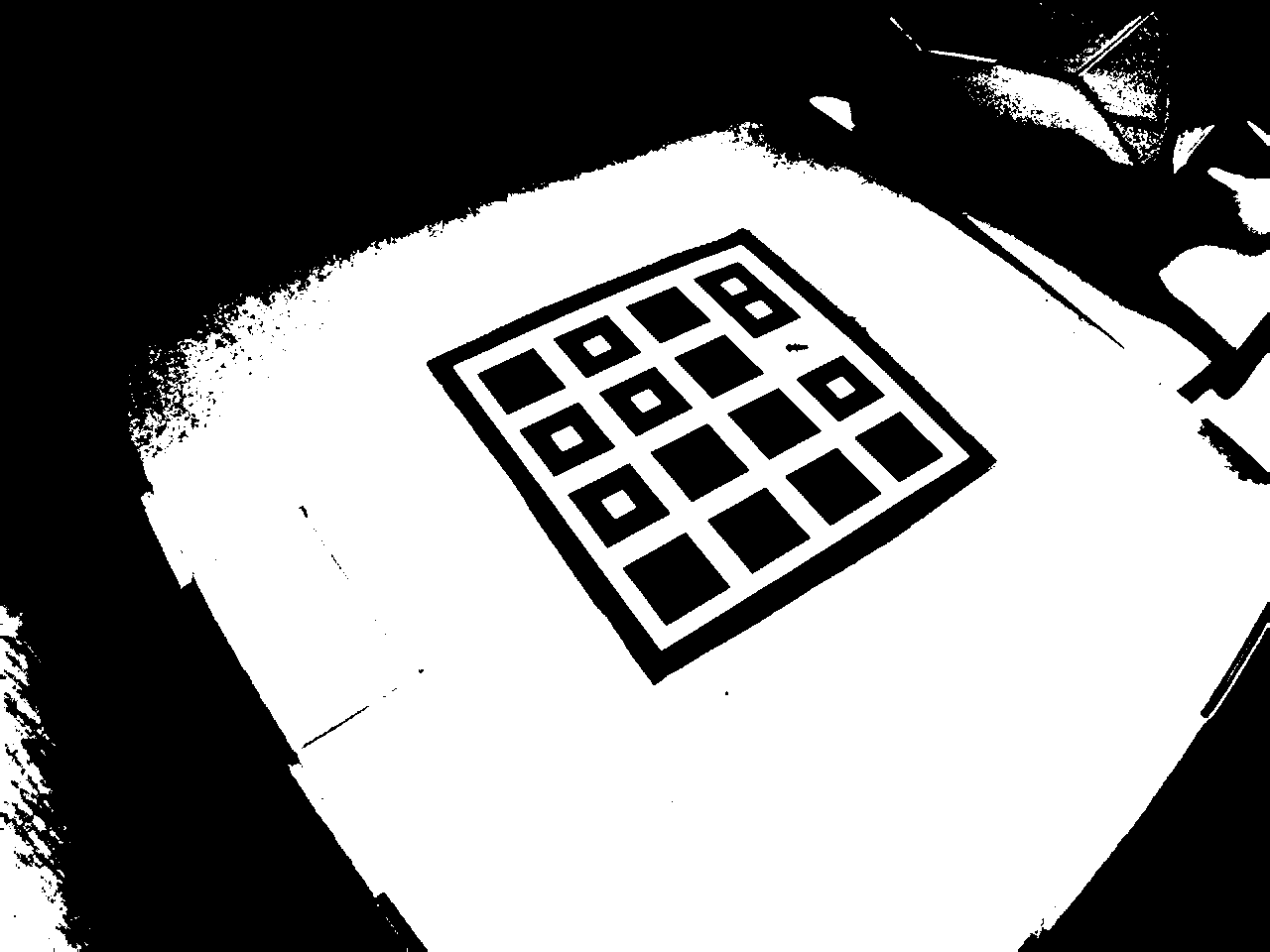}}
\hspace{0.01in}
\subfloat[Topological filtering. \label{fig:alg-steps-4}]{\includegraphics[width=0.24\textwidth]{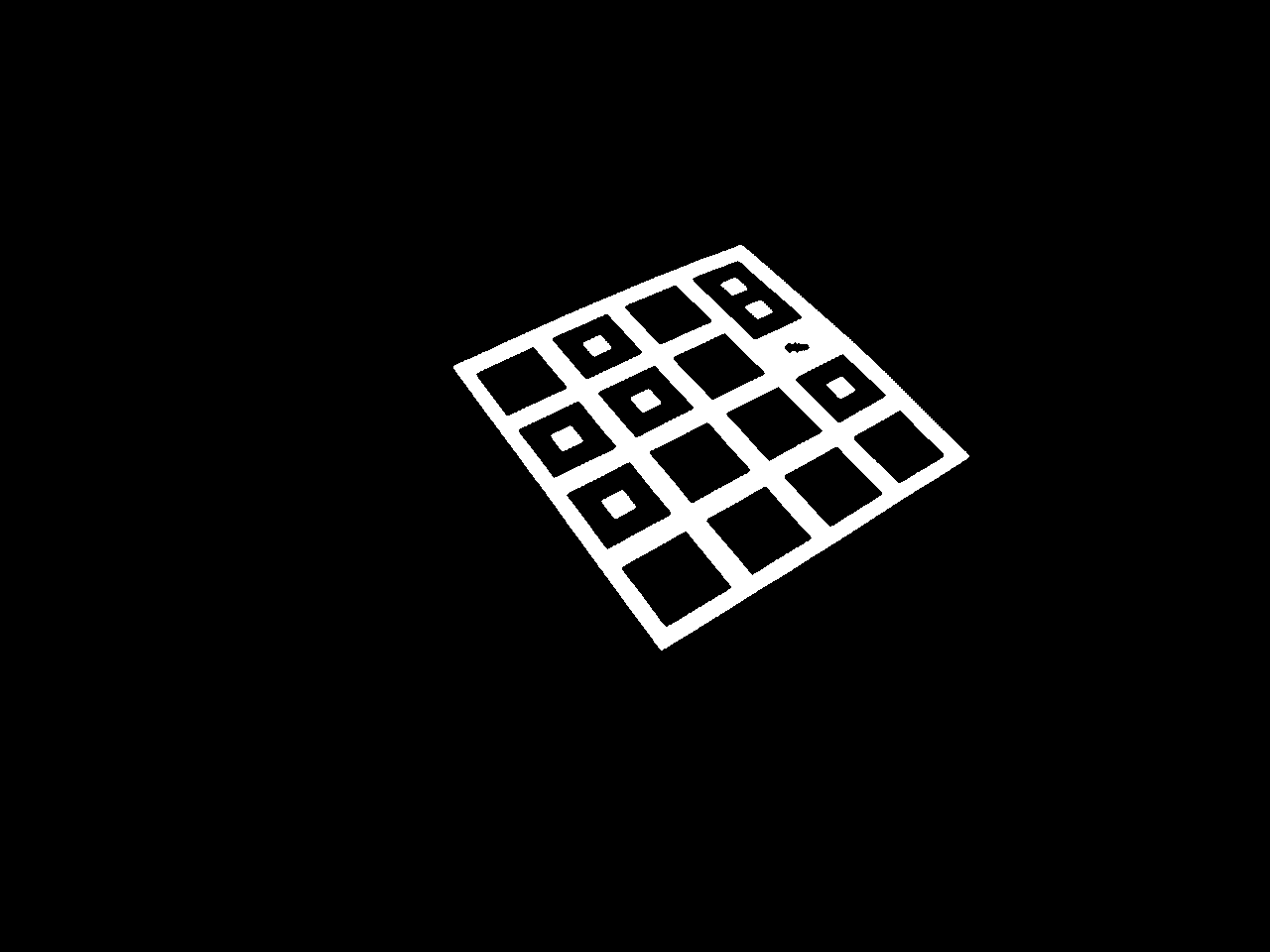}}
\hspace{0.01in}
\subfloat[Error correction. \label{fig:alg-steps-5}]{\includegraphics[width=0.24\textwidth]{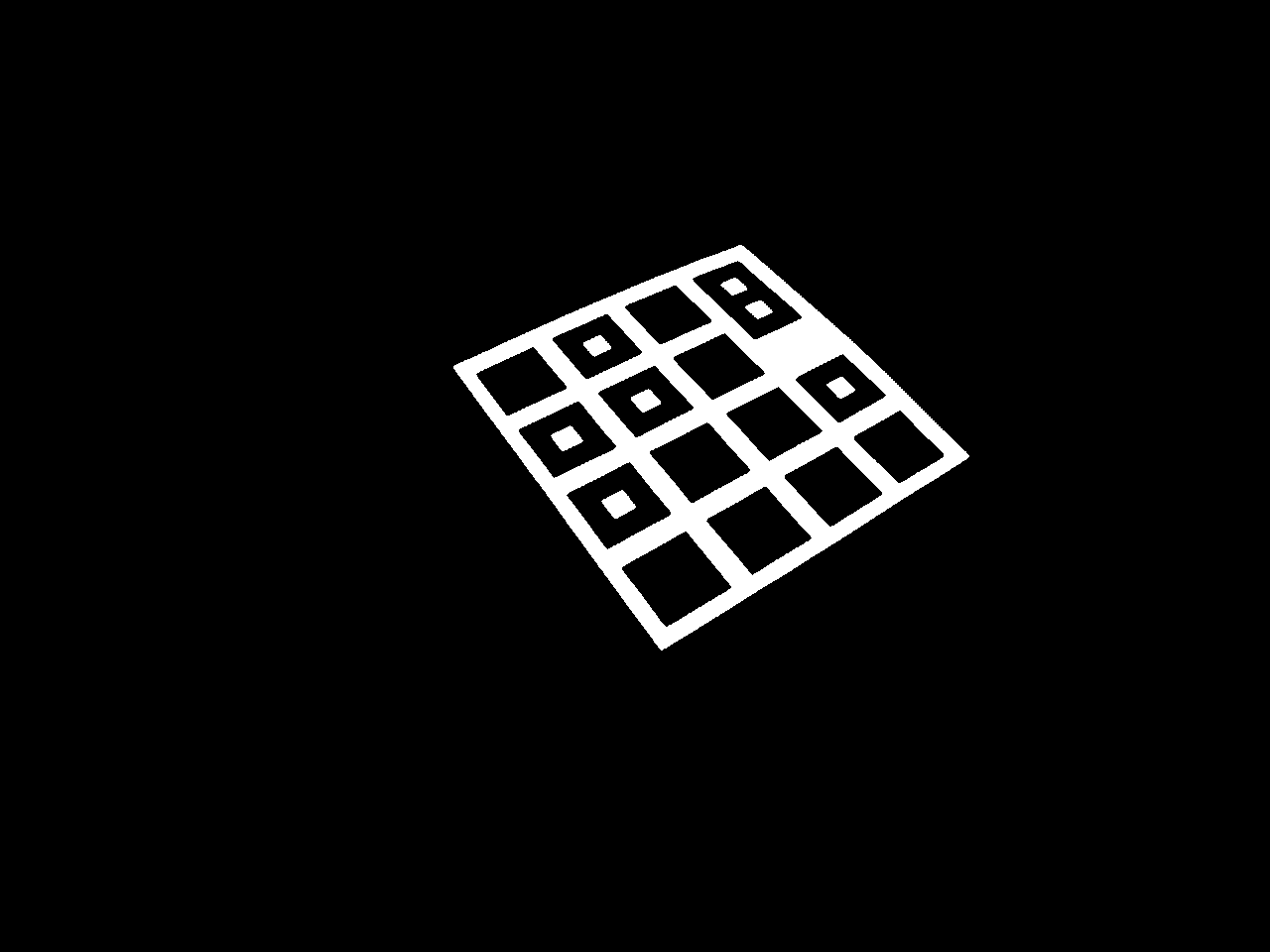}}
\hspace{0.01in}
\subfloat[Decoding. \label{fig:alg-steps-6}]{\includegraphics[width=0.24\textwidth]{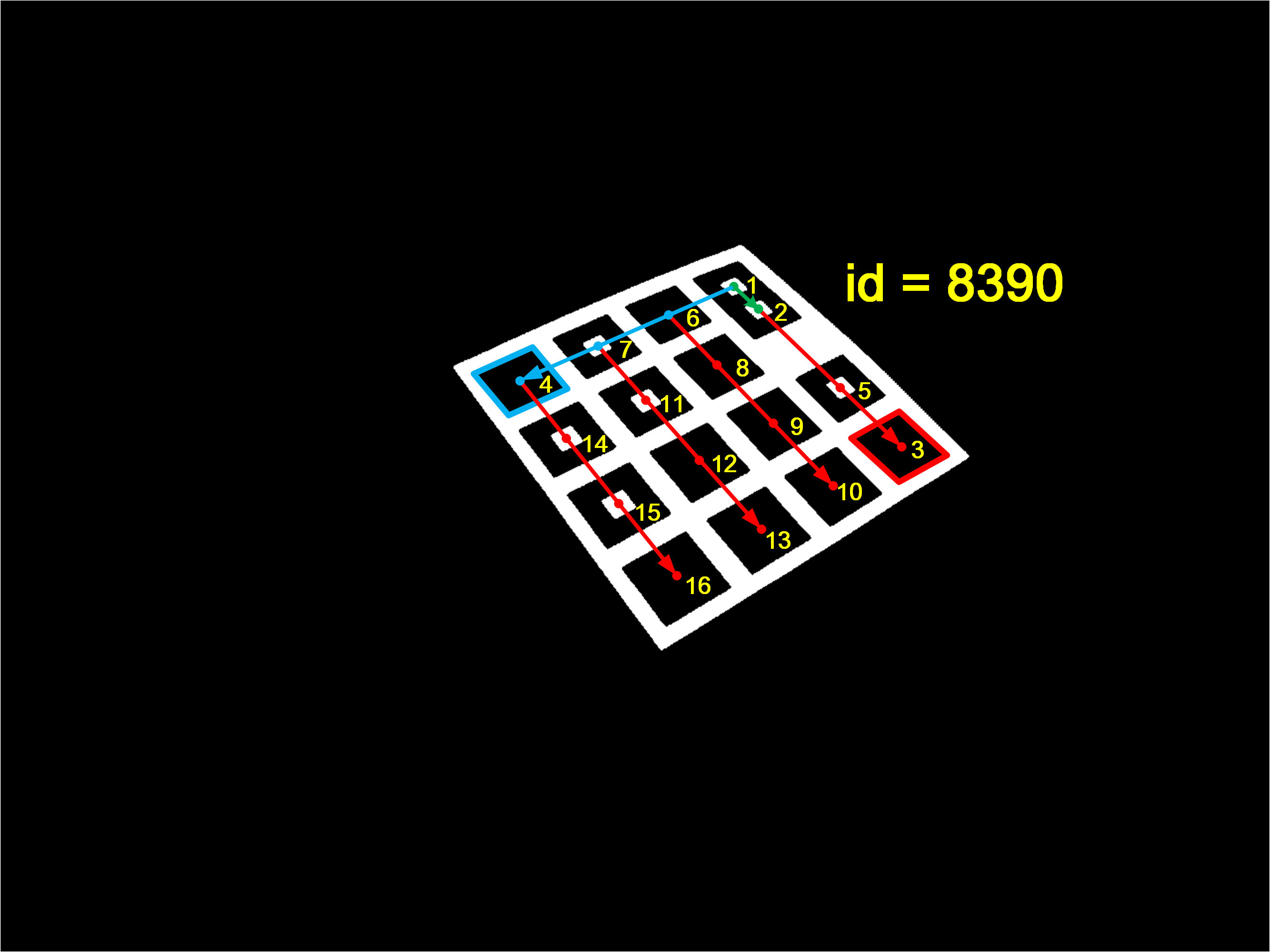}}
\hspace{0.01in}
\subfloat[Vertex estimation. \label{fig:alg-steps-7}]{\includegraphics[width=0.24\textwidth]{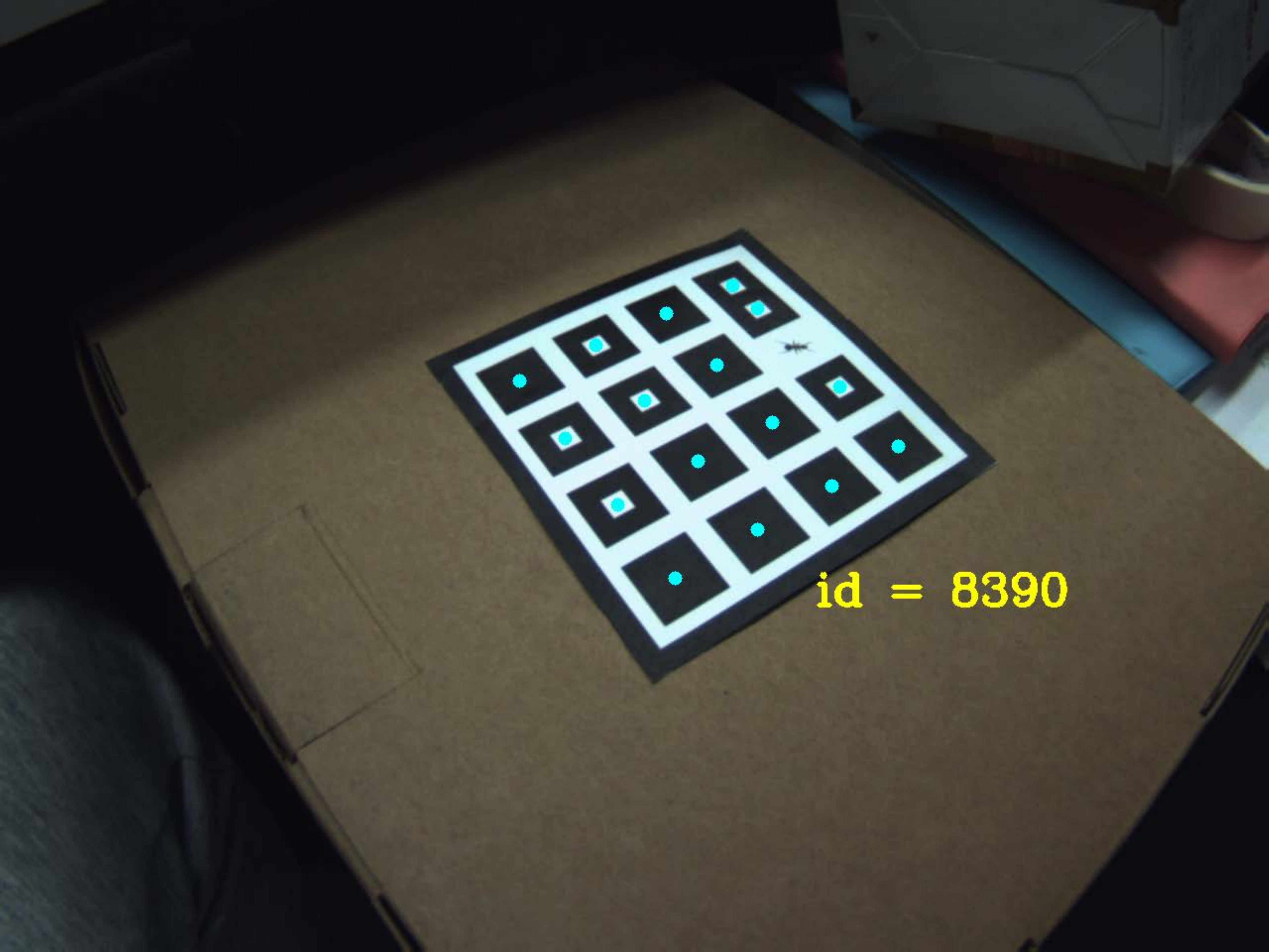}}
\hspace{0.01in}
\subfloat[Pose estimation. \label{fig:alg-steps-8}]{\includegraphics[width=0.24\textwidth]{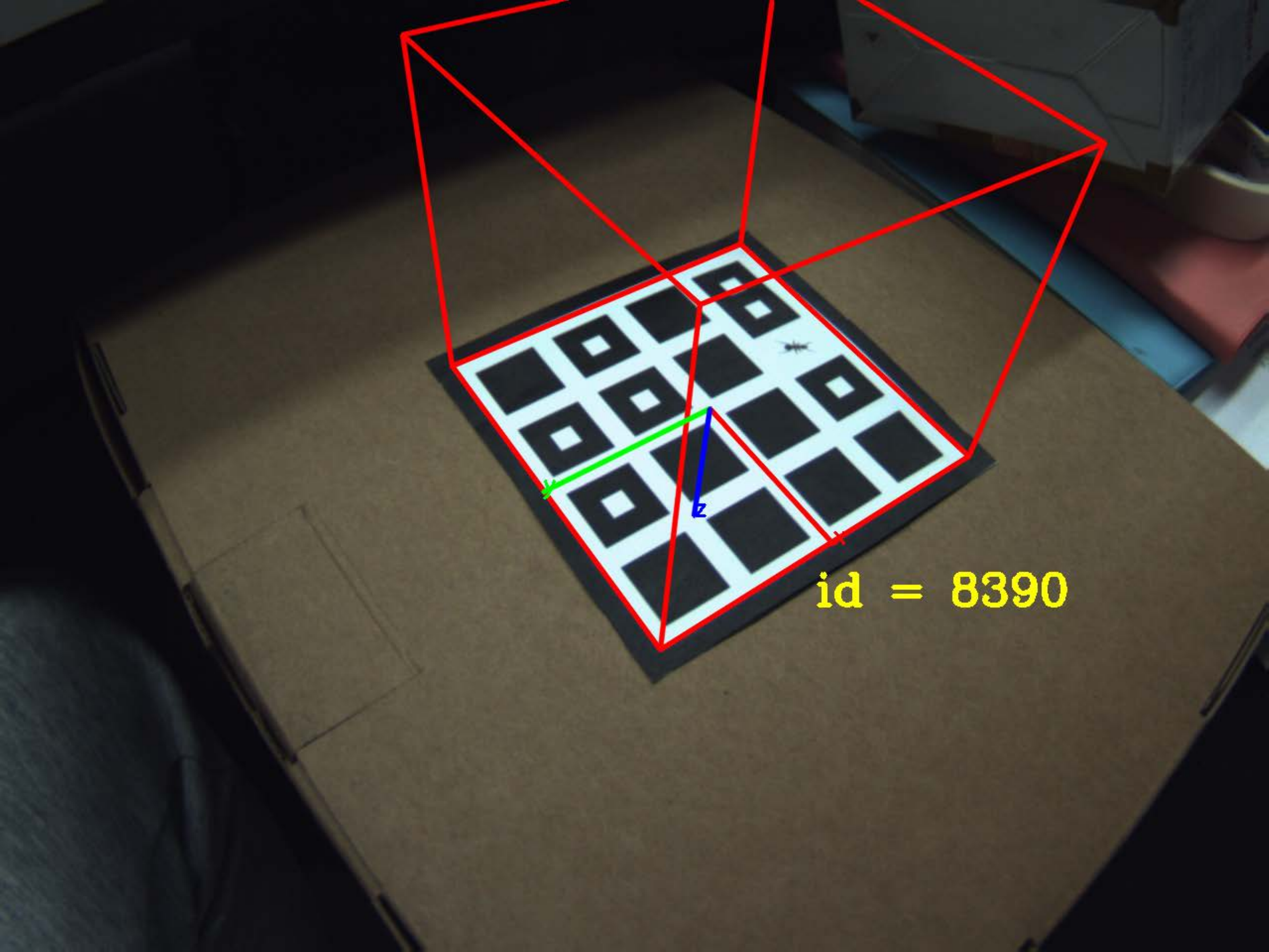}}
\caption{Main steps of TopoTag detection. (Best viewed in color)}
\label{fig:alg-steps}
\end{figure*}

\autoref{fig:alg-steps} outlines main steps of TopoTag detection. Topological information is extensively used for 2D marker detection, and further corresponding geometrical information for ID decoding. 3D pose estimation is achieved by taking advantage of all TopoTag vertices.

\subsection{2D Marker Detection}
\textbf{Threshold map estimation.} \quad Similar to the idea of adaptive thresholding, we estimate the threshold for each pixel by analyzing its neighboring pixels. The analysis can be conducted on the original image, however, in order to deal with the image noise and blur in real applications, analyzing a downsampled image (scalar $s_1$) is more accurate, which also brings speed benefits. Any pixel will be set to $\alpha$ if its value is less than $\alpha$ to remove pixels that are too dark. Average values are computed on a local region (window size $w$) on the downsampled image.
To further handle the image noise, the downsampled average map can be further downsampled (scalar $s_2$). The final threshold map is achieved by upsampling the downsampled average map by $s_1 \times s_2$ using bilinear interpolation, see Fig. \autoref{fig:alg-steps-2}.

\textbf{Binarization.} \quad Binarization is achieved by comparing the input image with the threshold map. A minimum brightness ($\beta$) is set to filter regions that are too small (i.e. set to black if pixel value is less than $\beta$). See Fig. \autoref{fig:alg-steps-3} for an example of binarization result.

\textbf{Topological filtering.} \quad After the binarization, we build the topological tree of the connected binary regions. To find candidate tags, we search the tree based on two conditions: (1) the number of children nodes should be within $[ \zeta_{\min} - \tau,  \zeta_{\max} + \tau\ ]$, where $\zeta_{\min}$ is the number of nodes for tag ID $=0$ with all black leaves except the baseline node and $\zeta_{\max}$ for the tag with maximum ID with no black leaves, and $\tau$ is the tolerance level allowed; (2) max depth of the tree should be exactly 3. See \autoref{fig:tag-tree} for examples of the topological trees for both $\zeta_{\min}$ and $\zeta_{\max}$ cases of 9-bit TopoTags. Fig. \autoref{fig:alg-steps-4} shows the result after the topological filtering.

\textbf{Error correction.} \quad There are possible error nodes within the tag region due to noise or occlusion. Fig. \autoref{fig:alg-steps-4} shows an example of one error node close to the baseline node because of one ant sitting on the tag. To correct these error nodes, we  first compute the area of the baseline node and then filter out smaller nodes if their areas are less than $\theta_1\%$ of the baseline node area. Fig. \autoref{fig:alg-steps-5} shows the result after error correction.

\subsection{ID Decoding}
\begin{figure}[tp]
\Blue{
\centering
\subfloat{\includegraphics[width=0.24\textwidth]{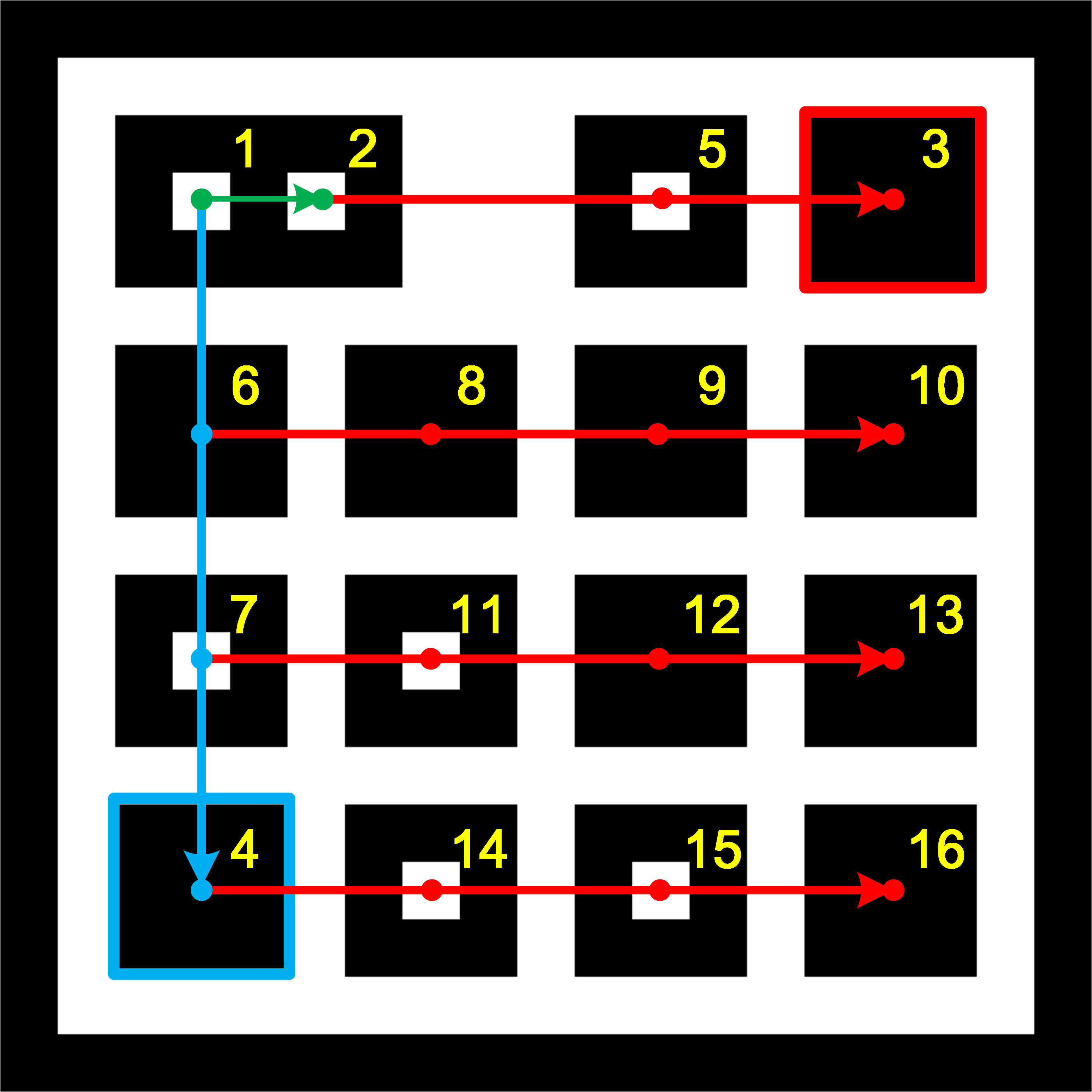}}
\caption{Vertex decoding order. (Best viewed in color)}
\label{fig:decode-order}
}
\end{figure}


To decode ID, we need to determine the node sequence and map it to a binary code string. Take a 16-bit TopoTag as an example, see \autoref{fig:decode-order} and Fig. \autoref{fig:alg-steps-6} of the sequence where we find for each node of the tag. To start, we first find the baseline node (including p1 and p2) and determine its search direction based on whether there are nodes along the direction with angel tolerance $\theta_2$, i.e. p1$\rightarrow$p2. Along the direction, we find the node with the largest distance, i.e. p3. For the remaining nodes, we first find the node with largest angle against the baseline direction p1$\rightarrow$p2 and then the largest distance along the direction, i.e. p4. p5 is determined along direction p1$\rightarrow$p3, p6 and p7 along p1$\rightarrow$p4. The remaining nodes are determined in order and in a similar way. After finding each node, we can simply map each node to 1 or 0 depending on whether it contains a white child node or not and then decode the tag based on the binary code string. For the example shown here and in \autoref{fig:alg-steps}, the binary code string is $10000011000110$, which is decoded with ID $=8390$.
\Blue{It's worth noting that ID decoding is processed on the images after removing the perspective distortion in which lines will still be lines in images with no distortion to improve the robustness of direction searching.}

\subsection{3D Pose Estimation}
For each node, we estimate the vertex by computing the centroid on the original image of its supporting region. The supporting region can be the binary mask or its dilated version (with dilate size $\delta$). The centroid can be determined via image moments, i.e. $\{\bar{u}, \bar{v}\} = \{\frac{M_{10}}{M_{00}}, \frac{M_{01}}{M_{00}}\}$.

For pose estimation, the exact correspondence between the 2D image features and the features of the associated model is needed (feature correspondence). At least four points are needed to recover unambiguous pose estimation for planar tags \cite{Owen2002}. Unlike most of previous work using only four corner points, all TopoTag vertices of tag bits are used for a better pose estimation. \Blue{As reported in \cite{Collins:2014:IPP:2660592.2660598}, a larger number of feature correspondences consistently leads to lower error and better pose estimation to noise for various PnP methods. We refer the reader to \cite{Chen:2009:EAH:1507555.1507568} for a detailed analysis on the stability of homography estimation by 1st-order perturbation theory.} For 16-bit tag, 16 vertex correspondences are used, including two baseline white nodes and 14 normal black nodes. 6-DoF pose estimation is achieved by solving the PnP problem and Levenberg-Marquardt algorithms \cite{10.2307/2098941, Hartley:2003:MVG:861369} based on these feature correspondences.

\section{Results and Discussion} \label{sec:Results}
\textbf{Algorithm setup.} \quad Throughout the experiment, we use $s_1=4, s_2=8, w=5, \alpha=45, \beta=50$ for segmentation, $\tau=0, \theta_1=30, \theta_2=0.1\text{ rad}$ for decoding, $\delta=\max\{2, \floor[\big]{\frac{l}{10}}\}$ for vertex estimation, where $l$ is the short length of the binary mask region.

All of the experiments have been performed on a typical laptop PC equipped with an Intel Core i7-7700HQ processor (8 cores @2.8Ghz) and 8GB of RAM.

\subsection{Dataset}
The previous work, like \cite{Goyal2011, Bergamasco2011}, mainly focused on evaluating performance on synthetic images. Although some of the work evaluated parts of the performance on more realistic scenes, e.g. ARToolKitPlus \cite{Wagner2007} evaluates the speed on several handheld devices and AprilTag \cite{Goyal2011, Wang2016} evaluates false positive on LabelMe \cite{Russell2008} dataset which is designed for general object detection and recognition research, there is still no uniform dataset for fiducial marker evaluation. This makes it difficult to reproduce the result and compare with others. More recently, in ChromaTag \cite{Degol2017} work, they collected a dataset to compare their work with AprilTag \cite{Goyal2011}, CCTag \cite{Calvet2016}, and RuneTag \cite{Bergamasco2011}. However, different tags are placed side-by-side during their dataset collection, thus it is not ideal for comparison, especially when tags viewed from a large angle as different markers will have different distances and facing angles towards the camera.

\begin{figure}[tp]
\centering
\subfloat{\includegraphics[width=0.24\textwidth]{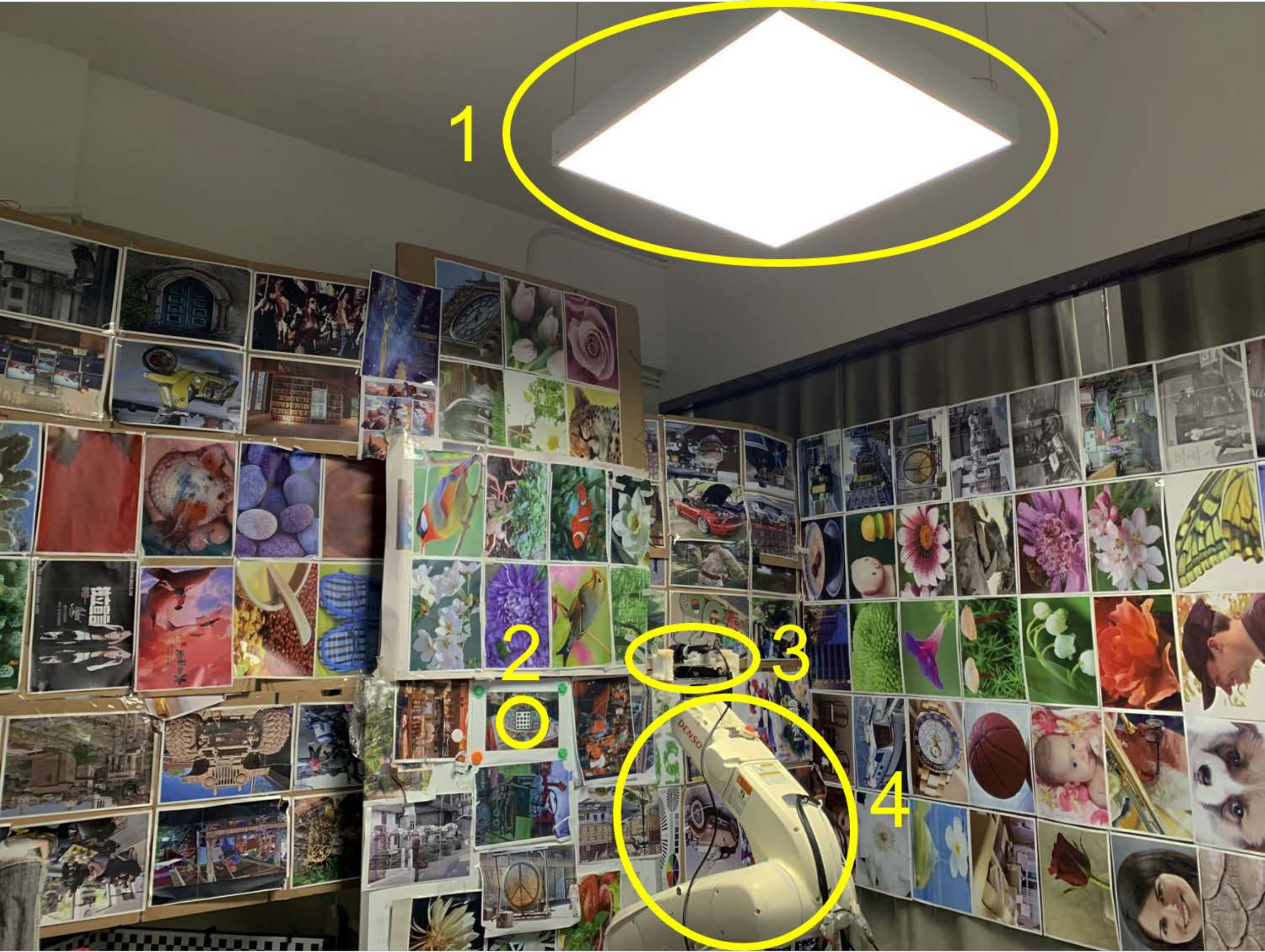}}
\hspace{0.01in}
\subfloat{\includegraphics[width=0.24\textwidth]{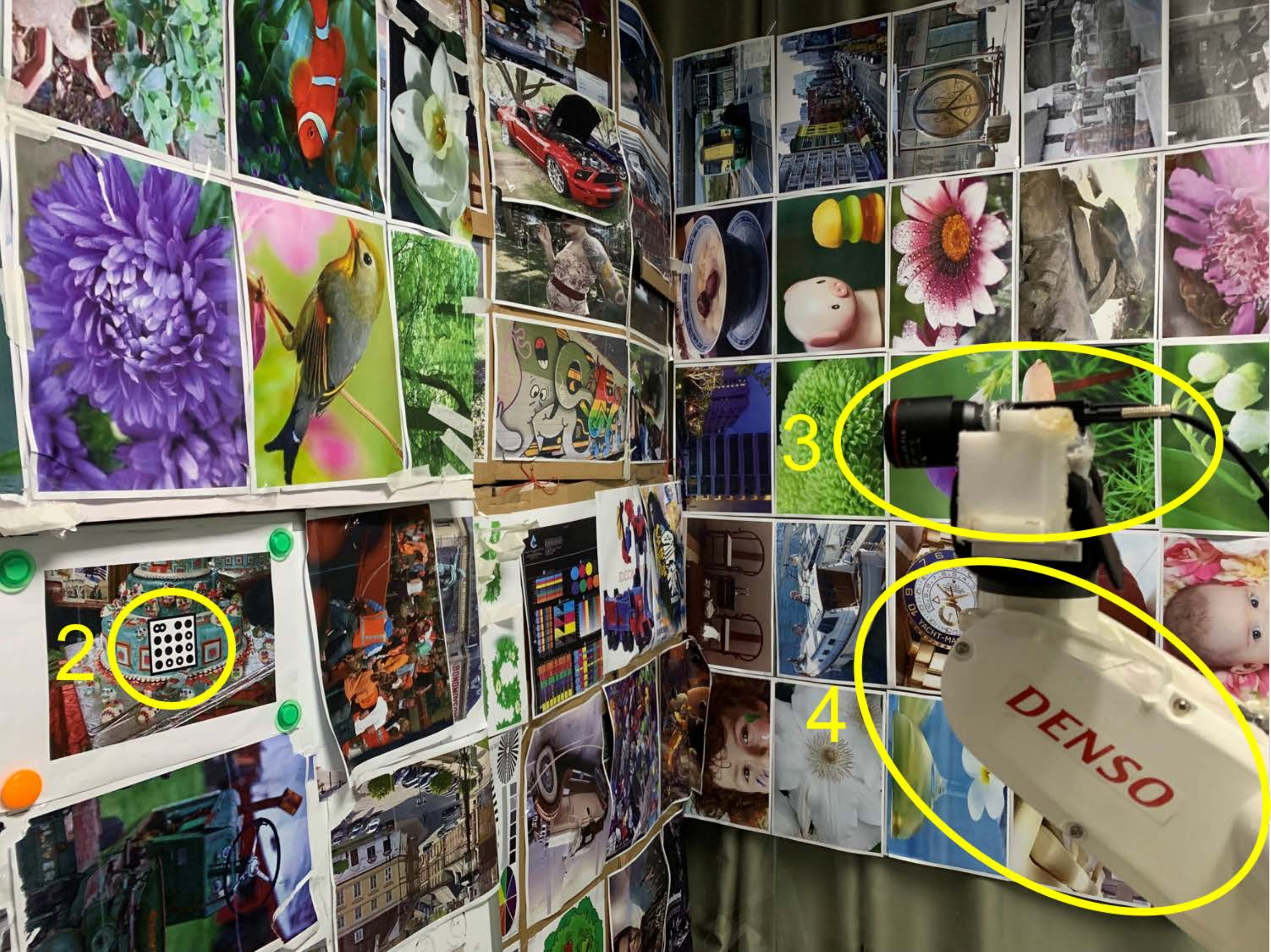}}
\caption{Dataset collection setup. We collect dataset by putting tags (label \#2) in a rich textured background of an indoor environment with fixed lighting (label \#1). The camera (label \#3) is fixed to a robot arm (label \#4) to ensure the same trajectories for different tags.}
\label{fig:test-setup}
\end{figure}

\begin{figure}[tp]
\centering
\subfloat{\includegraphics[width=0.155\textwidth]{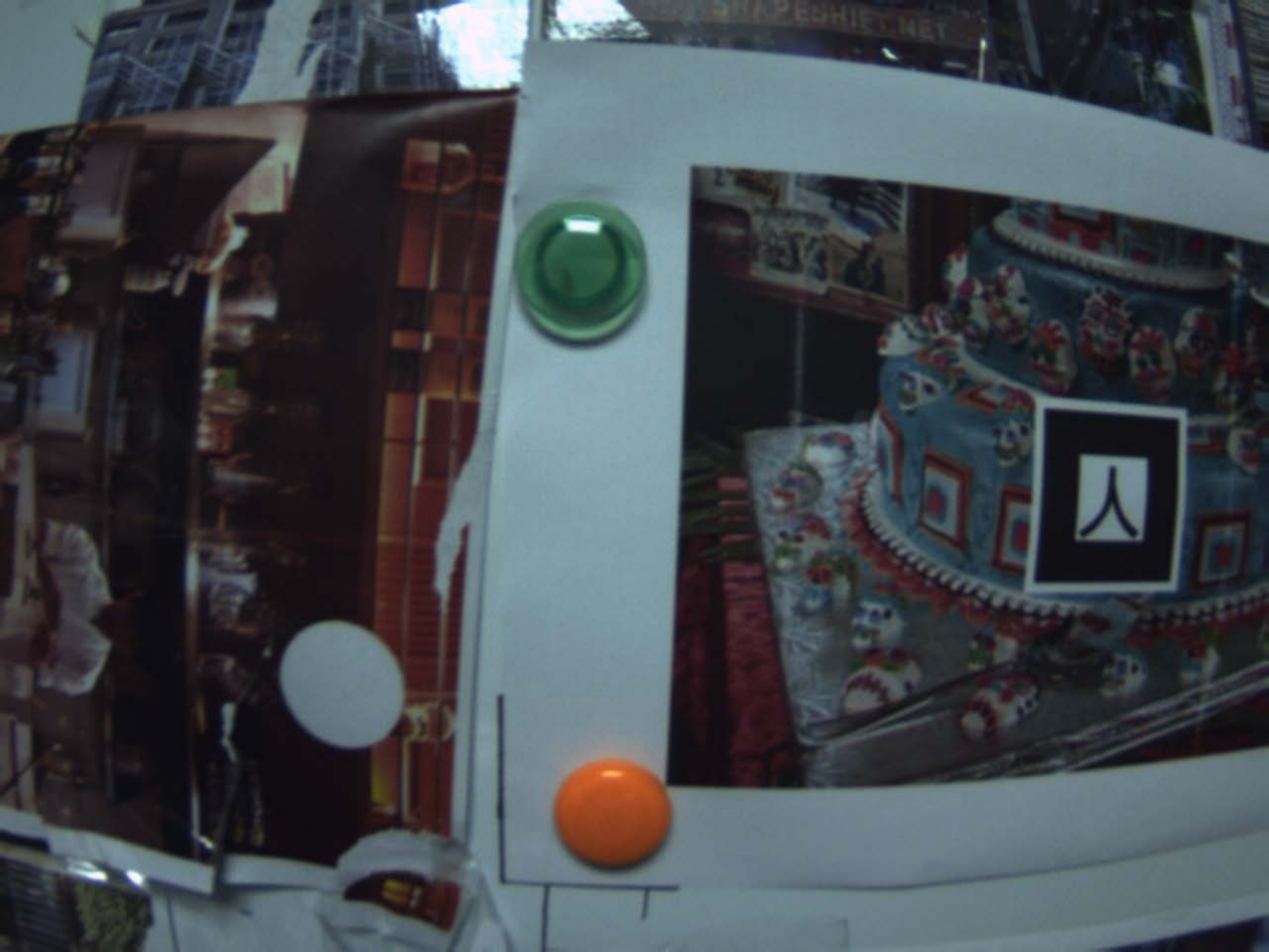}}
\hspace{0.01in}
\subfloat{\includegraphics[width=0.155\textwidth]{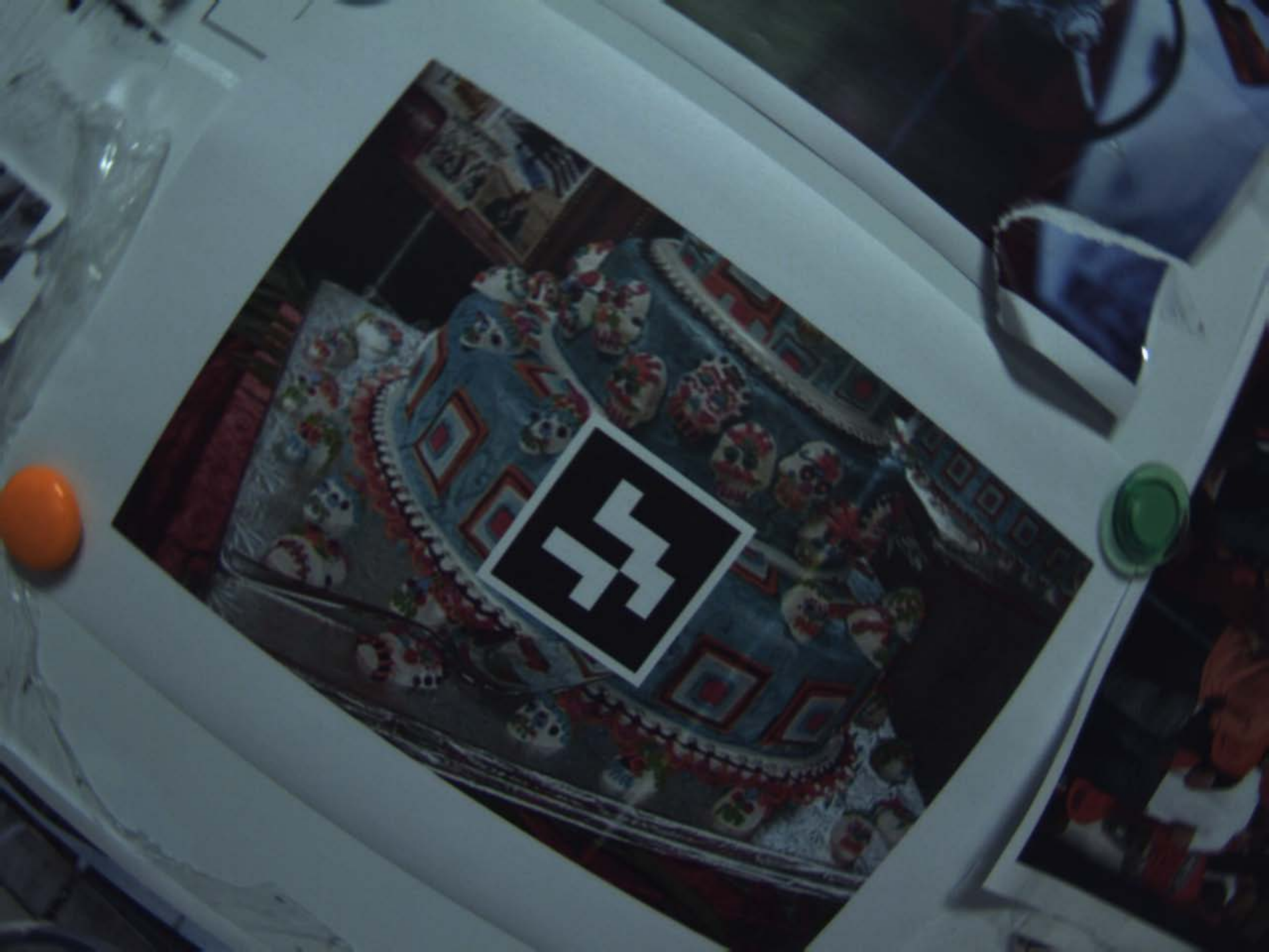}}
\hspace{0.01in}
\subfloat{\includegraphics[width=0.155\textwidth]{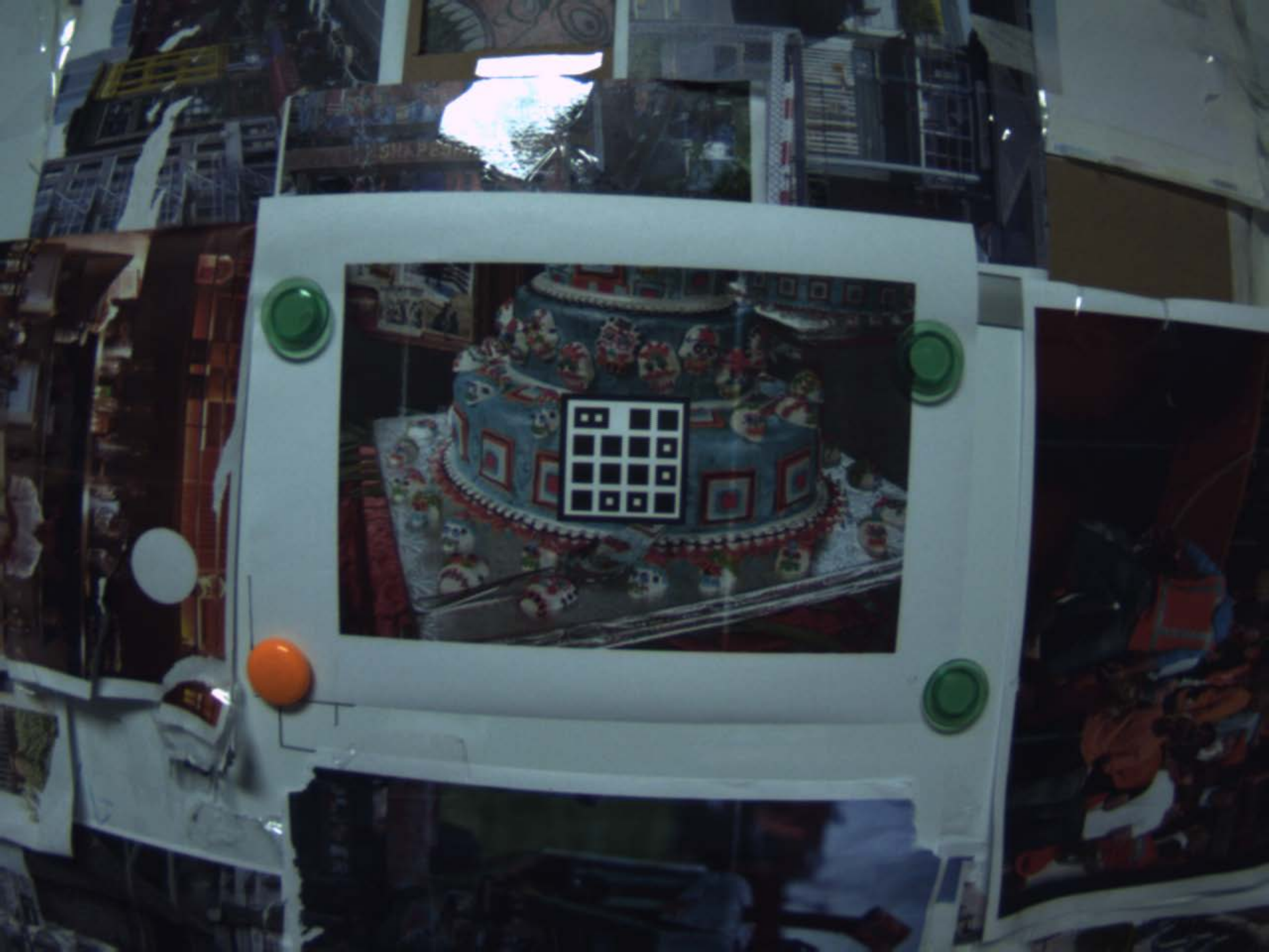}}
\caption{Sample images from the dataset. Images are from Seq \#1 (with ARToolKit), Seq \#2 (with AprilTag 25h9) and Seq \#3 (with TopoTag) from left to right respectively.}
\label{fig:samples}
\end{figure}

\begin{figure*}[tp]
\centering
\subfloat{\includegraphics[height=0.2475\textwidth]{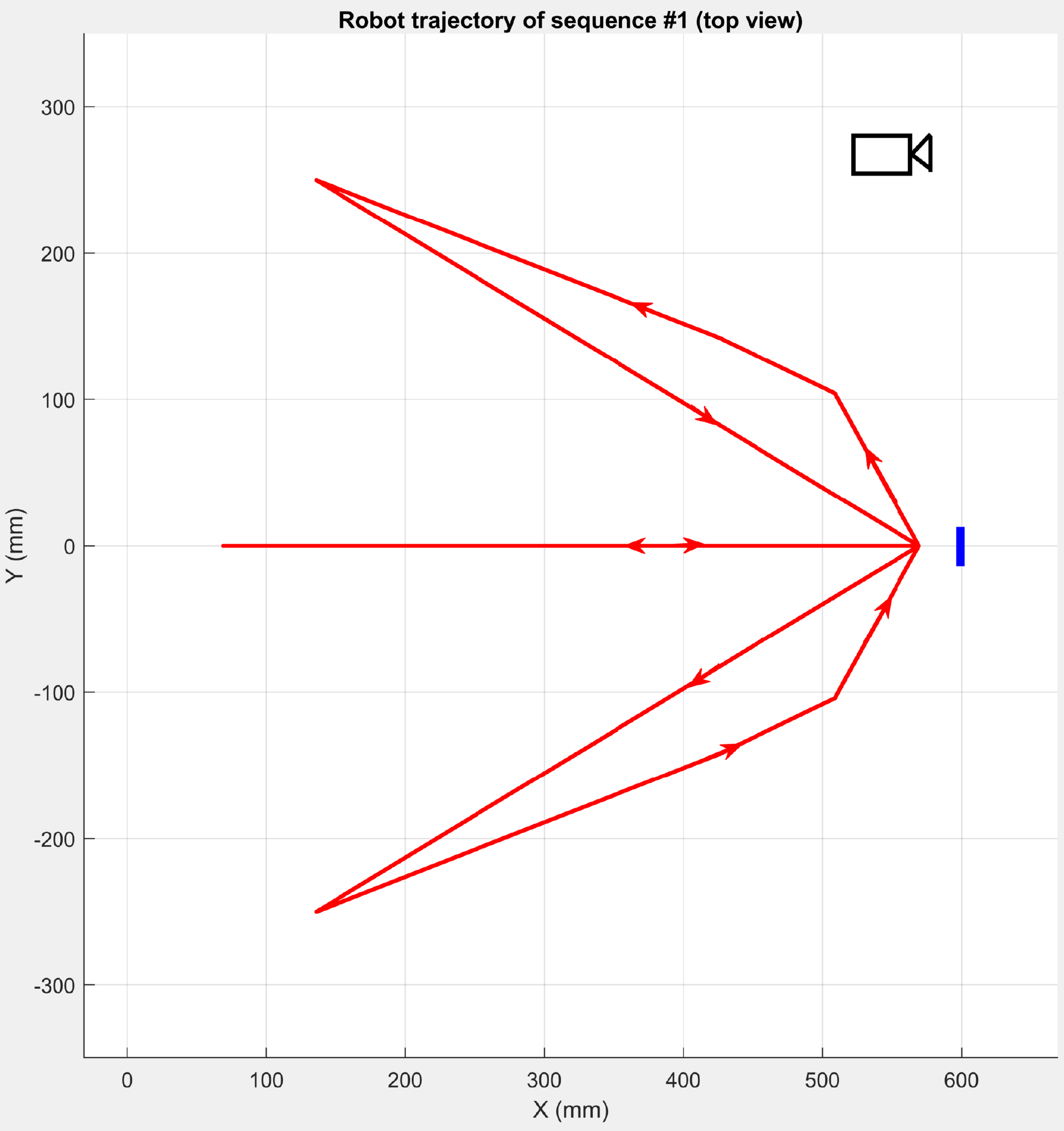}}
\hspace{0.01in}
\subfloat{\includegraphics[height=0.2475\textwidth]{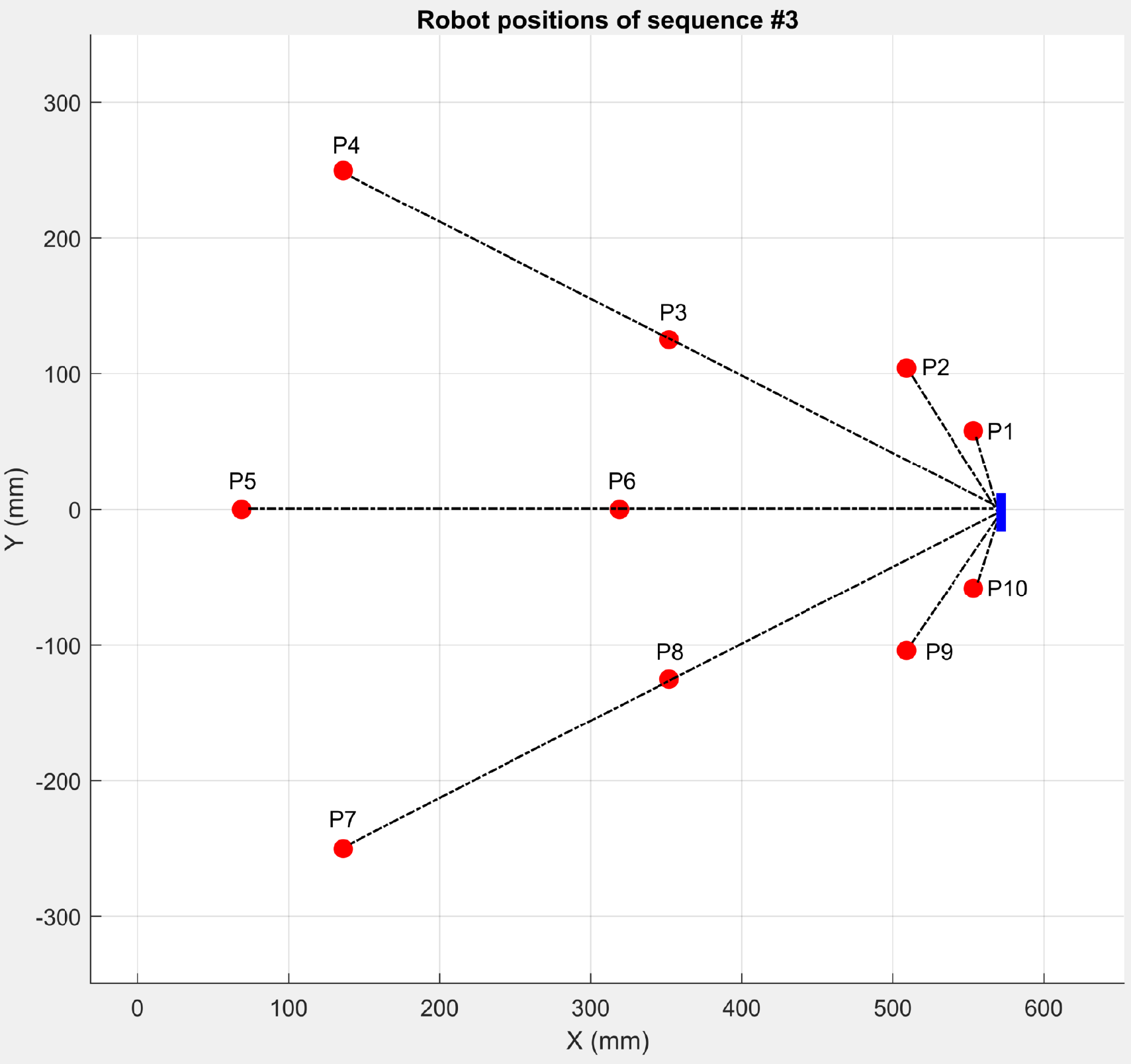}}
\hspace{0.01in}
\subfloat{\includegraphics[height=0.2475\textwidth]{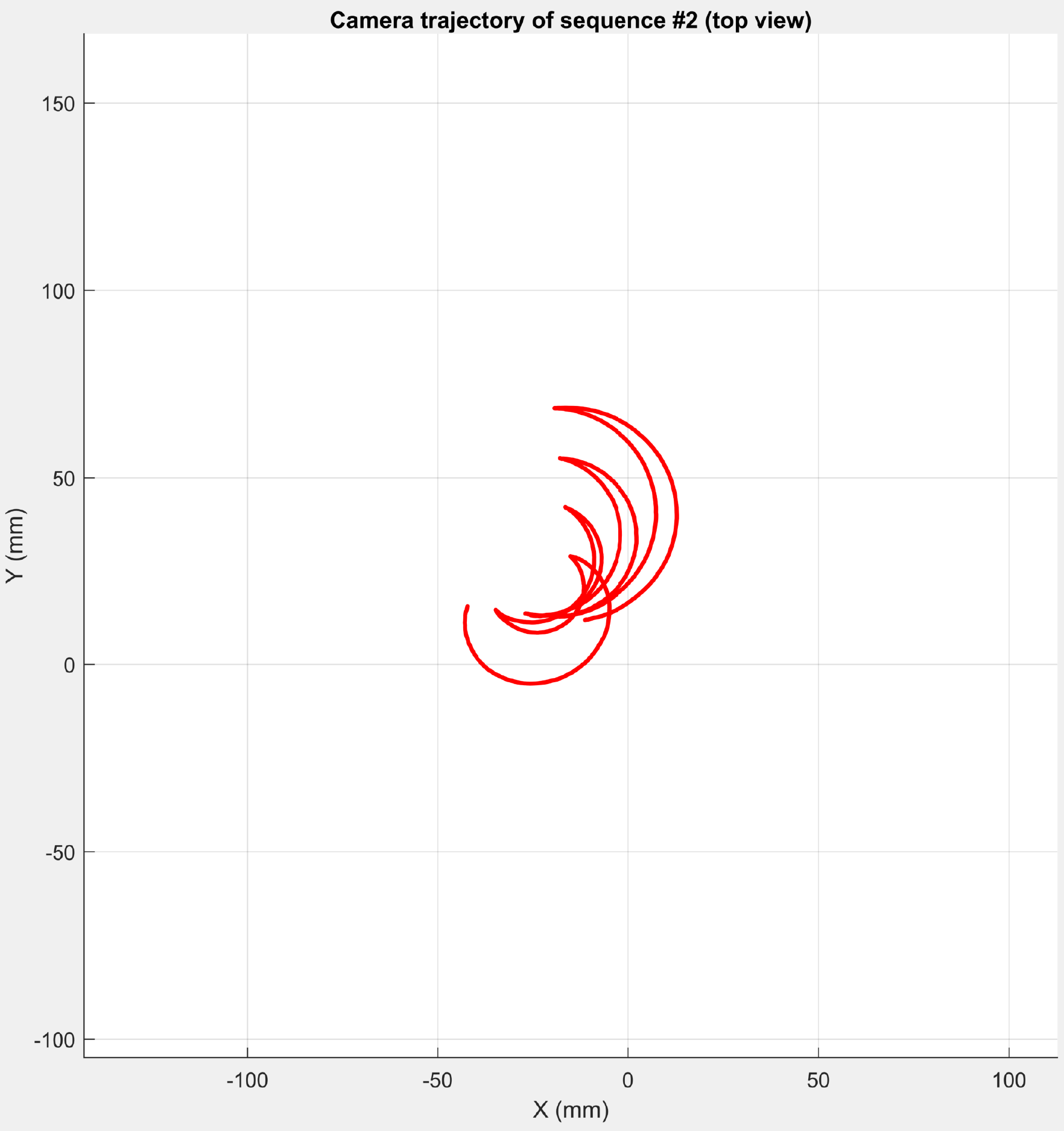}}
\hspace{0.01in}
\subfloat{\includegraphics[height=0.2475\textwidth]{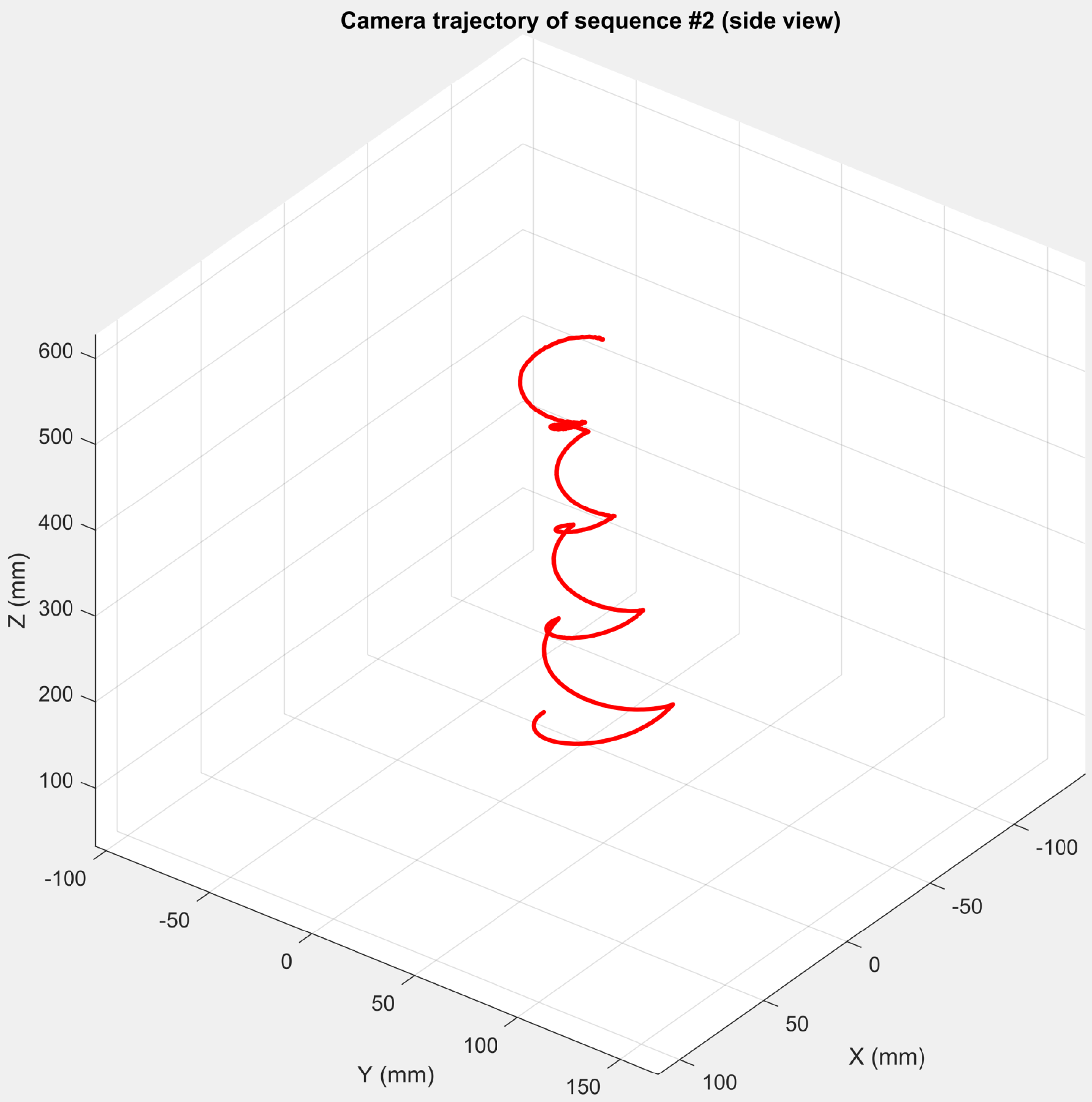}}
\caption{Robot arm trajectory/points in different sequences (1st image for Seq \#1, 2nd image for Seq \#3). Camera trajectory is shown for Seq \#2 for better visualization (3rd and 4th images). Tag position is shown in blue. (Best viewed in color)}
\label{fig:pose-viz}
\end{figure*}

In this work, we try to fill this gap by collecting a large dataset, including a total of 169,713 images, which include in-plane and out-of-plane rotations, image blur, various distances and cluttered backgrounds, etc. Please refer to the supplementary material for details of our dataset variations. We use an industrial camera with a global shutter that has 1280$\times$960 resolution streaming at 38.8 fps and 98\degree \ diagonal field of view. The exposure time is fixed at 10 ms. Using relatively long exposure guarantees sufficient brightness of the captured images, which at the same time introduces the image blur phenomenon for more challenging use cases (see the first image in \autoref{fig:samples} for an example). The camera is fixed to a robot arm\footnote{We use a robot arm from DENSO (VS-6556). Link: \url{https://www.denso-wave.com/en/robot/product/five-six/vs.html}} to ensure the same trajectories for different tags. \autoref{fig:test-setup} shows the dataset collection setup. Three sequences will be collected for each tag, and the trajectory for each sequence is shown in \autoref{fig:pose-viz}. In all the three sequences, the camera keeps facing the front as shown in the first image of \autoref{fig:pose-viz}. In Seq \#1, the camera moves along several lines at a constant speed, with different out-of-plane rotations for each line including 0\degree \ (i.e. camera faces the tag right ahead), 30\degree \ and 60\degree. In Seq \#2, the camera moves back and at the same time rotates in-plane within 0-180\degree \ at a constant speed back and forth. Note that, as we can only rotate around the end joint of the robot arm and there is an offset between the camera and arm, the camera's trajectory will not be an ideal half circle. In Seq \#3, the camera is placed at 10 fixed positions (P1$\rightarrow$P10). Besides 0\degree, 30\degree and 60\degree\ out-of-plane rotations as in Seq \#1, we further collect data with 75\degree \ (P1 and P10). In all three sequences, the background is filled with rich textured images to simulate more complex use scenarios.

We collect the dataset for TopoTag and previous tags including ARToolKit \cite{Kato1999}, ARToolKitPlus \cite{Wagner2007}, ArUco \cite{Garrido2014}, RuneTag \cite{Bergamasco2011}, ChromaTag \cite{Degol2017} and AprilTag \cite{Wang2016}. A 16-bit TopoTag is used throughout the experiment as it provides the most unique identities, see \autoref{tab:track-dist} for details. And, without loss of generality, the tag comes with square internal and external shapes (see the first image in \autoref{fig:topotag}). For systems with multiple tag families, we collect data for each tag family, including 16h3, 25h7, 36h12 for ArUco and 16h5, 25h7, 25h9, 36h9, 36h11 for AprilTag. For each tag family (including TopoTag), we randomly select one ID for evaluation.
In our experiment, we randomly selected ID $=1$ for ARToolKit, 262 for ARToolKitPlus, [104, 90, 136] for ArUco's [16h3, 25h7, 36h12], 107 for RuneTag, 0 for ChromaTag, [0, 204, 25, 1314, 343] for AprilTag's [16h5, 25h7, 25h9, 36h9, 36h11] and 278 for TopoTag.
\Blue{Note that, for AprilTag, there are three shared tag families, i.e. 16h5, 25h9 and 36h11, for AprilTag-1 \cite{Goyal2011} and AprilTag-2 \cite{Wang2016}, and 25h7, 36h9 only exist in AprilTag-1. In following sections, we will report the best result of AprilTag-1 and AprilTag-2 for these shared tag families if not otherwise specified.}
For evaluation fairness, outer border sizes of all tags are kept at the same of 5 cm. For each tag, there are $\approx$100,000 images collected, including $\approx$1,000 for Seq \#1, $\approx$1,200 for Seq \#2 and $\approx$7,800 for Seq \#3. Please see \autoref{fig:samples} for sample images for each sequence.


It's worth noting that segmentation is crucial for marker detection and pose estimation for all marker systems. Thus, for fair comparison, we fine tune the segmentation parameters for each marker algorithm unless it already uses advanced approaches like adaptive thresholding, line detection, etc. Specifically, we use a threshold of 60 instead of default 100 for ARToolKit, 15 and 2 for {\verb`AdaptiveThresholdWindowSize`} and {\verb`AdaptiveThresWindowSize_range`} instead of default -1 and 0 for ArUco. Please refer to the supplementary material for the performance comparison between their default setups and our finely tuned versions.

\subsection{Dictionary Size vs. Tracking Distance}
\begin{figure}[tp]
\Blue{
\centering
\subfloat{\includegraphics[width=0.24\textwidth]{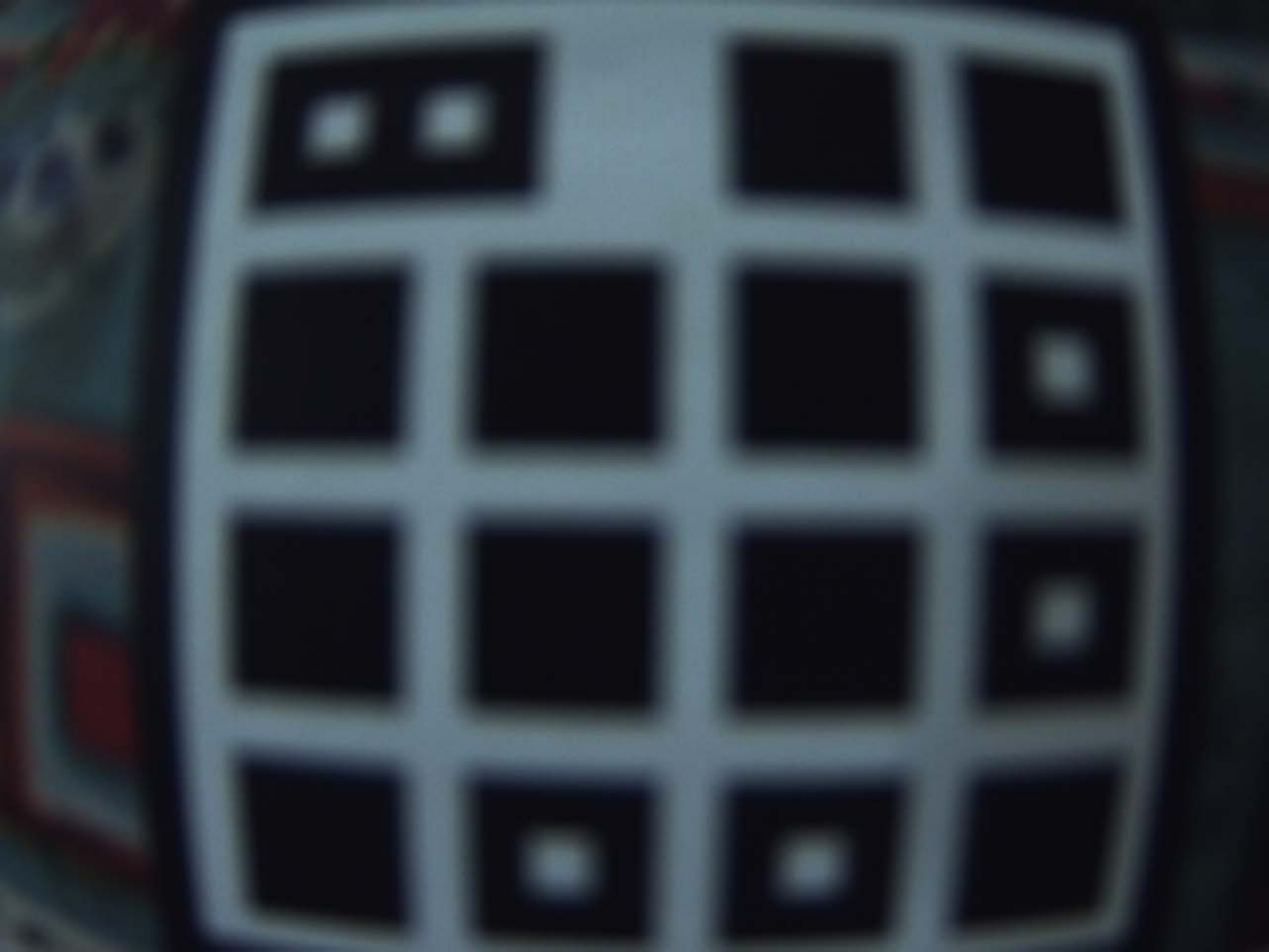}}
\hspace{0.01in}
\subfloat{\includegraphics[width=0.24\textwidth]{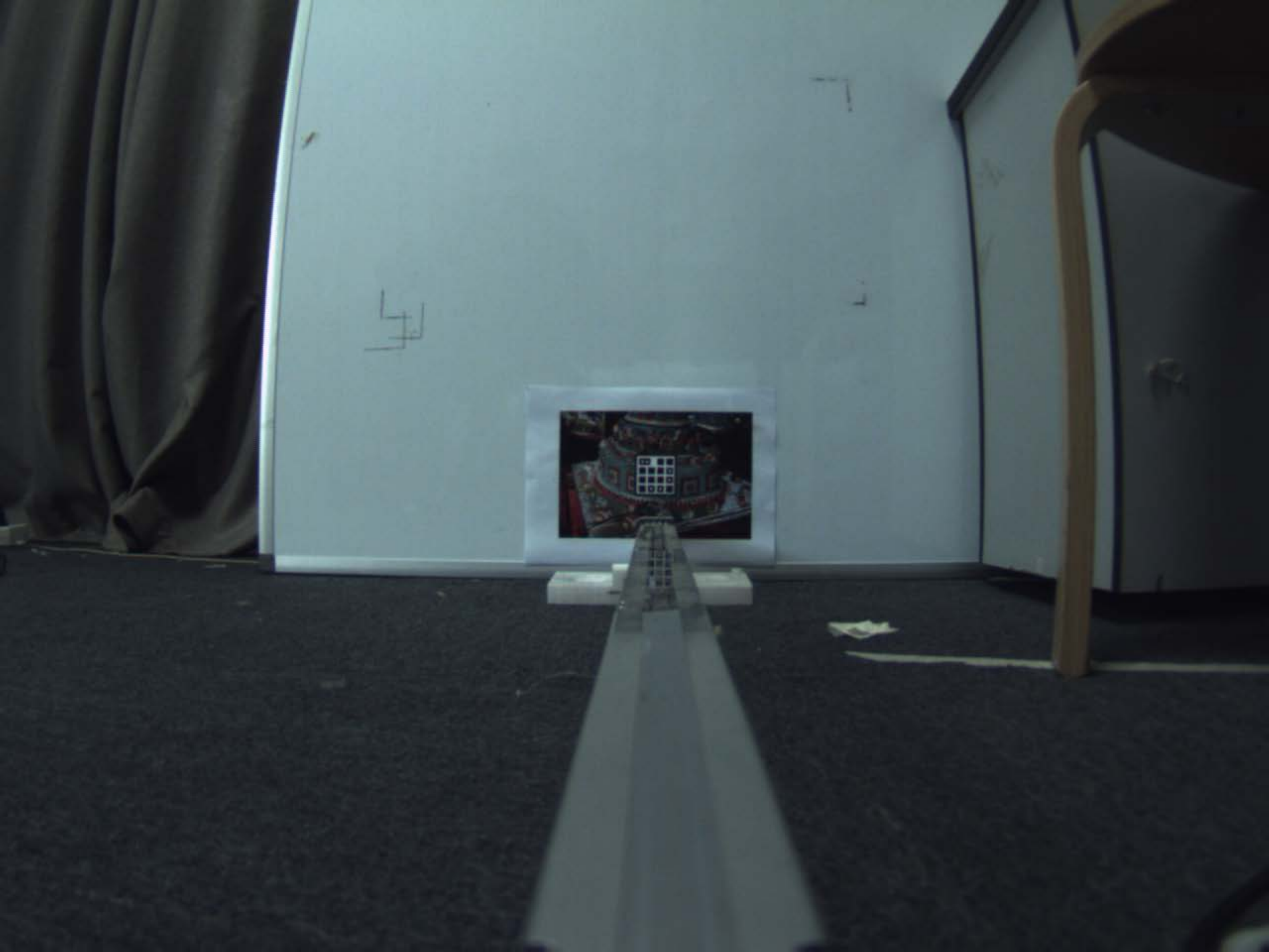}}
\caption{Images of TopoTag at minimum and maximum tracking distance respectively. Note that, there are partial marker occlusion because of the camera FoV limitation and blur issues especially at the min distance.}
\label{fig:distance}
}
\end{figure}

\autoref{tab:track-dist} shows the comparison of dictionary size vs. tracking distance \Blue{(both min and max)} of different tag systems. Generally speaking, more tag bits offer more spaces to encode identities, but sacrifice maximum tracking distance as region for each bit becomes smaller. \Blue{On the other hand, minimum tracking distance is affected by the marker occlusion because of the camera FoV limitation and blur issue at the small range of a fixed-focus camera. \autoref{fig:distance} shows the images of TopoTag at minimum and maximum tracking distance respectively. TopoTag achieves a state-of-the-art minimum tracking distance, which further demonstrates the robustness of TopoTag under partial occlusion and out-of-focus image blur.}
TopoTag also achieves comparable \Blue{maximum} tracking range when the dictionary size is small (9-bit), while offering a significantly larger tracking range when the dictionary size extends to tens of thousands (16-bit vs. RuneTag with the state-of-the-art most identities). In addition, TopoTag offers the scalability of extending the dictionary size to millions with a still acceptable tracking distance (25-bit).
\Blue{Interestingly, AprilTag-2 achieves better minimum tracking range but worse maximum tracking range over AprilTag-1 by a large margin. We suspect that this is due to the tag detection strategy change from gradient computing based in AprilTag-1 to adaptive thresholding based in AprilTag-2 for speedup.}
\Blue{It's worth noting that, we also tested markers with different sizes, including 2.5 cm and 10 cm. The conclusions for both min and max tracking ranges still hold.}

\begin{table}
\centering
\caption{Dictionary size vs. tracking distance. \Blue{For shared tag families of AprilTag-1 and AprilTag-2, results of both versions are reported with format ``AprilTag-1$\rightarrow$AprilTag-2".}}
\begin{tabular}{|c|c|c|c|}
\hline
\textbf{Tag}       & \begin{tabular}{@{}c@{}}\textbf{Dictionary} \\ \textbf{Size}\end{tabular}  & \begin{tabular}{@{}c@{}}\Blue{\textbf{Min}} \\ \Blue{\textbf{Distance (m)}}\end{tabular}       & \begin{tabular}{@{}c@{}}\textbf{Max} \\ \textbf{Distance (m)}\end{tabular} \\ \hline
ARToolKit          & 10\footnotemark   & \Blue{0.047}       					 & 1.199             \\ \hline
ARToolKitPlus      & 512               & \Blue{0.087}       					 & 1.154             \\ \hline
ArUco (16h3)       & 250               & \Blue{0.117}       					 & 1.309             \\ \hline
ArUco (25h7)       & 100               & \Blue{0.117}       					 & 1.187             \\ \hline
ArUco (36h12)      & 250               & \Blue{0.120}       					 & 1.199             \\ \hline
RuneTag            & 17,000            & \Blue{0.103}       					 & 0.221             \\ \hline
ChromaTag          & 30                & \Blue{0.547}       					 & 0.560             \\ \hline
AprilTag (16h5)    & 30                & \Blue{0.161 $\rightarrow$ 0.043}       & 1.220 \Blue{$\rightarrow$ 0.757}            \\ \hline
AprilTag (25h7)    & 242               & \Blue{0.160}     						 & 1.171             \\ \hline
AprilTag (25h9)    & 35                & \Blue{0.156 $\rightarrow$ 0.040}       & 1.226 \Blue{$\rightarrow$ 0.968}            \\ \hline
AprilTag (36h9)    & 5,329             & \Blue{0.163}      					    & 1.223             \\ \hline
AprilTag (36h11)   & 587               & \Blue{0.163 $\rightarrow$ 0.042}       & 1.168 \Blue{$\rightarrow$ 0.906}            \\ \hline
TopoTag (3x3)      & 128               & \Blue{0.029}       					 & 1.204             \\ \hline
TopoTag (4x4)      & 16,384            & \Blue{0.029}       					 & 1.055             \\ \hline
TopoTag (5x5)      & 8,388,608         & \Blue{0.029}       					 & 0.670             \\ \hline
\end{tabular}
\label{tab:track-dist}
\end{table}
\footnotetext{10 tags are provided in ARToolKit package. Theoretically, any pattern can be used for tag design, but the author didn't provide the approach.}

\subsection{Detection Accuracy}
\begin{figure}[tp]
\centering
\subfloat{\includegraphics[width=0.24\textwidth]{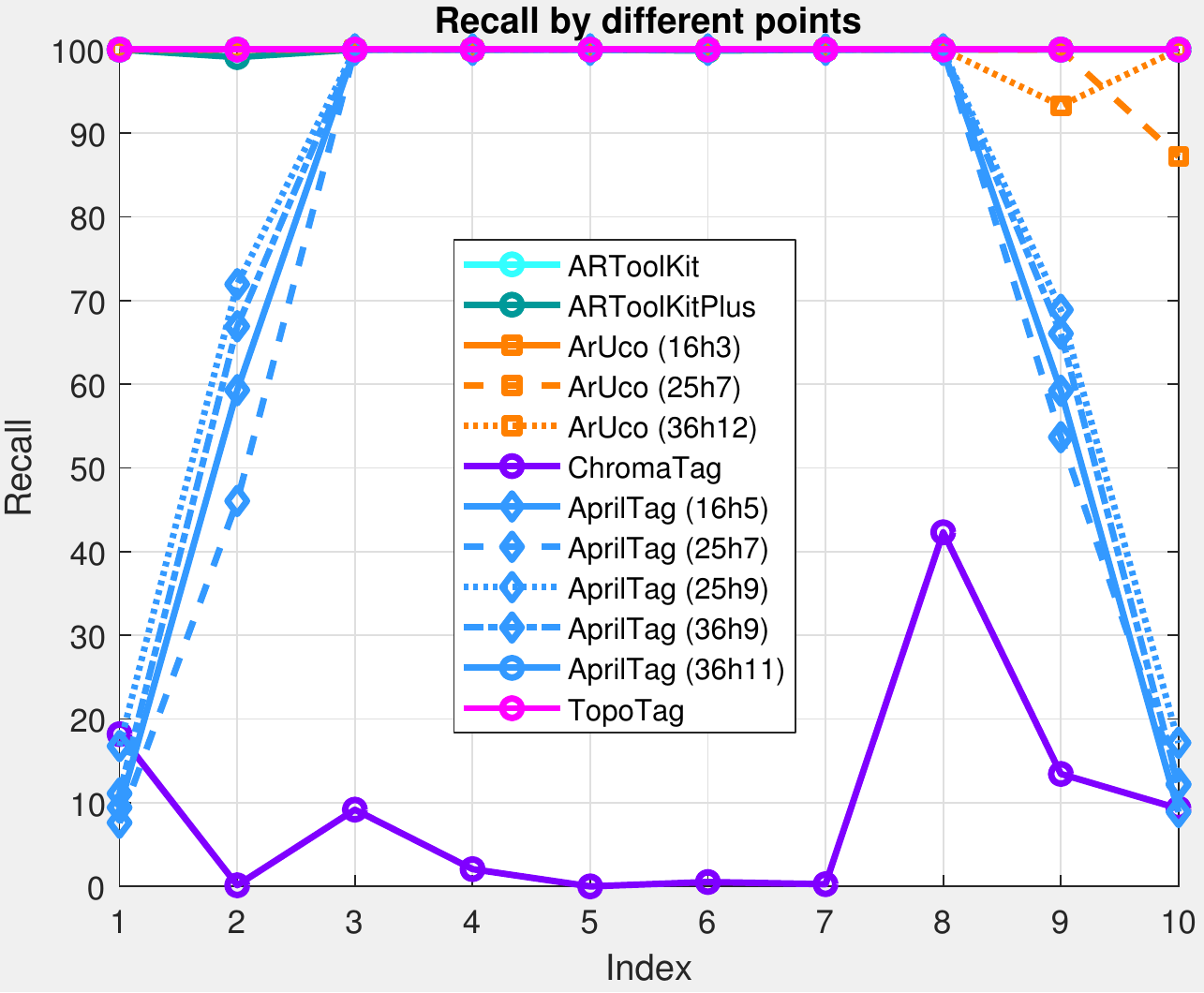}}
\hspace{0.01in}
\subfloat{\includegraphics[width=0.24\textwidth]{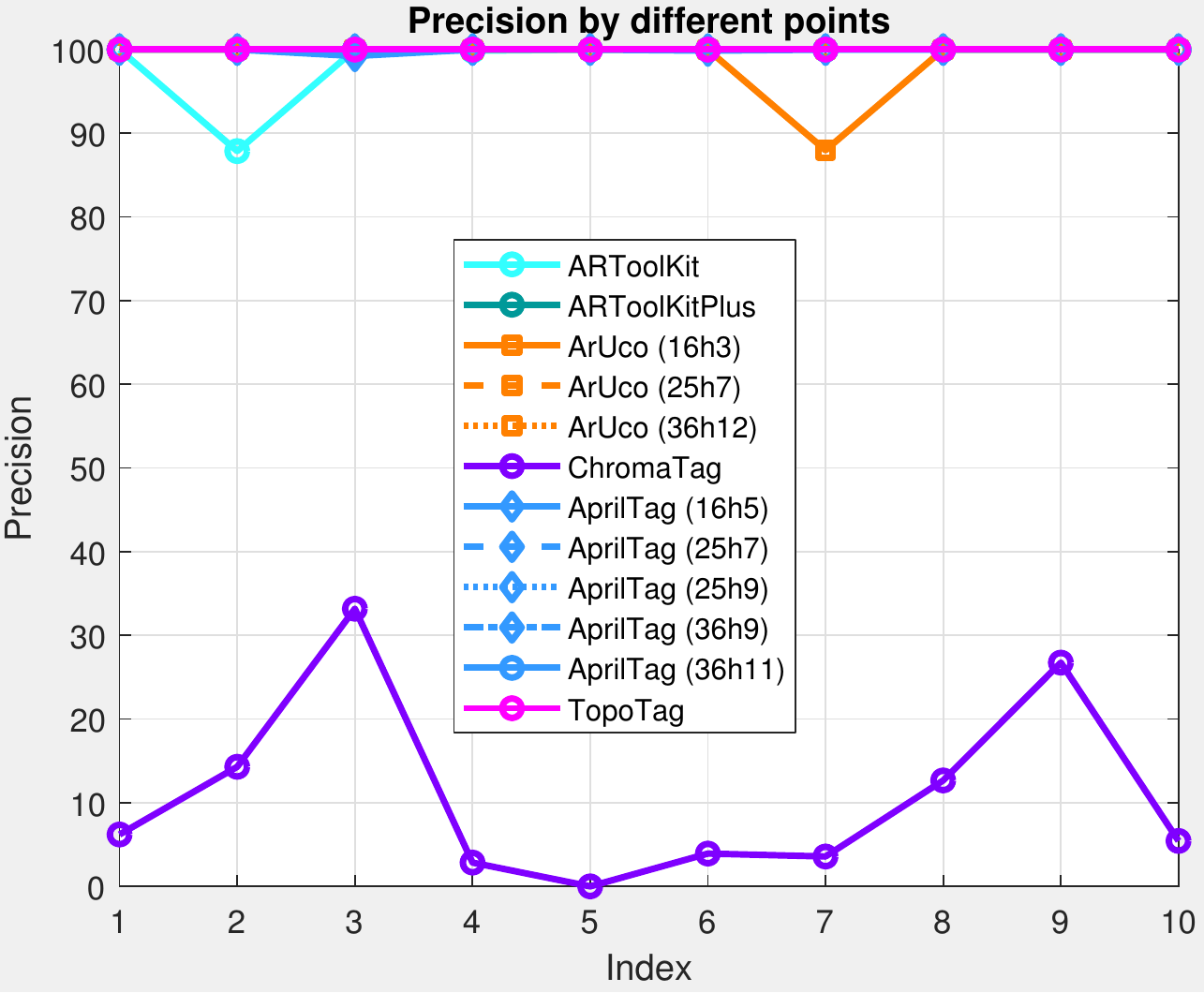}}
\caption{Recall and precision by different points on Seq \#3. (Best viewed in color)}
\label{fig:recall-precision-by-points}
\end{figure}

\autoref{tab:detect-acc} summarizes the detection results for TopoTag compared to previous marker systems. \autoref{fig:recall-precision-by-points} highlights the recall and precision of different captured points on Seq \#3. We follow the metrics used in \cite{Degol2017}. True positives (TP) are defined as when the tag is correctly detected, including locating the tag and correctly identifying the ID. Correct identification of the tag is determined by having at least 50\% intersection over union between the detection and the ground truth. False positives (FP) are defined as detections returned by the detection algorithms that do not identify the location and ID correctly. False negatives (FN) are defined as any marker that is not identified correctly. Precision is $\frac{TP}{TP+FP}$ and recall is $\frac{TP}{TP+FN}$.

\autoref{tab:detect-acc} shows that TopoTag performs perfectly on all three sequences, achieving 100\% across both recall and precision. All tested marker systems, except ChromaTag, work great and achieve $>99.5\%$ on precision due to their unique false positive rejection techniques. However, most systems except ARToolKit, ARToolKitPlus and ArUco fail to achieve a high recall, i.e. $<81\%$.
\autoref{fig:recall-precision-by-points} shows that all previous systems degrade on recall or precision or both when markers are viewed from wide angles. This result is probably from large distortion, decreased lightness and blur issues, which distract marker detection. TopoTag, on the other hand, has no obvious degradation on these issues.

ChromaTag performs worse on both recall and precision possibly due to the cluttered colored background and relative low brightness of collected images distracting its detection based on color information. For ablation study, we tried to replace all ChromaTag's background with pure white pixels and keep only the marker region. With such a setup, ChromaTag achieves the same recall (i.e. same number of false negatives) and the number of false positives decreased from 8962 to 257, which further validates that ChromaTag is sensitive to cluttered background (i.e. detecting false positives).

RuneTag performs the worst with lowest recall $<0.3\%$ and fails to detect any frame on Seq \#3 where it fails to find enough confident ellipses on the images. As also found in ChromaTag work \cite{Degol2017}, RuneTag requires larger tag sizes for detection, which is the major cause of its lesser performance on our dataset with small marker size in long distance and challenging blur. In our experiment, we found that RuneTag cannot be detected when the marker is smaller than $180\times180$ pixels.

\begin{table}
\centering
\caption{Detection accuracy (with run time). \Blue{For shared tag families of AprilTag-1 and AprilTag-2, run time of both versions are reported with format ``AprilTag-1$\rightarrow$AprilTag-2".}}
\begin{tabular}{|c|c|c|c|}
\hline
\textbf{Tag}       & \textbf{Recall (\%)} & \textbf{Precision (\%)} & \textbf{Time (ms)} \\ \hline
ARToolKit          & 99.990          & 99.880      & 5.864       \\ \hline
ARToolKitPlus      & 98.297          & 100.000     & 9.314       \\ \hline
ArUco (16h3)       & 100.000         & 99.910      & 54.319       \\ \hline
ArUco (25h7)       & 99.009          & 100.000     & 53.930       \\ \hline
ArUco (36h12)      & 99.470          & 100.000     & 56.001       \\ \hline
RuneTag            & 0.281           & 100.000     & 455.832       \\ \hline
ChromaTag          & 9.088           & 9.190       & 9.103       \\ \hline
AprilTag (16h5)  & 77.285          & 99.883        & 246.762 \Blue{$\rightarrow$ 15.114}     \\ \hline
AprilTag (25h7)  & 75.711          & 100.000       & 244.433     \\ \hline
AprilTag (25h9)  & 80.405          & 100.000       & 251.275 \Blue{$\rightarrow$ 13.603}    \\ \hline
AprilTag (36h9)  & 78.704          & 100.000       & 240.694     \\ \hline
AprilTag (36h11) & \Blue{100.000}  & \Blue{99.990} & 241.314 \Blue{$\rightarrow$ 13.431}    \\ \hline
TopoTag            & 100.000         & 100.000     & 33.638       \\ \hline
\end{tabular}
\label{tab:detect-acc}
\end{table}

\textbf{False Positive Rejection.} \quad Since all of the images in our dataset contain valid tags, FP mainly focuses on the background excluding the tag regions. To better evaluate FP, as in \cite{Wang2016}, we further run the experiment on LabelMe \cite{Russell2008} dataset, which consists of 207,883\footnote{This is the latest LabelMe dataset size, which is slightly different the size of 180,829 from that was used in \cite{Goyal2011, Wang2016}.} images of natural scenes from a wide variety of indoor and outdoor environments, none of which contain any valid fiducial markers. We run this test for ARToolKit, ARToolKitPlus, ArUco, AprilTag and TopoTag as they achieve top detection accuracy results on our dataset as shown in \autoref{tab:detect-acc}. In addition, we further run this test for reacTIVision \cite{Bencina2005_1, Bencina2005_2, Kaltenbrunner2007} which only recovers 2D location and orientation by default. There are 49321 false positives returned by AprilTag (16h5), 9756 by ARToolKit, 348 by reacTIVision and 146 by ArUco (16h3). In contrast, TopoTag and ARToolKitPlus both have no false positives.

\subsection{Localization Jitter and Accuracy}
We evaluate localization jitter (including 6-DoF pose jitter and 2D vertex jitter) and accuracy (i.e. 6-DoF pose accuracy) on Seq \#3.

\subsubsection{Pose Error}
\begin{figure}[tp]
\centering
\subfloat{\includegraphics[width=0.24\textwidth,height=0.2\textwidth]{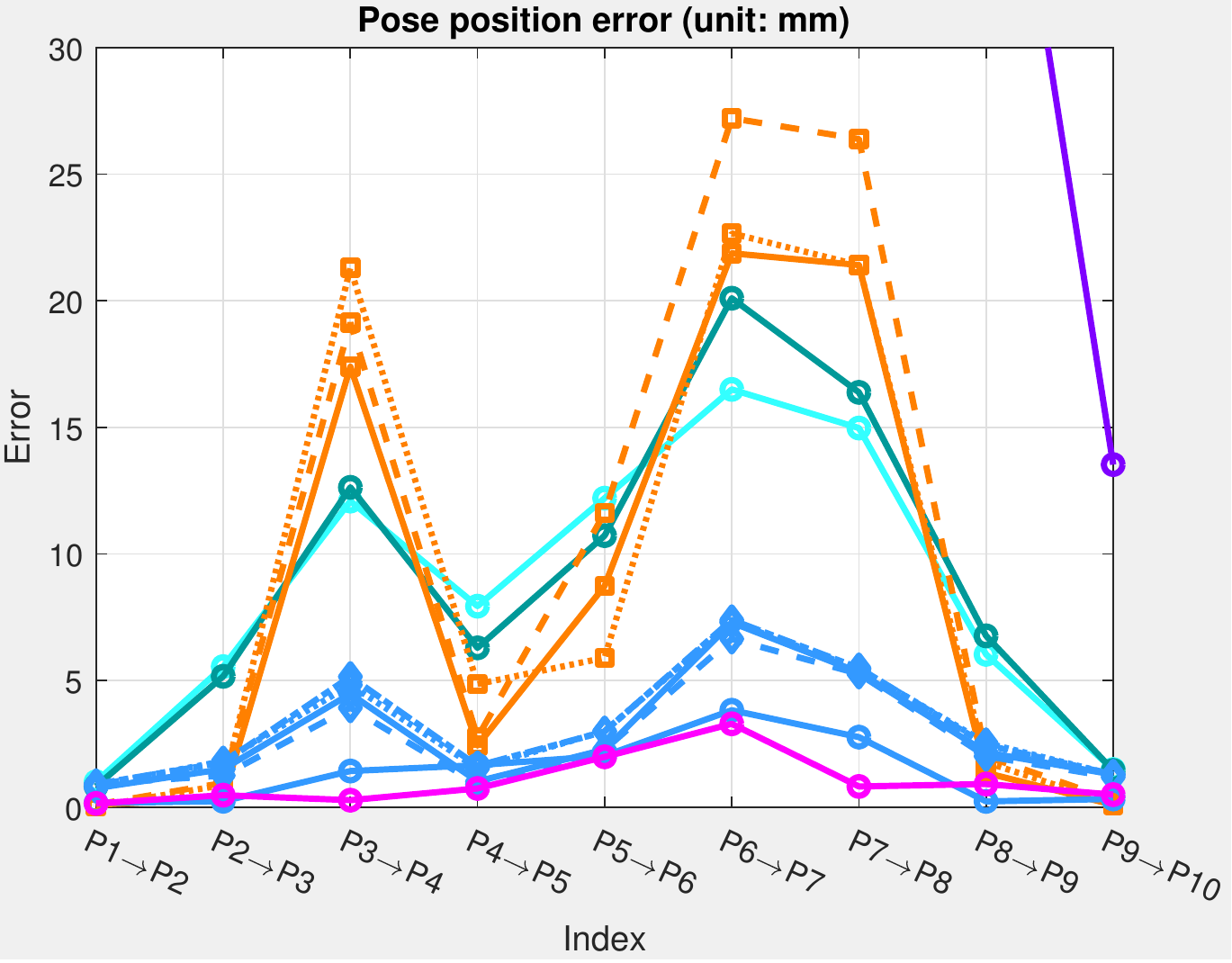}}
\hspace{0.01in}
\subfloat{\includegraphics[width=0.24\textwidth,height=0.2\textwidth]{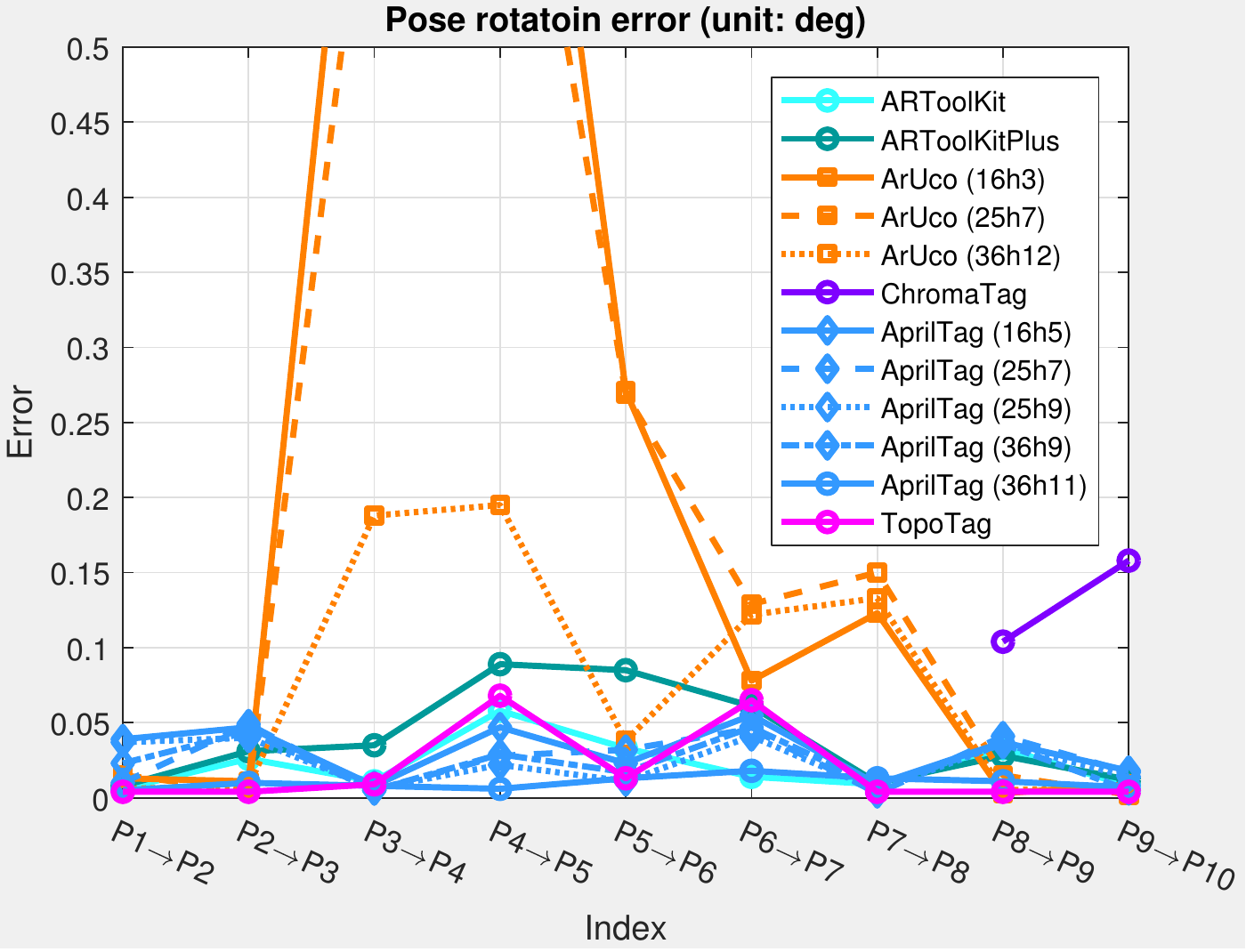}}
\caption{Pose position (left) and rotation (right) error comparison. We have trimmed the figures for better visualization. Please refer to the supplementary material for full figures. (Best viewed in color)}
\label{fig:pose-acc}
\end{figure}

We evaluate the accuracy between each point and its adjacent point. The robot's measurements serve as the groundtruth. Since there are in total 10 points in Seq \#3, nine accuracy values will be computed. See \autoref{fig:pose-acc} for the results of both position and rotation accuracies. Average and maximum pose errors for each tag are listed in \autoref{tab:pose-acc}. TopoTag outperforms all previous systems in position error by a large margin (\Blue{about 28\%} error reduction by average and \Blue{14\%} by max compared to the 2nd best) and is comparable with the state-of-the-art on rotation error \Blue{($<$0.1 degree for both average and max)}.
\Blue{A further two-sample Kolmogorov-Smirnov test shows that TopoTag significantly outperforms the 2nd best (i.e. AprilTag) in position error with $p=0.000$.}

\begin{table}
\centering
\caption{Average and maximum pose errors of each tag. Best results are shown in bold and underlined.}
\begin{tabular}{|c|c|c|c|c|}
\hline
\multirow{2}{*}{\textbf{Tag}} & \multicolumn{2}{c|}{\textbf{position (mm)}} & \multicolumn{2}{c|}{\textbf{rotation (deg)}} \\ \cline{2-5}
                     & \textbf{avg}     & \textbf{max}    & \textbf{avg}     & \textbf{max}     \\ \hline
ARToolKit            & 8.639            & 16.499          & 0.022            & 0.058            \\ \hline
ARToolKitPlus        & 8.923            & 20.101          & 0.040            & 0.089            \\ \hline
ArUco (16h3)         & 8.191            & 21.876          & 0.248            & 0.908            \\ \hline
ArUco (25h7)         & 10.049           & 27.212          & 0.225            & 0.765            \\ \hline
ArUco (36h12)        & 8.768            & 22.663          & 0.078            & 0.195            \\ \hline
ChromaTag            & 29.586           & 45.643          & 0.131            & 0.158            \\ \hline
AprilTag (16h5)    & 2.894            & 7.287           & 0.031            & 0.055            \\ \hline
AprilTag (25h7)    & 2.704            & 6.641           & 0.026            & 0.049            \\ \hline
AprilTag (25h9)    & 3.178            & 7.320           & 0.024            & 0.041            \\ \hline
AprilTag (36h9)    & 3.228            & 7.394           & 0.024            & 0.047            \\ \hline
AprilTag (36h11)   & \Blue{1.402}            & \Blue{3.824}           & \Blue{\underline{\textbf{0.010}}}            & \Blue{\underline{\textbf{0.018}}}            \\ \hline
TopoTag            & \underline{\textbf{1.011}}   & \underline{\textbf{3.289}}  & 0.019   & 0.068            \\ \hline
\end{tabular}
\label{tab:pose-acc}
\end{table}

\subsubsection{Pose Jitter}
\begin{figure}[tp]
\centering
\subfloat{\includegraphics[width=0.24\textwidth]{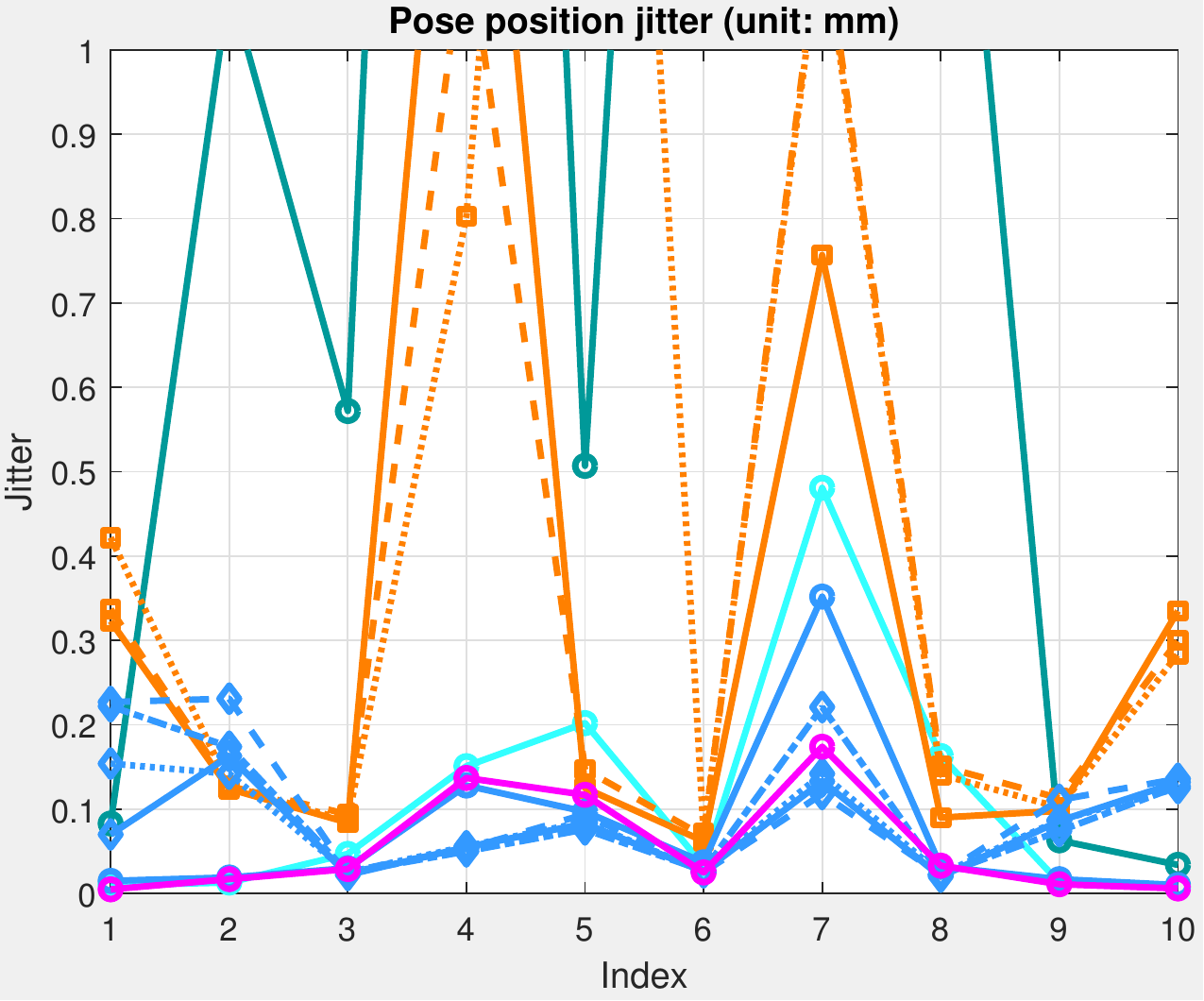}}
\hspace{0.01in}
\subfloat{\includegraphics[width=0.24\textwidth]{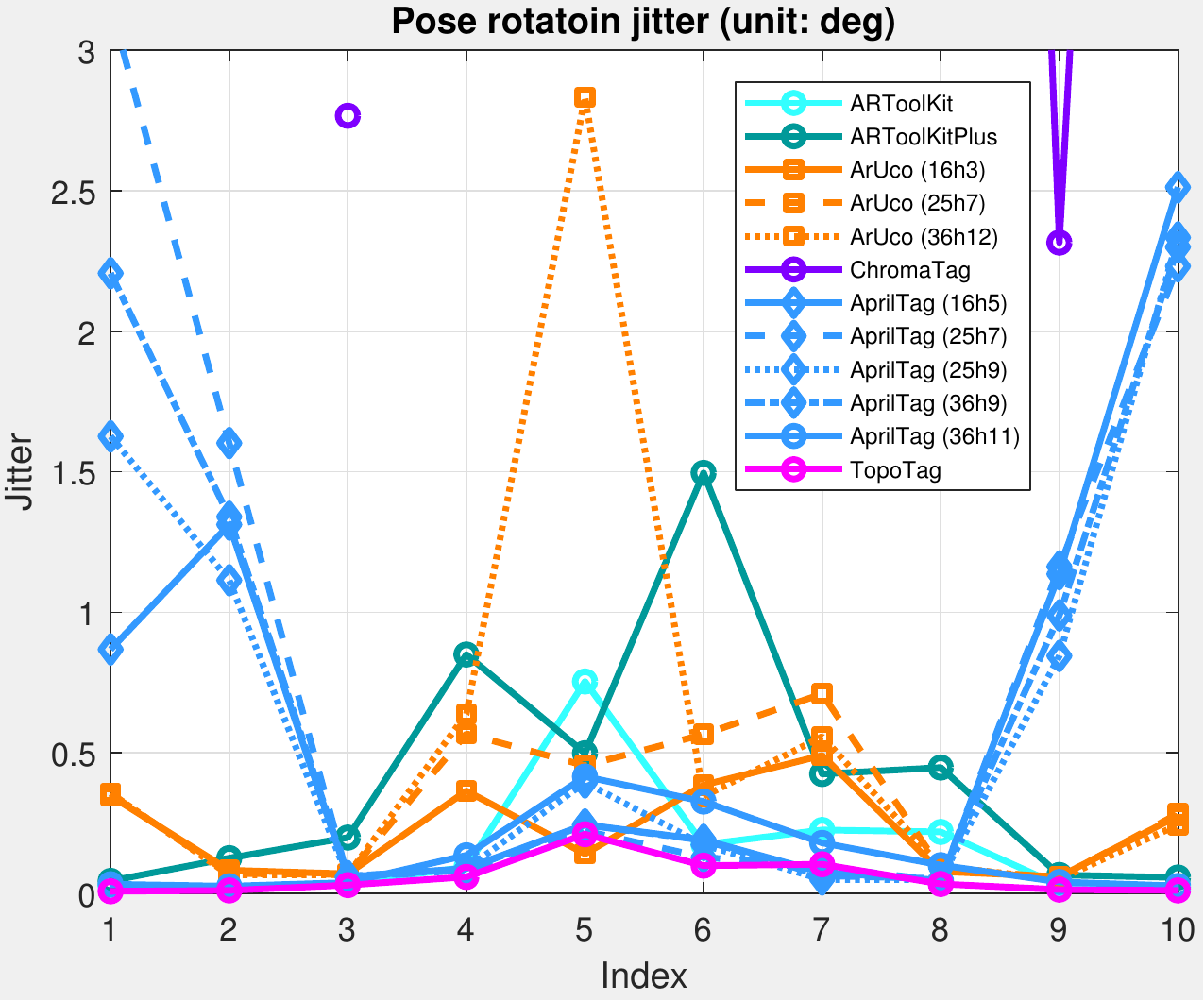}}
\caption{Pose position (left) and rotation (right) jitter comparison. We trim the figures for better visualization. Please refer to the supplementary material for full figures. (Best viewed in color)}
\label{fig:pose-jitter}
\end{figure}

Both position and rotation jitters are evaluated at each point using the standard deviation (STD) metric. See \autoref{fig:pose-jitter} for the result. Average and maximum jitter for each tag can be seen in \autoref{tab:pose-jitter}. TopoTag outperforms all previous systems in rotation jitter by a significant margin (\Blue{about 56\%} average jitter reduction and \Blue{49\%} by max compared to the 2nd best). This result is comparable with the state-of-the-art on position jitter \Blue{($<$0.1 mm for average and $<$0.2 mm for max)}.
\Blue{A further two-sample Kolmogorov-Smirnov test shows that TopoTag significantly outperform the 2nd best (i.e. AprilTag) in rotation jitter with $p=0.000$.}

\subsubsection{Vertex Jitter}
Vertex jitter measures the noise of the 2D feature point estimation, whose errors will propagate to the estimation of the 6-DoF pose. To evaluate vertex jitter, we compare two of the best previous methods, AprilTag and ArUco. Both AprilTag and ArUco are square markers, which use intersections of quad lines to achieve sub-pixel vertex precision. RUNE-Tag and ChromaTag are not evaluated as they fail to reliably detect all positions in Seq \#3, i.e. the number of detected frames for a point is less than 50\footnote{ChromaTag fails to reliably detect P2, P4, P5, P6 and P7; and all 10 positions are failed for RUNE-Tag.}. Square markers, like ARToolKitPlus and ChromaTag, theoretically will have similar performance as AprilTag and ArUco. ARToolKit is not evaluated as it uses correlation against a database to detect instead of finding fixed corners. All candidate methods are evaluated on markers with 16 bits (i.e. AprilTag's 16h5 and ArUco's 16h3). Similar to pose jitter evaluation, STD metric is used.

\begin{table}
\centering
\caption{Average and maximum pose jitters of each tag. Best results are shown in bold and underlined.}
\begin{tabular}{|c|c|c|c|c|}
\hline
\multirow{2}{*}{\textbf{Tag}} & \multicolumn{2}{c|}{\textbf{position (mm)}} & \multicolumn{2}{c|}{\textbf{rotation (deg)}} \\ \cline{2-5}
                     & \textbf{avg}     & \textbf{max}    & \textbf{avg}     & \textbf{max}     \\ \hline
ARToolKit            & 0.112            & 0.481           & 0.160            & 0.754            \\ \hline
ARToolKitPlus        & 1.134            & 3.584           & 0.421            & 1.496            \\ \hline
ArUco (16h3)         & 0.363            & 1.636           & 0.230            & 0.491            \\ \hline
ArUco (25h7)         & 0.364            & 1.155           & 0.322            & 0.710            \\ \hline
ArUco (36h12)        & 0.573            & 2.553           & 0.526            & 2.832            \\ \hline
ChromaTag            & 49.880           & 130.958         & 8.479            & 14.616           \\ \hline
AprilTag (16h5)    & 0.079            & 0.163           & 0.654            & 2.512            \\ \hline
AprilTag (25h7)    & 0.104            & 0.231           & 0.879            & 3.160            \\ \hline
AprilTag (25h9)    & 0.087            & \underline{\textbf{0.154}}  & 0.673            & 2.333            \\ \hline
AprilTag (36h9)    & 0.102            & 0.222           & 0.753            & 2.299            \\ \hline
AprilTag (36h11)   & \Blue{0.074}            & \Blue{0.352}           & \Blue{0.133}            & \Blue{0.416}            \\ \hline
TopoTag            & \underline{\textbf{0.055}}   & 0.173           & \underline{\textbf{0.058}}   & \underline{\textbf{0.211}}   \\ \hline
\end{tabular}
\label{tab:pose-jitter}
\end{table}

\begin{figure}[tp]
\centering
\subfloat{\includegraphics[width=0.24\textwidth]{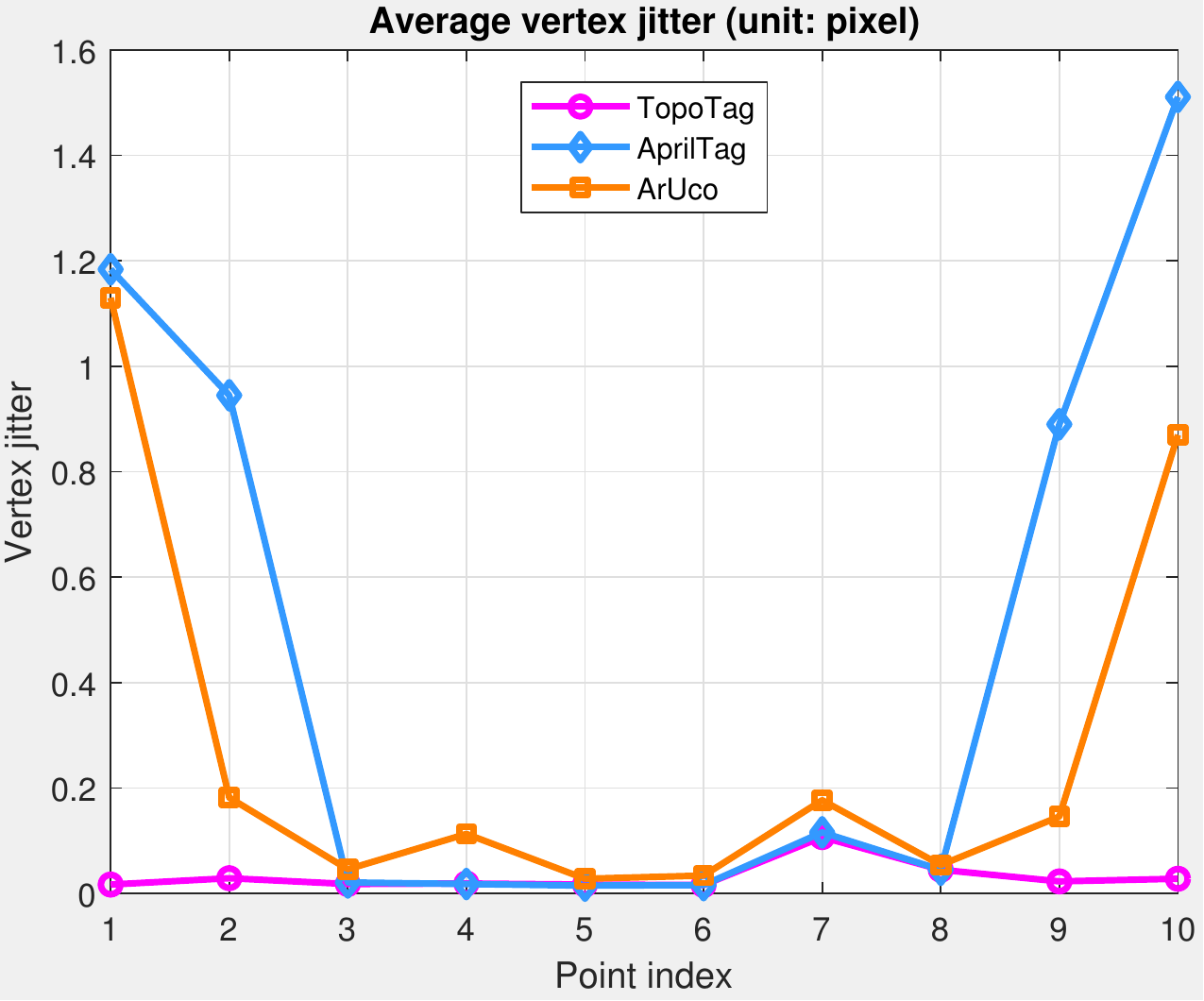}}
\hspace{0.01in}
\subfloat{\includegraphics[width=0.24\textwidth]{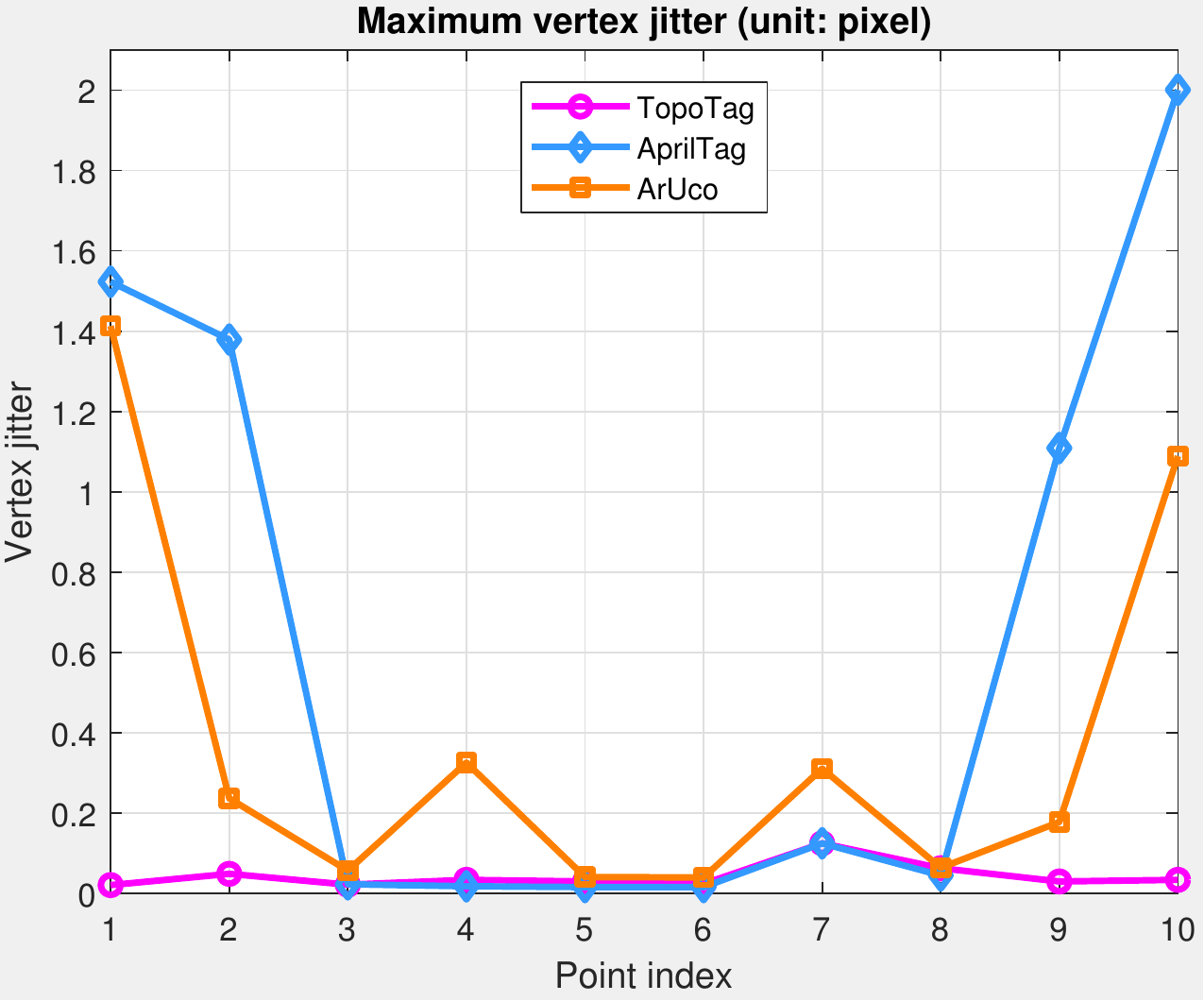}}
\caption{Average and maximum vertex jitter comparison \Blue{by different points on Seq \#3}. (Best viewed in color)}
\label{fig:vertex-jitter}
\end{figure}

Results can be seen in \autoref{fig:vertex-jitter}. It is evident that TopoTag performs consistently the best or comparable to the state-of-the-art across all points, especially when the marker angles become greater (e.g. $\geq 60 \degree$) and with more image blur (see P1, P2, P9 and P10). AprilTag performs better than ArUco where marker angles are relatively small ($\leq 30 \degree$, see P3-P8) thanks to its edge refinement, but become worse where the marker has larger angle w.r.t. the camera.

\subsection{Speed}
\subsubsection{Dictionary Computation}
Dictionary computation is usually a time-consuming operation due to the specially designed lexicode generation algorithm and Hamming distance strategy required to achieve high detection robustness. Although there is no need to do dictionary computation online normally, it's still meaningful to make this step efficient enough. ArUco takes approximately 8, 20 and 90 minutes respectively for dictionaries of sizes 10, 100 and 1000 \cite{Garrido2014}, while it can take several days to generate 36-bit tags for AprilTag \cite{Goyal2011}.
\Blue{As TopoTag supports full tag bits for identity encoding, it is extremely fast for dictionary computation as an ID can be directly mapped to the binary code string. In our experiment, it takes only 4.1 seconds to generate dictionary of size 8,388,608 (i.e. TopoTag-5$\times$5).}

\subsubsection{Tag Detection}
The last column of \autoref{tab:detect-acc} shows the running time comparison. TopoTag takes less time than ArUco ($38\%\Downarrow$), \Blue{AprilTag-1} ($86\%\Downarrow$) and RuneTag ($93\%\Downarrow$). Though ARToolKit, ChromaTag\Blue{, AprilTag-2} and ARToolKitPlus run faster than TopoTag, they offer significantly less unique identities. See \autoref{tab:track-dist} for details.
For TopoTag, most time is spent on segmentation ($68.8\%$), followed by decoding and vertex estimation ($29.7\%$). Pose estimation takes the least time ($1.5\%$).

It's worth noting that no parallelization is utilized in current TopoTag implementation, which will normally bring further speed-up. To demonstrate the possible applications on mobile, we have implemented the 2D marker detection process in a single pipeline on a Lattice FPGA (LFE5UM-45 with 44k LUTs, 1.9 Mb RAM and without using external DDR), which is decreased to $<100$ us achieving $230\times$ speedup.


\subsection{Flexible Shape Support}
\begin{figure}[tp]
\centering
\subfloat{\includegraphics[width=0.24\textwidth]{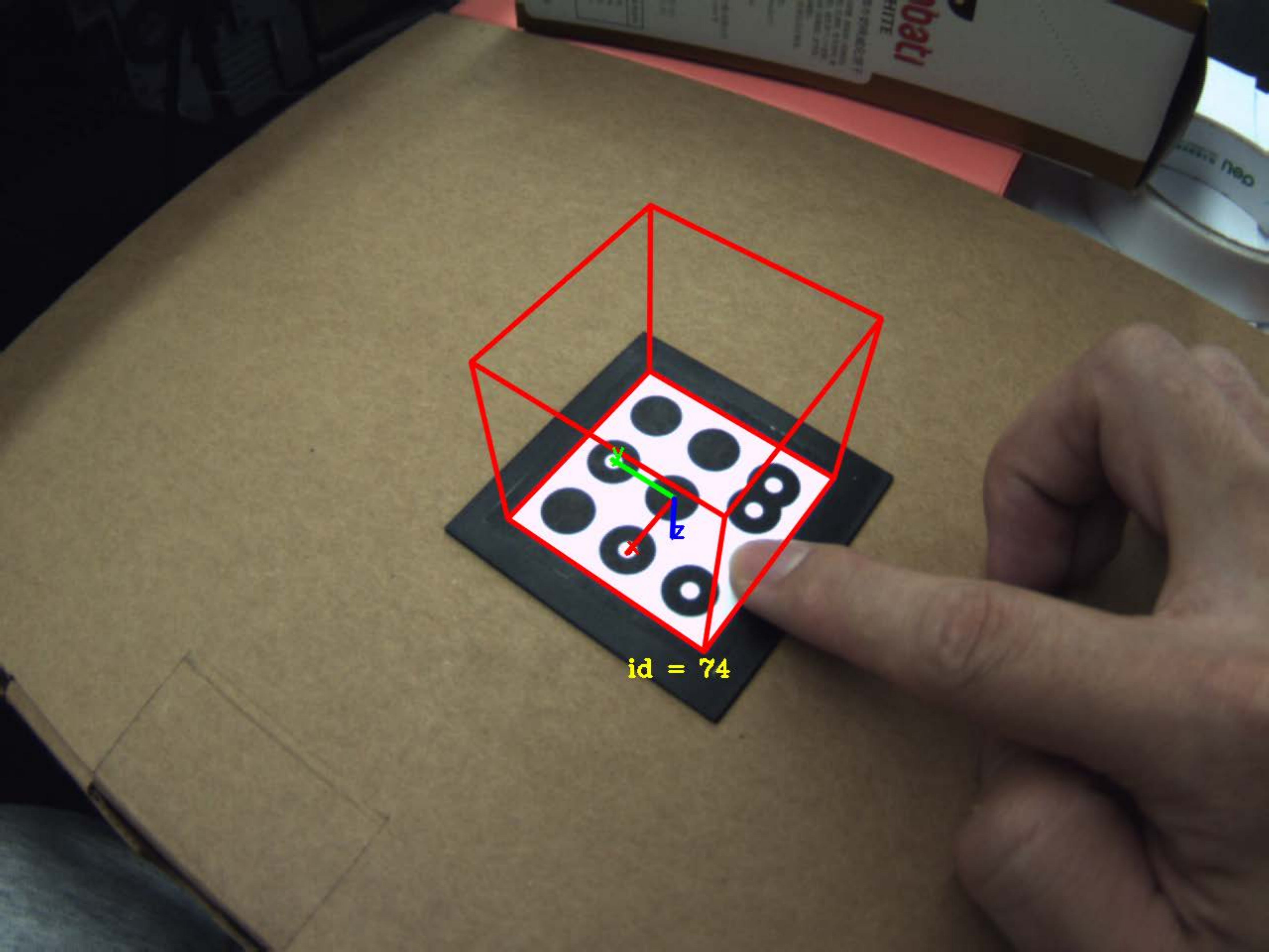}}
\hspace{0.01in}
\subfloat{\includegraphics[width=0.24\textwidth]{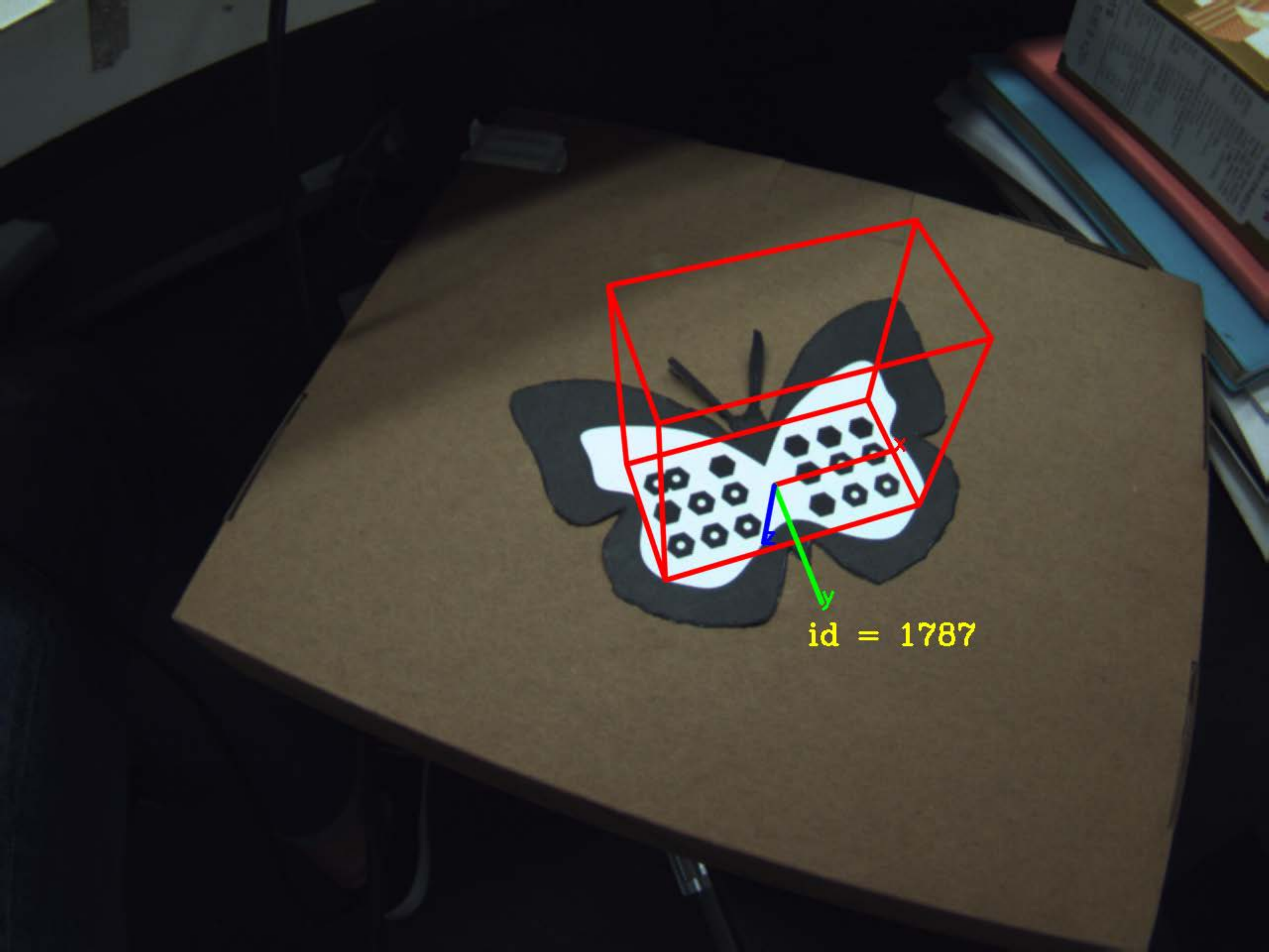}}
\caption{Detection and pose estimation of two customized TopoTags. (Best viewed in color)}
\label{fig:diff-shape}
\end{figure}

TopoTag supports both customized external and internal shapes as long as the topological structure is maintained. \autoref{fig:topotag} shows three TopoTags with various internal shapes like square, circle, hexagon and different external shapes including square and butterfly. \autoref{fig:diff-shape} shows our algorithm running upon these customized TopoTags.

Experiments show that tags with different shapes have compatible results. Please see \autoref{tab:diff-shapes} for detailed comparison. It's worth noting that all these four different TopoTags have 100\% result on both detection recall and precision which further validates the robustness of the TopoTag system.

\begin{table*}
\centering
\caption{Pose estimation of different bits and shapes.}
\begin{tabular}{|c|c|c|c|c|c|c|c|c|}
\hline
\multirow{3}{*}{\begin{tabular}{@{}c@{}} \textbf{Different} \textbf{Bits \& Shapes}\end{tabular} } & \multicolumn{4}{c|}{\textbf{Pose Accuracy}}     & \multicolumn{4}{c|}{\textbf{Pose Jitter}}    \\ \cline{2-9}
                     & \multicolumn{2}{c|}{\textbf{position (mm)}} & \multicolumn{2}{c|}{\textbf{rotation (deg)}} & \multicolumn{2}{c|}{\textbf{position (mm)}} & \multicolumn{2}{c|}{\textbf{rotation (deg)}} \\ \cline{2-9}
                     & \textbf{avg} & \textbf{max}              & \textbf{avg} & \textbf{max}              & \textbf{avg} & \textbf{max}             & \textbf{avg} & \textbf{max}             \\ \hline
3x3, circle & 1.073 & 3.299 & 0.016 & 0.048 & 0.069 & 0.265 & 0.080 & 0.205 \\ \hline
3x3, square & 0.837 & 2.780 & 0.022 & 0.065 & 0.085 & 0.383 & 0.081 & 0.276 \\ \hline
4x4, circle & 0.995 & 2.867 & 0.017 & 0.057 & 0.058 & 0.192 & 0.073 & 0.272 \\ \hline
4x4, square & 1.011 & 3.289 & 0.019 & 0.068 & 0.055 & 0.173 & 0.058 & 0.211 \\ \hline
\end{tabular}
\label{tab:diff-shapes}
\end{table*}

\subsection{Occlusion Support} \label{sec:occlusion}
\begin{figure}[tp]
\Blue{
\centering
\subfloat{\includegraphics[width=0.5\textwidth]{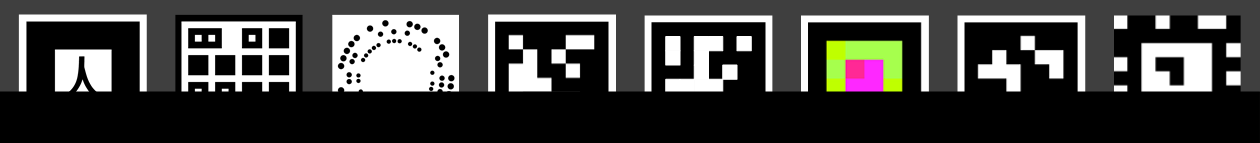}}
\caption{Occlusion test by blocking 40\% marker area starting from away the baseline node. Makers from left to right are ARToolKit, TopoTag, RuneTag, ArUco, ARToolKitPlus, ChromaTag, AprilTag-1\&2 and AprilTag-3 respectively. (Best viewed in color)}
\label{fig:occlusion}
}
\end{figure}

TopoTag can handle occlusion as long as topological structure is preserved. The left image of \autoref{fig:diff-shape} is an example working under occlusion.
\Blue{Similar to \cite{10.1007/978-3-319-99582-3_26}, we conduct an occlusion test by blocking different percentages (10\%$\rightarrow$100\% with 10\% step size) of the marker area. As TopoTag is used with a unique baseline node, for fairness, we conduct the occlusion test twice, i.e. one starting from the baseline node side and the other away from it. \autoref{fig:occlusion} shows an example of an occlusion test setup with result in \autoref{tab:occlusion}. We can see that all markers except TopoTag and RuneTag fail all occlusion tests. RuneTag achieves the best occlusion performance with max 30\% occlusion and TopoTag can work well with up to 10\% occlusion. Note that, as shown in above results, RuneTag has limitations of low detection rate and narrow tracking range due to its requirement of finding enough confident ellipses.}

\begin{table}
\Blue{
\centering
\caption{Occlusion test result. ``top$\rightarrow$bottom occlusion" and ``bottom$\rightarrow$top occlusion" means occlusion starting from and away from the baseline node side respectively.}
\resizebox{0.5\textwidth}{!}{
\begin{tabular}{|c|c|c|c|c|c|c|c|c|}
\hline
\multirow{2}{*}{Tag} & \multicolumn{4}{c|}{top$\rightarrow$bottom occlusion} & \multicolumn{4}{c|}{bottom$\rightarrow$top occlusion} \\ \cline{2-9}
                     & 10\%        & 20\%        & 30\%		& $\geq$40\%        & 10\%       & 20\%      & 30\%  	 & $\geq$40\%   \\ \hline
ARToolKit            & \xmark      & \xmark      & \xmark	& \xmark                  		 & \xmark     & \xmark    & \xmark    & \xmark                   \\ \hline
ARToolKitPlus        & \xmark      & \xmark      & \xmark	& \xmark                  		 & \xmark     & \xmark    & \xmark    & \xmark                   \\ \hline
ArUco                & \xmark      & \xmark      & \xmark	& \xmark                  		 & \xmark     & \xmark    & \xmark    & \xmark                   \\ \hline
RuneTag              & \cmark      & \cmark      & \cmark	& \xmark                  		 & \cmark     & \cmark    & \cmark    & \xmark                   \\ \hline
ChromaTag            & \xmark      & \xmark      & \xmark	& \xmark                  		 & \xmark     & \xmark    & \xmark    & \xmark                   \\ \hline
AprilTag-1\&2        & \xmark      & \xmark      & \xmark	& \xmark                  		 & \xmark     & \xmark    & \xmark    & \xmark                   \\ \hline
AprilTag-3           & \xmark      & \xmark      & \xmark	& \xmark                  		 & \xmark     & \xmark    & \xmark    & \xmark                   \\ \hline
TopoTag              & \cmark      & \xmark      & \xmark	& \xmark                  		 & \cmark     & \xmark    & \xmark    & \xmark                   \\ \hline
\end{tabular}
}
\label{tab:occlusion}
}
\end{table}



\begin{figure}[tp]
\centering
\subfloat{\includegraphics[width=0.24\textwidth]{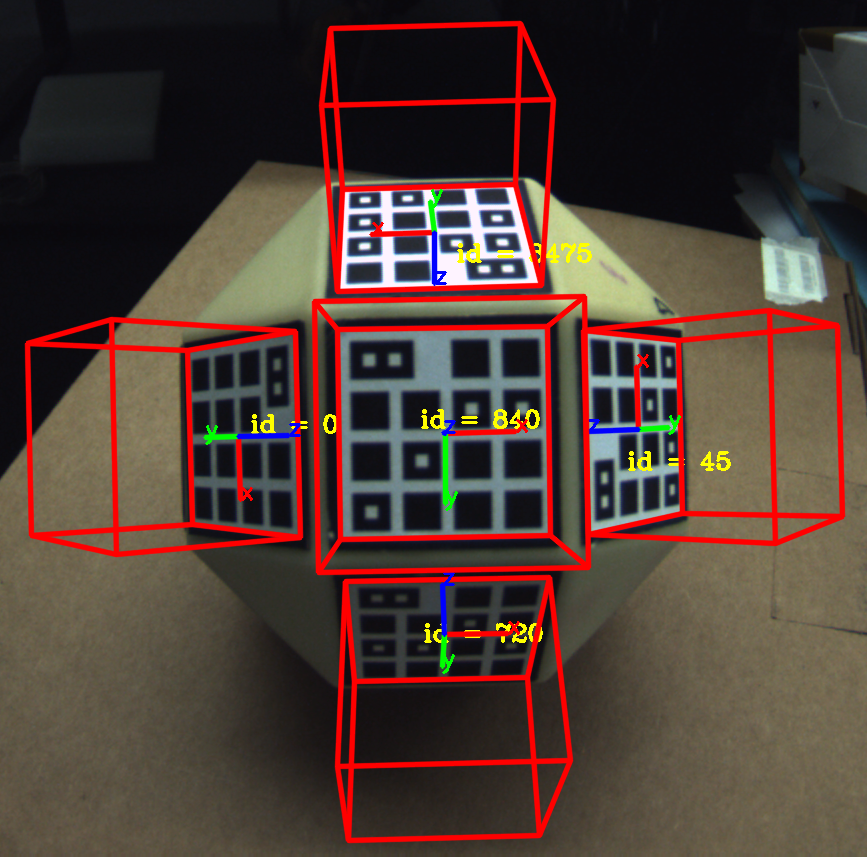}}
\caption{$360\degree$-freedom tracking via using 18 TopoTags on a rhombicuboctahedron-shaped object. (Best viewed in color)}
\label{fig:cube}
\end{figure}

To handle more severe occlusions, similar to \cite{Atcheson2010, Garrido2014}, we can use multiple tags in a grid to increase the probability of detecting complete markers and other forms can be also considered. \autoref{fig:cube} shows an example of achieving $360\degree$-freedom tracking using 18 TopoTags on a rhombicuboctahedron-shaped object.

\subsection{Noise Handing}
\begin{figure}[tp]
\centering
\subfloat[Original image.]{\includegraphics[width=0.24\textwidth]{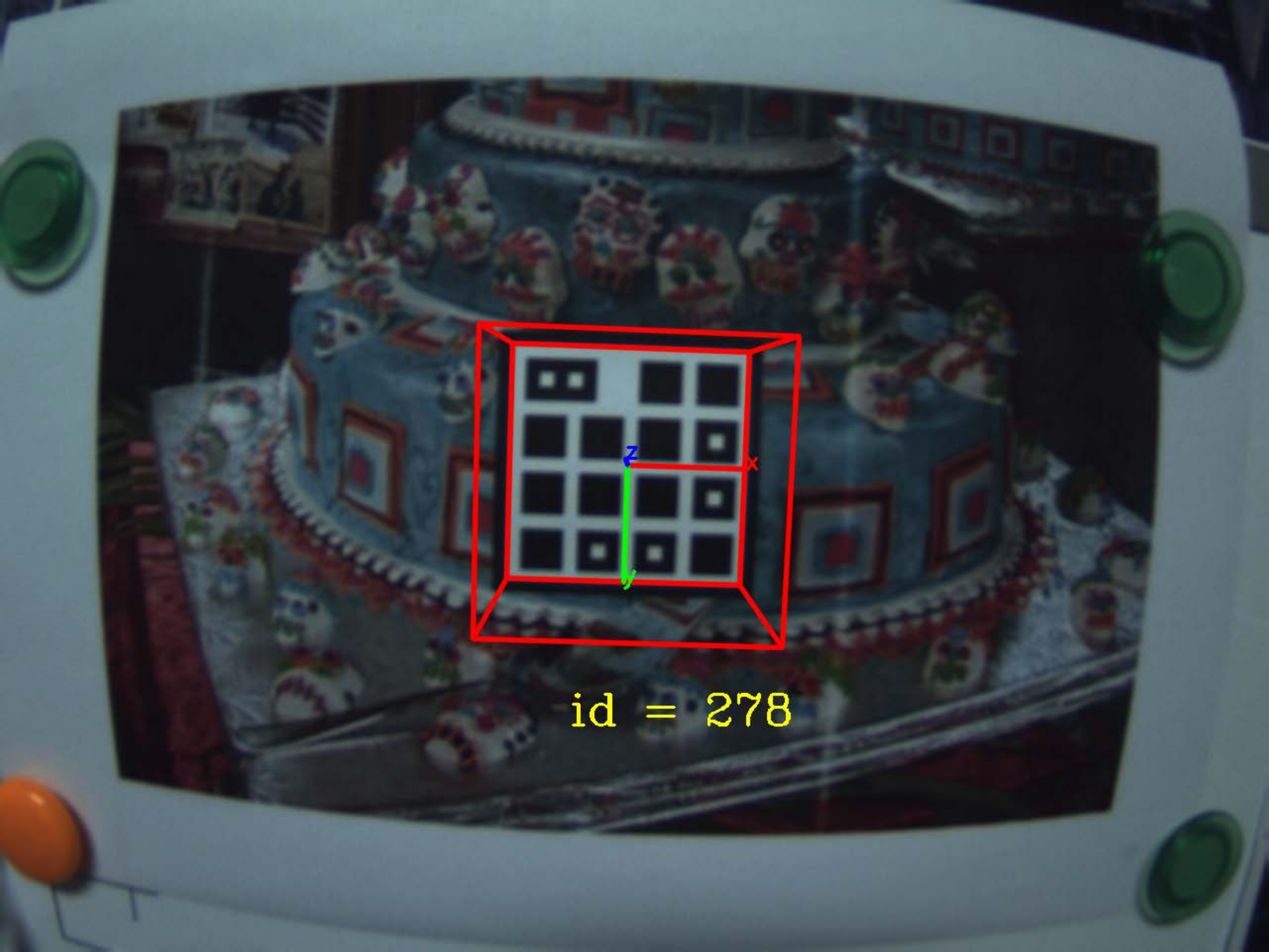}}
\hspace{0.01in}
\subfloat[Image after adding noise.]{\includegraphics[width=0.24\textwidth]{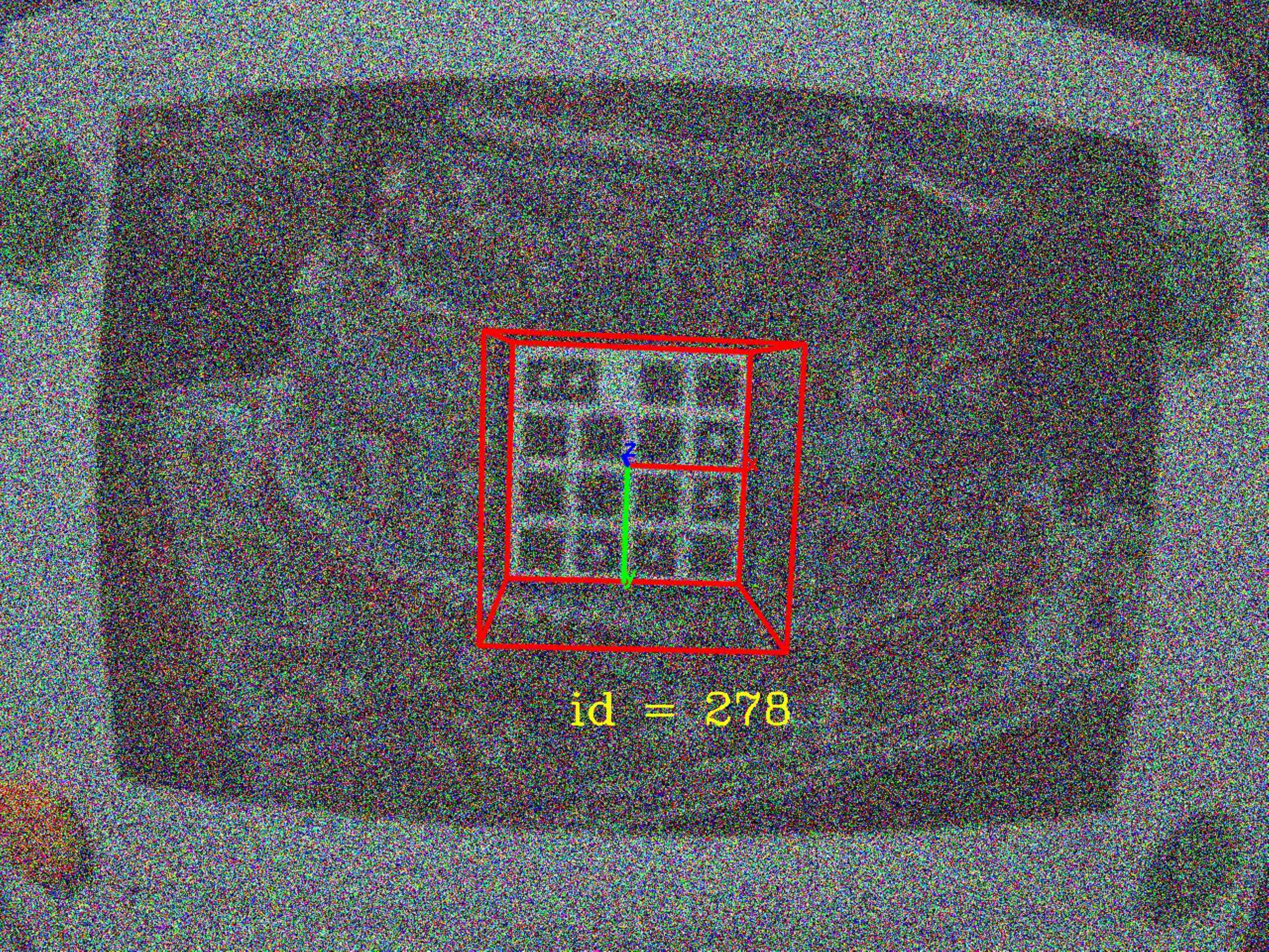}}
\caption{Example of TopoTag working under severe noise. Original image is from Seq \#1. Noise image is obtained by adding Gaussian noise $\sigma=0.45$ to the original image.}
\label{fig:noise}
\end{figure}

TopoTag can handle certain noise due to our specially designed threshold map estimation, topological filtering and error correction. In \autoref{fig:noise}, we show an example of TopoTag working under severe noise (adding Gaussian noise $\sigma=0.45$ to the original image) by introducing image smoothing (i.e. Gaussian blur with kernel size $=5, \ \sigma_x = 5.5, \ \sigma_y = 5.5$) as the pre-processing step.

\begin{figure*}[tp]
\Blue{
\centering
\subfloat[Dark.]{\includegraphics[width=0.24\textwidth]{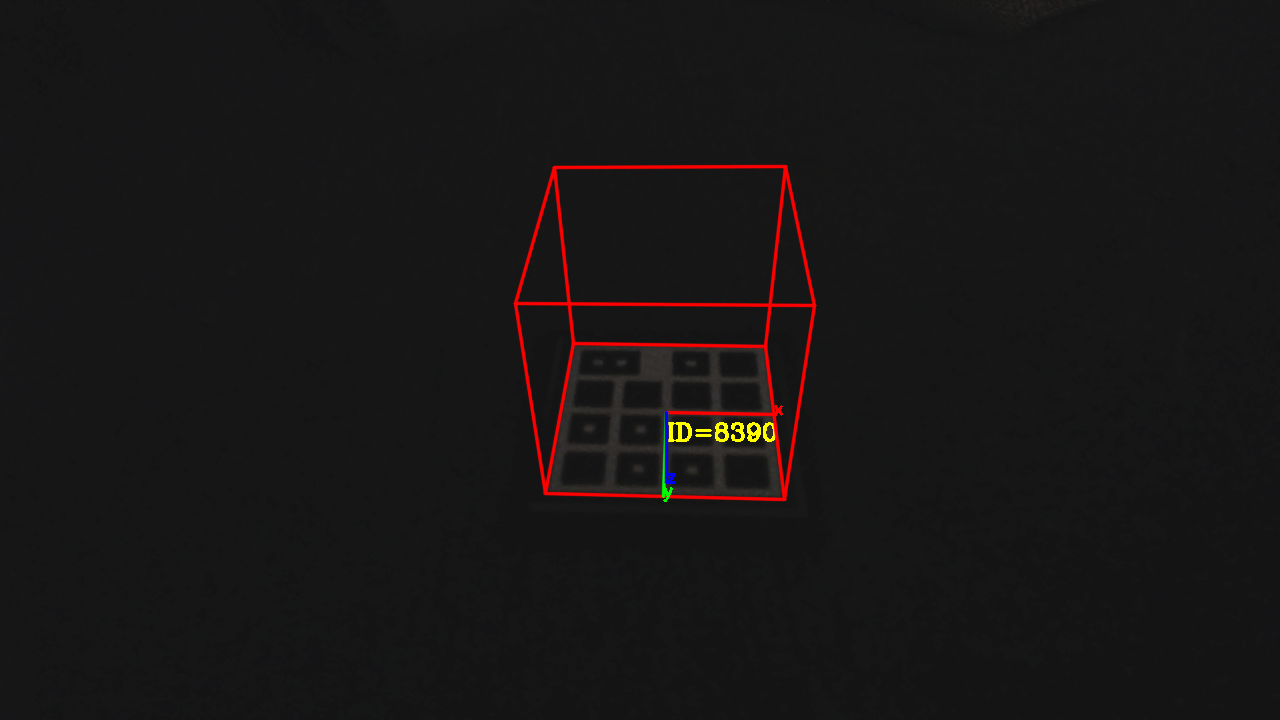}}
\hspace{0.01in}
\subfloat[Bright outdoor.]{\includegraphics[width=0.24\textwidth]{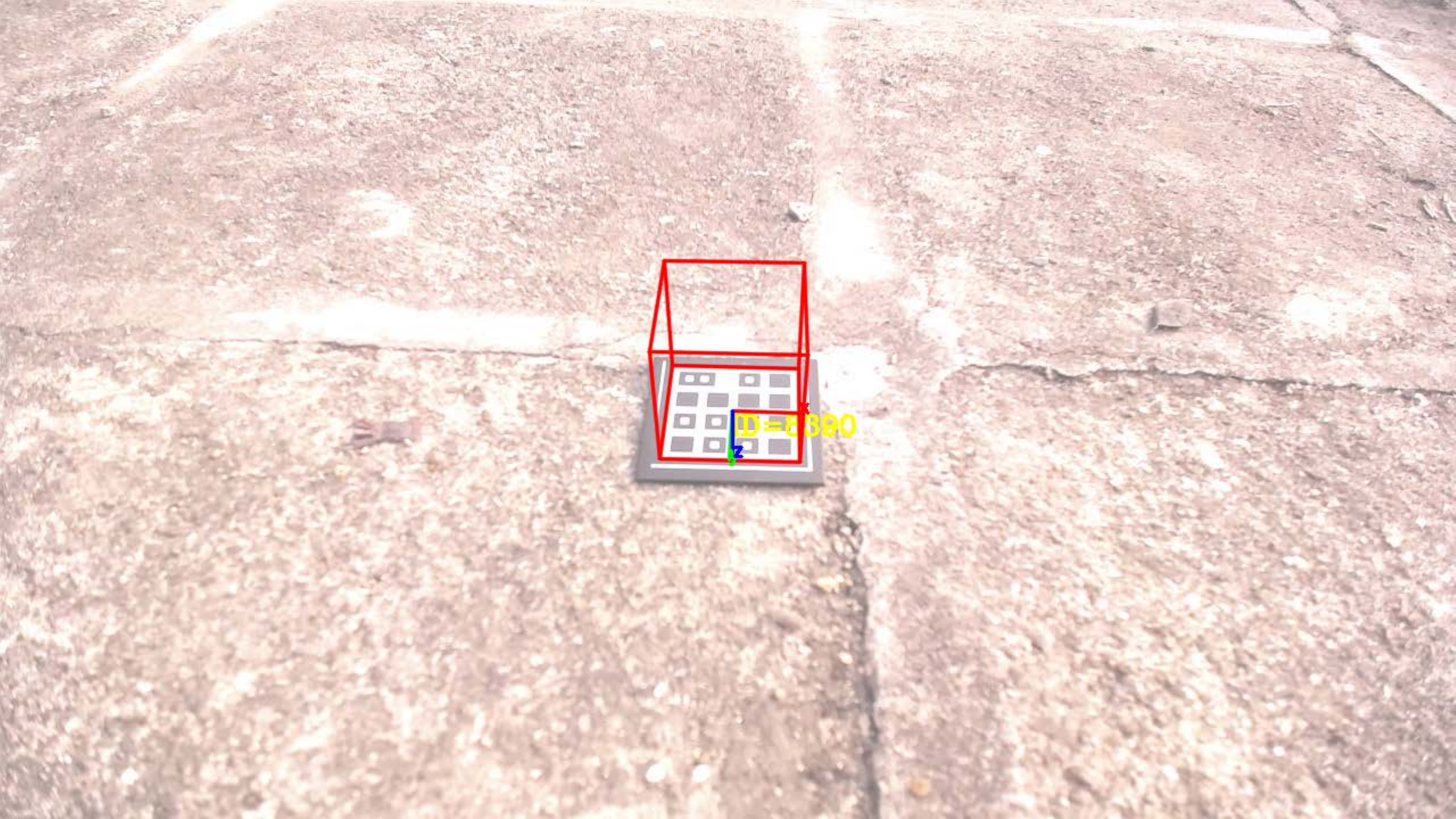}}
\hspace{0.01in}
\subfloat[Shadow.]{\includegraphics[width=0.24\textwidth]{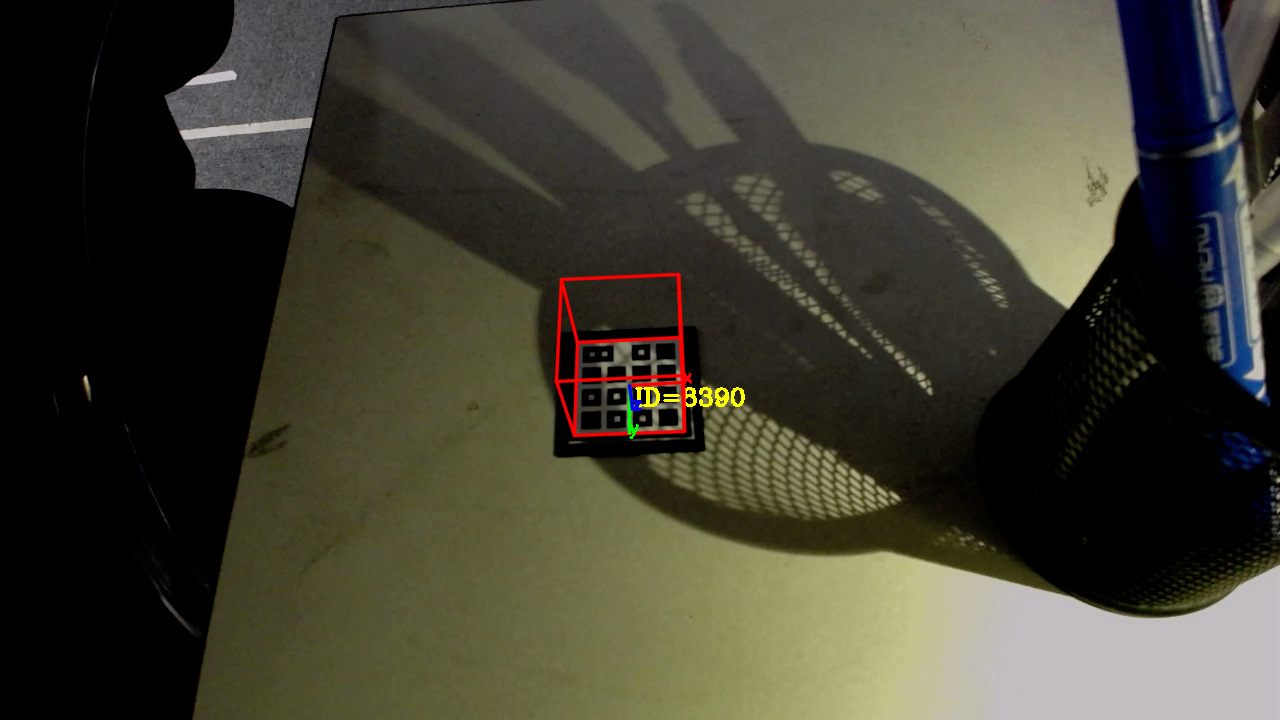}}
\hspace{0.01in}
\subfloat[Motion blur.]{\includegraphics[width=0.24\textwidth]{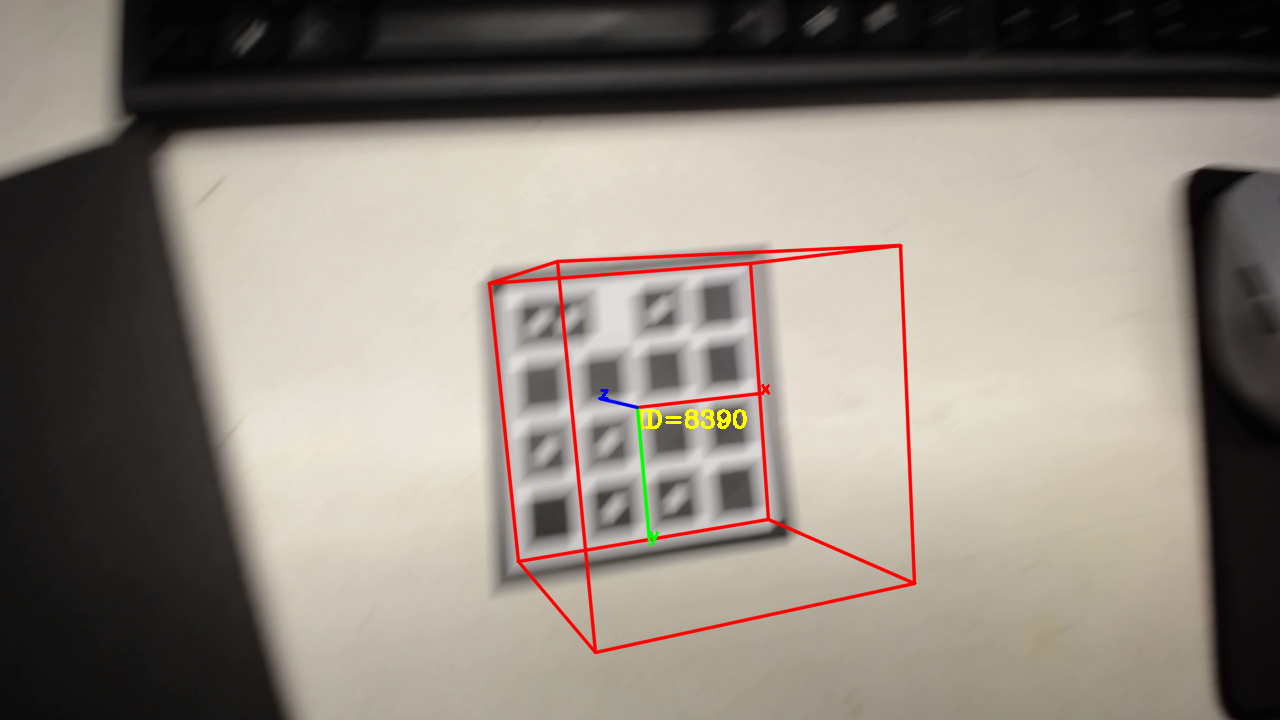}}
\caption{TopoTag detection in different real scene tests with a rolling shutter camera. (Best viewed in color)}
\label{fig:rolling-test}
}
\end{figure*}

\Blue{
\subsection{Real Scene Test with a Rolling Shutter Camera}
}
\Blue{Besides the above laboratory testing with a global shutter camera, we further conduct real indoor and outdoor scene tests with a rolling shutter camera which is widely used in mobile phones and other smart devices. Specifically, we use a Logitech C930E webcam with 1280$\times$720 resolution at 30 fps and 90 $\deg$ diagonal field of view. The experiment is conducted in four different scenarios, including dark, bright outdoor, shadow and motion blur. \autoref{fig:rolling-test} shows the test setup with TopoTag detection overlay. For dark, bright outdoor and shadow scenarios, we also evaluate pose jitter and compare it with existing markers including the latest AprilTag-3 \cite{krogius2019iros}. For fairness, for markers with multiple tag families, we randomly select one tag from the tag family with closest and smaller dictionary size compared with used TopoTag-4x4. Results are evaluated for a fixed length sequence of 100 frames and results shown in \autoref{tab:pose-jitter-rolling}. We can see that TopoTag, together with ARToolKitPlus and AprilTag-3, performs well in all test scenarios, while all other markers fail in at least one scenarios. TopoTag also achieves the best (9 out of 12) or 2nd best (3 out of 12) performance in terms of position and rotation jitter for all scenarios.}

\begin{table*}[tp]
\centering
\Blue{
\caption{Average and maximum pose jitters of each tag under different real test scenarios. ``---" means detection failure. Best results are shown in bold and underlined.}
\begin{tabular}{|c||c|c|c|c||c|c|c|c||c|c|c|c|}
\hline
\multirow{3}{*}{\textbf{Tag}}
    & \multicolumn{4}{c||}{\textbf{Dark}} & \multicolumn{4}{c||}{\textbf{Bright}}  & \multicolumn{4}{c|}{\textbf{Shadow}} \\ \cline{2-13}
        & \multicolumn{2}{c|}{\textbf{position (mm)}} & \multicolumn{2}{c||}{\textbf{rotation (deg)}}
        & \multicolumn{2}{c|}{\textbf{position (mm)}} & \multicolumn{2}{c||}{\textbf{rotation (deg)}}
        & \multicolumn{2}{c|}{\textbf{position (mm)}} & \multicolumn{2}{c|}{\textbf{rotation (deg)}} \\ \cline{2-13}
                     & \textbf{avg}     & \textbf{max}    & \textbf{avg}     & \textbf{max}  & \textbf{avg}     & \textbf{max}    & \textbf{avg}     & \textbf{max}  & \textbf{avg}     & \textbf{max}    & \textbf{avg}     & \textbf{max}  \\ \hline
ARToolKit            & ---   & ---   & ---   & ---   & ---   & ---   & ---   & ---   & 0.504             & 0.879           & 0.093	         & 0.210            \\ \hline
ARToolKitPlus        & 0.103 & 1.301 & 0.068 & 0.200 & \underline{\textbf{0.020}} & 0.113 & 0.030 & 0.110 & 1.399	         & 3.711	       & 0.469	         & 0.960            \\ \hline
ArUco                & ---   & ---   & ---   & ---   & 0.056 & 0.183 & 0.023 & 0.077 & 1.768	         & 7.793	       & 1.199	         & 4.913            \\ \hline
RuneTag              & ---   & ---   & ---   & ---   & ---   & ---   & ---   & ---   & ---               & ---             & ---             & ---              \\ \hline
ChromaTag            & ---   & ---   & ---   & ---   & ---   & ---   & ---   & ---   & ---               & ---             & ---             & ---              \\ \hline
AprilTag-1\&2        & 0.065 & 0.224 & 0.040 & 0.139 & 0.033 & 0.118 & 0.018 & 0.066 & ---               & ---             & ---             & ---              \\ \hline
AprilTag-3           & 0.112 & 0.334 & 0.060 & 0.228 & 0.044 & 0.462 & 0.020 & 0.099 & 0.103             & 0.523	       & \underline{\textbf{0.027}}	         & \underline{\textbf{0.099}}            \\ \hline
TopoTag              & \underline{\textbf{0.038}} & \underline{\textbf{0.105}} & \underline{\textbf{0.030}} & \underline{\textbf{0.113}}
                     & 0.021 & \underline{\textbf{0.066}} & \underline{\textbf{0.009}} & \underline{\textbf{0.025}}
                     & \underline{\textbf{0.067}} & \underline{\textbf{0.220}} & 0.039 & 0.116   \\ \hline
\end{tabular}
\label{tab:pose-jitter-rolling}
}
\end{table*}

\vspace{0.3in}
\subsection{Failure Cases}
\begin{figure}[tp]
\centering
\subfloat{\includegraphics[width=0.24\textwidth]{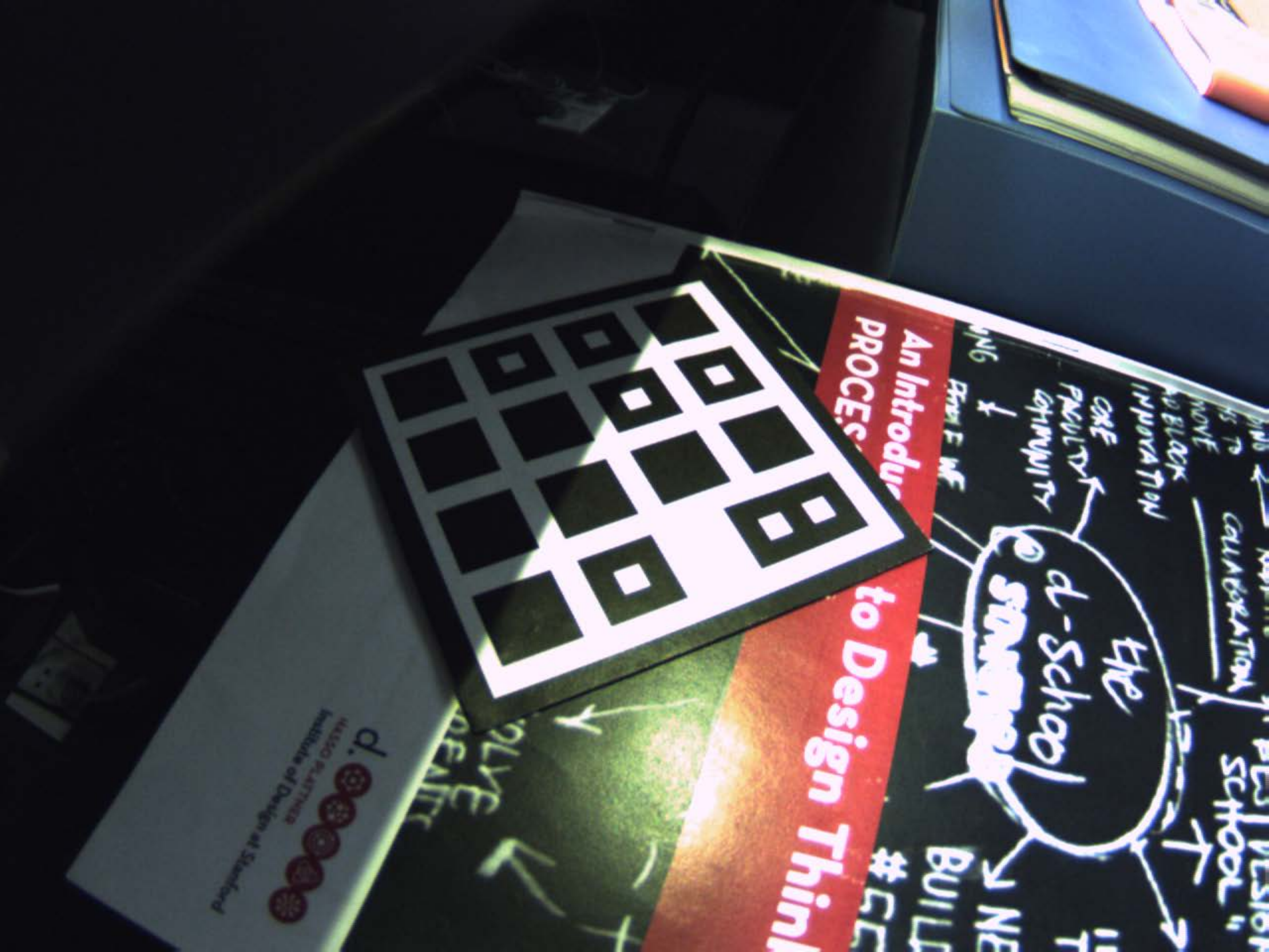}}
\hspace{0.01in}
\subfloat{\includegraphics[width=0.24\textwidth]{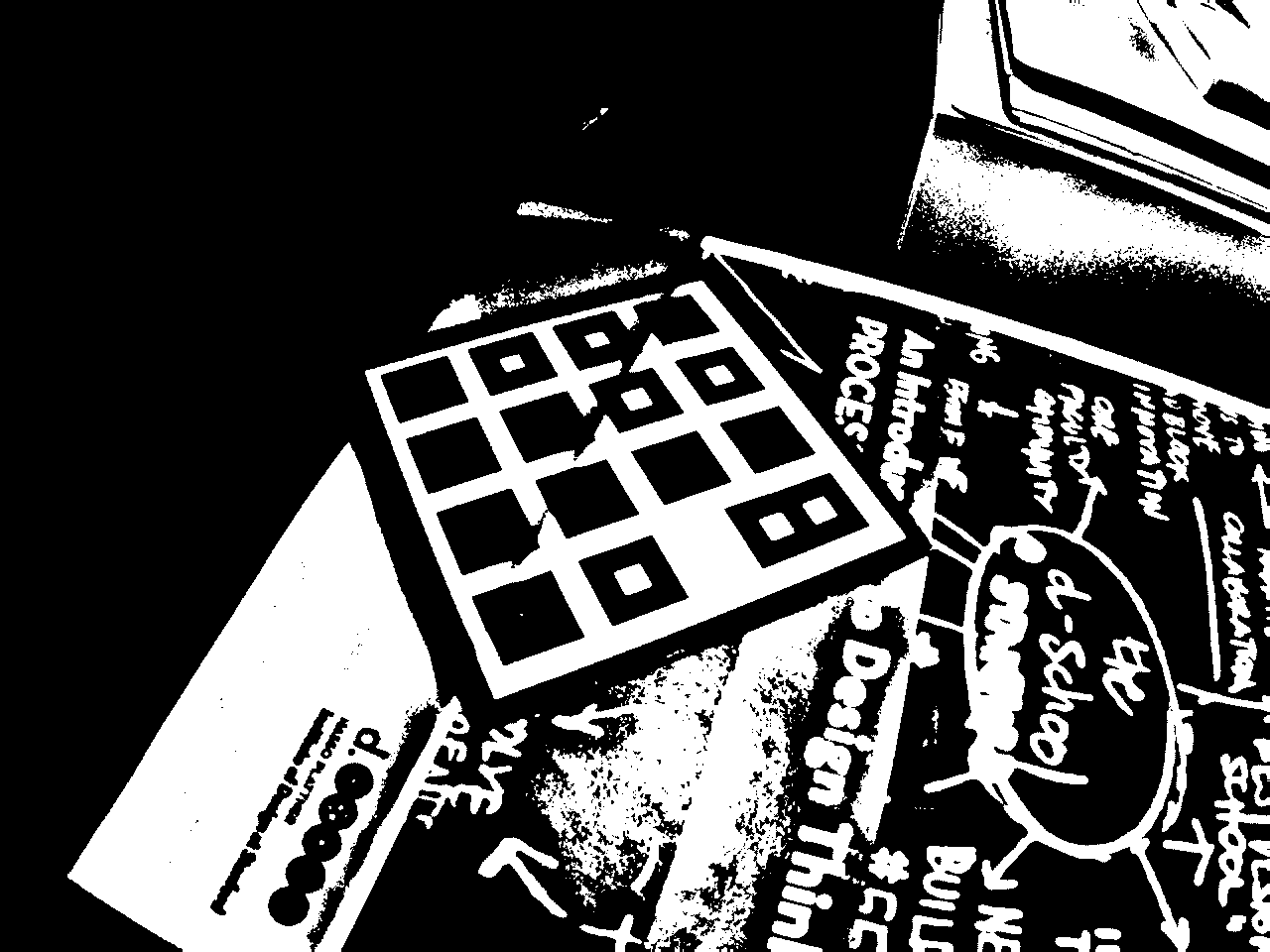}}
\hspace{0.01in}
\subfloat{\includegraphics[width=0.24\textwidth]{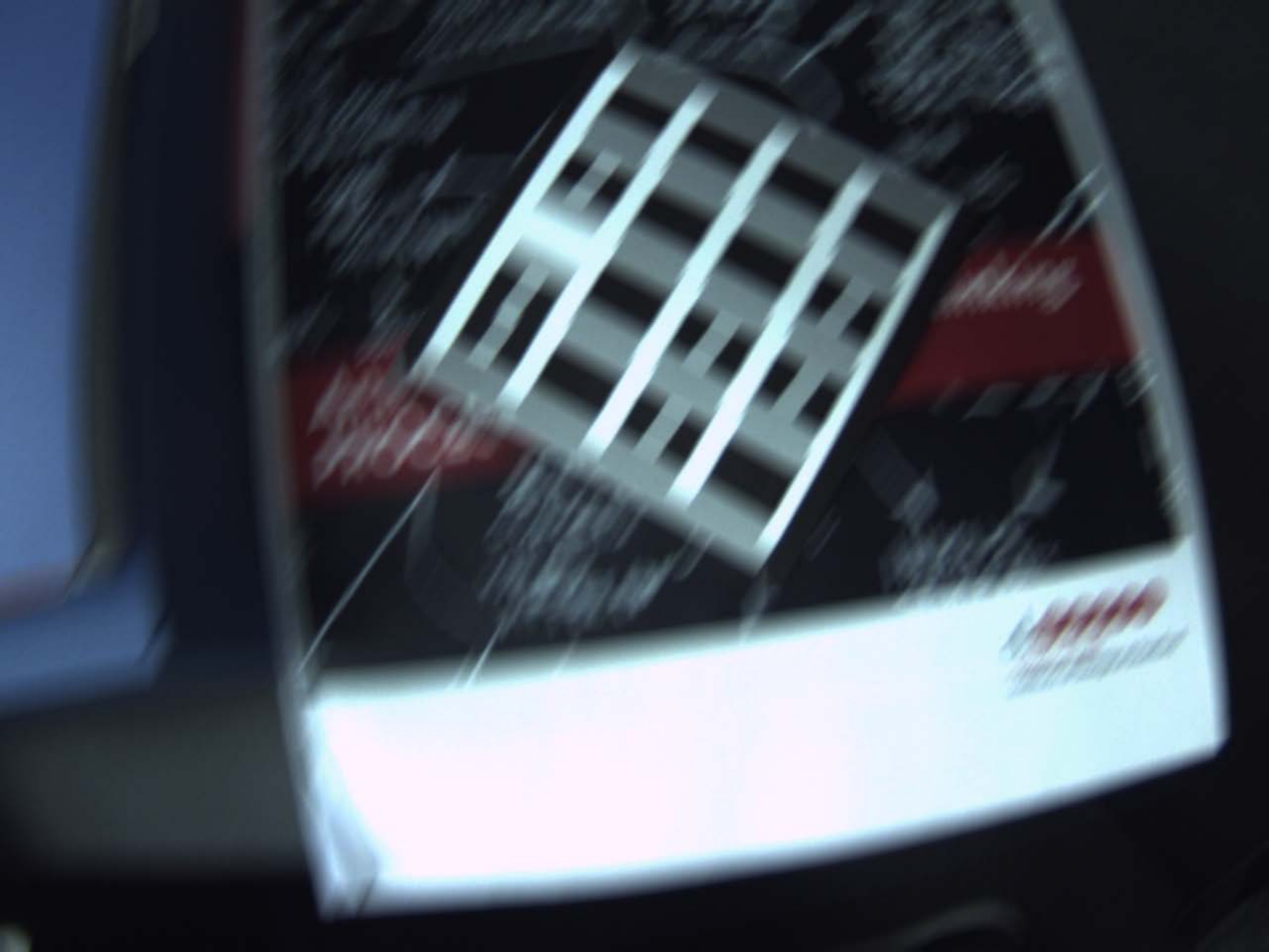}}
\hspace{0.01in}
\subfloat{\includegraphics[width=0.24\textwidth]{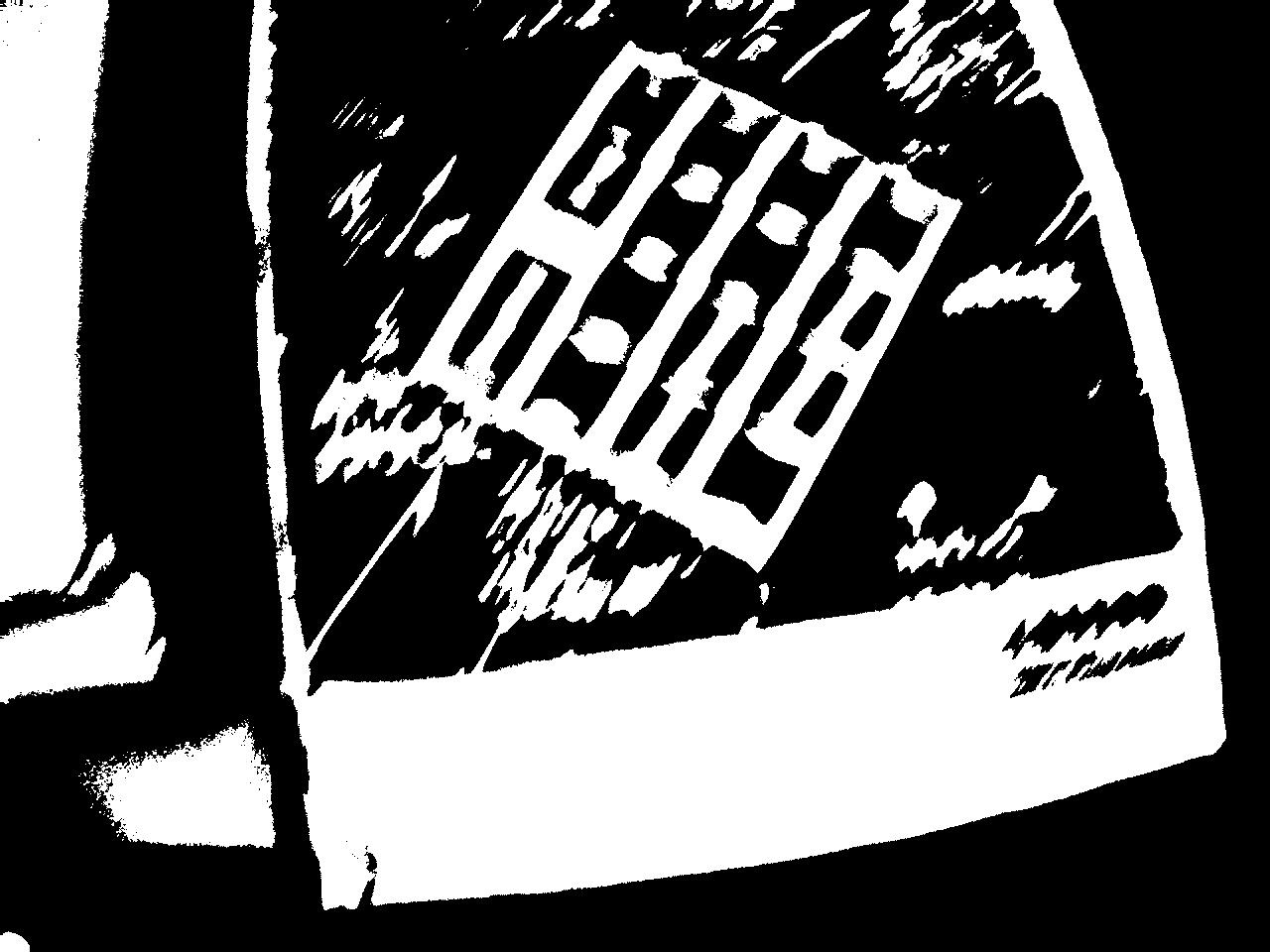}}
\caption{Failure cases. Top row shows one of the failure cases of dramatic lighting change, and bottom row shows one of the severe motion blur. On right, binarization results are shown for each case respectively.}
\label{fig:failure}
\end{figure}

TopoTag can handle lighting change and motion blur better due to our unique threshold map estimation and topological filtering modules\Blue{, see examples in \autoref{fig:rolling-test}}. However, it will fail to detect the markers where dramatic lighting change or severe motion blur happens over the marker region. \autoref{fig:failure} shows two typical failure cases as a result of dramatic lighting change and severe motion blur. Their binarization results show that markers' topological structure is dramatically changed. This change is the root cause of the detection failure.


\section{Conclusions} \label{sec:Conclusions}
We present TopoTag, a new topological-based fiducial marker and detection algorithm that utilizes topological information to achieve high robustness, and near-perfect detection accuracy. We show that all tag bits can be used to encode identities without sacrificing detection accuracy, thus achieving rich identification and scalability. TopoTag offers more feature correspondences for better pose estimation. We demonstrate that TopoTag achieves the best performance in various metrics including detection accuracy, localization jitter and accuracy, and at the same time supports occlusion and flexible shapes. We also collected a large dataset of TopoTag and other previous state-of-the-art tags for better evaluation, involving in-plane and out-of-plane rotations, image blur, various distances and cluttered background, etc.

For future research, we will explore novel ID encoding/decoding strategy. We believe that this is key for a better marker system with a goal of strong occlusion resistance and scalability in addition to high detection rate and long distance tracking range.

\ifCLASSOPTIONcaptionsoff
  \newpage
\fi



\bibliographystyle{IEEEtran}
\bibliography{IEEEabrv,bare_jrnl_compsoc}
%
%
%

\begin{IEEEbiography}[{\includegraphics[width=1in,height=1.25in,clip,keepaspectratio]{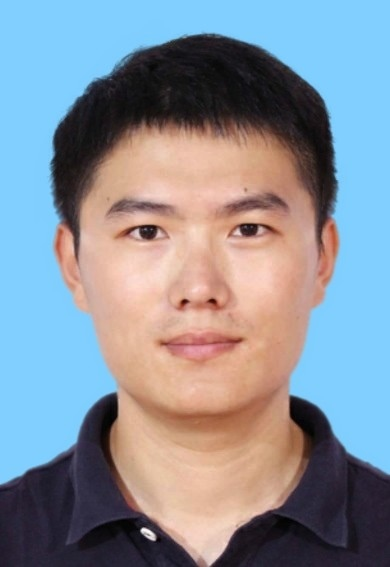}}]{Guoxing Yu}
received his B.Eng degree in electronic information engineering from Wuhan University of Science and Technology, Wuhan, China, in 2013, and the M.E. degree in information and communication engineering from Huazhong University of Science and Technology, Wuhan, China, in 2016.

He is currently with Guangdong Virtual Reality Co., Ltd. (aka. Ximmerse) as an algorithm engineer. Prior to joining Ximmerse, he was an algorithm engineer with Wuhan Guide Infrared Co., Ltd. Wuhan from Jul. 2016 to Aug. 2017. His research interests include computer vision, augmented reality and virtual reality.
\end{IEEEbiography}

\begin{IEEEbiography}[{\includegraphics[width=1in,height=1.25in,clip,keepaspectratio]{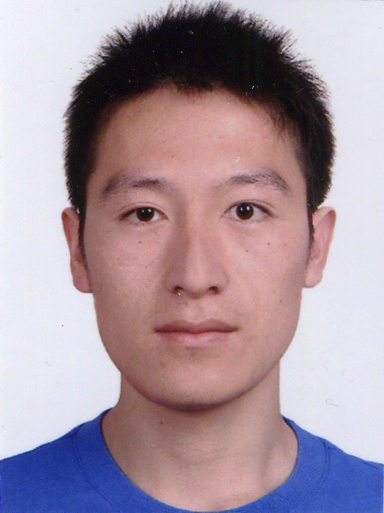}}]{Yongtao Hu}
received his B.Eng degree in computer science from Shandong University, Jinan, China, in 2010, and the Ph.D. degree in computer science from The University of Hong Kong, Hong Kong, in 2014.

He is currently with Guangdong Virtual Reality Co., Ltd. (aka. Ximmerse) as a research scientist. Prior to joining Ximmerse, he was a staff researcher with Image and Visual Computing Lab (IVCL), Lenovo Research, Hong Kong from Jan. 2015 to Oct. 2015, was a researcher assistant with IVCL from Jul. 2014 to Nov. 2014, and was a research intern at Internet Graphics Group in Microsoft Research Asia (MSRA) from Mar. 2010 to Jun. 2010. His research interests include computer vision, multimedia, machine learning, augmented reality and virtual reality.
\end{IEEEbiography}

\begin{IEEEbiography}[{\includegraphics[width=1in,height=1.25in,clip,keepaspectratio]{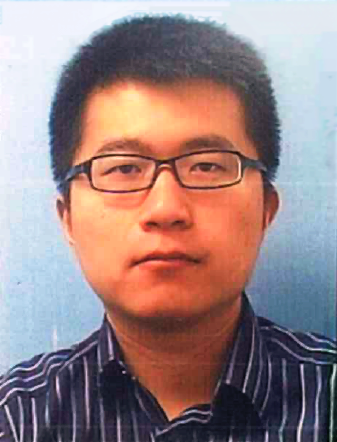}}]{Jingwen Dai (S'09 - M'12)}
received the B.E. degree in automation from Southeast University, Nanjing, China, in 2005, the M.E. degree in automation from Shanghai Jiao Tong University, Shanghai, China, in 2009, and the Ph.D. degree in mechanical and automation engineering from the Chinese University of Hong Kong, Hong Kong, in 2012.

He is currently with Guangdong Virtual Reality Co., Ltd. (aka. Ximmerse) as co-founder and chief technology officer. Prior to joining Ximmerse, he was a manager and advisory researcher with Image and Visual Computing Lab (IVCL), Lenovo Research, Hong Kong from Jan. 2014 to July 2015, and was a Post-Doctoral Research Associate with the Department of Computer Science, University of North Carolina at Chapel Hill, Chapel Hill, NC, USA from Oct. 2012 to Dec. 2013. His current research interests include computer vision and its applications in human-computer interaction, augmented reality and virtual reality.
\end{IEEEbiography}

%
%


\vfill


\end{document}